\documentclass{article}
\renewcommand{\baselinestretch}{0.968}
\usepackage{iclr2027_conference,times}
\usepackage[T1]{fontenc}
\usepackage[utf8]{inputenc}
\usepackage{amsmath,amssymb,bm,mathtools}
\usepackage{graphicx,booktabs,multirow,array,longtable,tabularx,adjustbox,longtable}
\usepackage{microtype}
\usepackage{placeins,float}
\usepackage{url}
\usepackage{algorithm}
\usepackage{algpseudocode}
\usepackage[hidelinks]{hyperref}
\hypersetup{pdftitle={SplatQ: Local Interaction-Aware Pruning of 3D Gaussian Splatting via Quadratic Selection},pdfauthor={Hongbin Ye, Haixu Liu, Shenwei Kang, Wenxiu Wei, Tinglan Wang, Hongdong Zhu, Fan Tang, Qi Gao, Yin Ma, Kai Wen, Hai Wei, Nuo Qian, Wei-Shi Zheng, Shaobo Chen, Fa-Ting Hong, Jingke Meng}}
\usepackage{etoolbox}
\makeatletter

\newif\ifSplatQCompactSpacing
\SplatQCompactSpacingtrue
\let\splatq@original@baselinestretch\baselinestretch
\renewcommand{\baselinestretch}{0.98}
\edef\splatq@original@parskip{\the\parskip}
\edef\splatq@original@textfloatsep{\the\textfloatsep}
\edef\splatq@original@floatsep{\the\floatsep}
\edef\splatq@original@intextsep{\the\intextsep}
\edef\splatq@original@dbltextfloatsep{\the\dbltextfloatsep}
\edef\splatq@original@dblfloatsep{\the\dblfloatsep}
\edef\splatq@original@abovedisplayskip{\the\abovedisplayskip}
\edef\splatq@original@belowdisplayskip{\the\belowdisplayskip}
\edef\splatq@original@abovedisplayshortskip{\the\abovedisplayshortskip}
\edef\splatq@original@belowdisplayshortskip{\the\belowdisplayshortskip}
\let\splatq@original@makecaption\@makecaption
\let\splatq@original@section\section
\let\splatq@original@subsection\subsection
\let\splatq@original@subsubsection\subsubsection
\let\splatq@original@paragraph\paragraph
\let\splatq@original@subparagraph\subparagraph

\long\def\@makecaption#1#2{%
  \ifSplatQCompactSpacing
    \begingroup
    \setlength{\baselineskip}{10.5pt}%
    \splatq@original@makecaption{#1}{#2}%
    \endgroup
  \else
    \splatq@original@makecaption{#1}{#2}%
  \fi}

\AtBeginEnvironment{figure}{%
  \ifSplatQCompactSpacing
    \setlength{\abovecaptionskip}{2pt}%
    \setlength{\belowcaptionskip}{0pt}%
  \fi}
\AtBeginEnvironment{figure*}{%
  \ifSplatQCompactSpacing
    \setlength{\abovecaptionskip}{2pt}%
    \setlength{\belowcaptionskip}{0pt}%
  \fi}
\AtBeginEnvironment{table}{%
  \ifSplatQCompactSpacing
    \setlength{\abovecaptionskip}{0pt}%
    \setlength{\belowcaptionskip}{2.5pt}%
  \fi}

\renewcommand{\section}{\@startsection{section}{1}{\z@}%
  {-1.4ex plus -0.2ex minus -0.1ex}{0.65ex plus 0.1ex}%
  {\large\sc\raggedright}}
\renewcommand{\subsection}{\@startsection{subsection}{2}{\z@}%
  {-1.0ex plus -0.2ex minus -0.1ex}{0.45ex plus 0.1ex}%
  {\normalsize\sc\raggedright}}
\renewcommand{\subsubsection}{\@startsection{subsubsection}{3}{\z@}%
  {-0.8ex plus -0.1ex minus -0.1ex}{0.35ex plus 0.1ex}%
  {\normalsize\sc\raggedright}}
\renewcommand{\paragraph}{\@startsection{paragraph}{4}{\z@}%
  {1.0ex plus 0.1ex minus 0.1ex}{-1em}{\normalsize\bf}}
\renewcommand{\subparagraph}{\@startsection{subparagraph}{5}{\z@}%
  {1.0ex plus 0.1ex minus 0.1ex}{-1em}{\normalsize\sc}}

\patchcmd{\@maketitle}{\vskip 0.3in minus 0.1in}{\vskip 6pt}{}%
  {\PackageError{compact_spacing}{Title spacing patch failed}%
    {Check the supplied ICLR title definition.}}

\renewenvironment{abstract}{%
  \par\vspace{2pt}%
  \centerline{\large\sc Abstract}%
  \vspace{1pt}%
  \begin{list}{}{%
    \setlength{\leftmargin}{\leftmargini}%
    \setlength{\rightmargin}{\leftmargin}%
    \setlength{\topsep}{2pt}%
    \setlength{\partopsep}{0pt}%
    \setlength{\parsep}{0pt}%
    \setlength{\itemsep}{0pt}%
  }\item\relax
}{\end{list}\par\vspace{2pt}}

\AtBeginDocument{\raggedbottom}

\newcommand{\SplatQRestoreSpacing}{%
  \SplatQCompactSpacingfalse
  \let\baselinestretch\splatq@original@baselinestretch
  \selectfont
  \setlength{\parskip}{\splatq@original@parskip}%
  \setlength{\abovedisplayskip}{\splatq@original@abovedisplayskip}%
  \setlength{\belowdisplayskip}{\splatq@original@belowdisplayskip}%
  \setlength{\abovedisplayshortskip}{\splatq@original@abovedisplayshortskip}%
  \setlength{\belowdisplayshortskip}{\splatq@original@belowdisplayshortskip}%
  \let\section\splatq@original@section
  \let\subsection\splatq@original@subsection
  \let\subsubsection\splatq@original@subsubsection
  \let\paragraph\splatq@original@paragraph
  \let\subparagraph\splatq@original@subparagraph
}
\newcommand{\SplatQRestoreFloatSpacing}{%
  \setlength{\textfloatsep}{\splatq@original@textfloatsep}%
  \setlength{\floatsep}{\splatq@original@floatsep}%
  \setlength{\intextsep}{\splatq@original@intextsep}%
  \setlength{\dbltextfloatsep}{\splatq@original@dbltextfloatsep}%
  \setlength{\dblfloatsep}{\splatq@original@dblfloatsep}%
}
\makeatother

\title{SplatQ: Local Interaction-Aware Pruning of 3D Gaussian Splatting via Quadratic Selection}
\author{\begin{tabular}{c}
Hongbin Ye\textsuperscript{*}, Haixu Liu\textsuperscript{*}, Shenwei Kang\textsuperscript{*},
Wenxiu Wei, Tinglan Wang, Hongdong Zhu,\\
Fan Tang, Qi Gao, Yin Ma, Kai Wen, Hai Wei, Nuo Qian, Wei-Shi Zheng, \\
Shaobo Chen\textsuperscript{$\dagger$},
Fa-Ting Hong\textsuperscript{$\dagger$}, Jingke Meng\textsuperscript{$\dagger$}\\[4pt]
\texttt{mengjk3@mail.sysu.edu.cn}\\[2pt]
{\small \textsuperscript{*}Equal contribution \quad \textsuperscript{$\dagger$}Corresponding authors}
\end{tabular}}
\iclrfinalcopy %

\begin{document}
\maketitle

\begin{abstract}

Existing 3D Gaussian Splatting (3DGS) pruning methods primarily rely on global Top-K selection based on independent importance scores, lacking regional budget control and ignoring joint deletion effects. Consequently, they cannot ensure balanced reconstruction quality across a scene. We propose SplatQ, a combinatorial pruning algorithm that coordinates regional budget allocation and joint Gaussian selection under a fixed budget. SplatQ partitions Gaussians by spatial proximity and appearance similarity into size-limited blocks, decomposing global selection into subproblems of a specific size. It then smooths importance-based block retention quotas to prevent excessive budget concentration. Finally, it constructs a quadratic approximation of joint deletion effects using the responses of representative pixel colors to changes in Gaussian opacity. Individual responses, pairwise interactions, and quota penalties are incorporated into a quadratic unconstrained binary optimization (QUBO) objective, solved using simulated annealing (SA) or a coherent Ising machine (CIM) to jointly select Gaussians. Experiments on 5 benchmarks demonstrate that SplatQ outperforms state-of-the-art pruning methods in aggregate reconstruction quality, spatial quality balance, and rendering speed. Ablations confirm that budget smoothing improves local detail reconstruction, while interaction modeling improves Gaussian selection quality under a fixed budget.

\end{abstract}
\section{Introduction}
\label{sec:introduction}

Gaussian Splatting ~\citep{kerbl2023_3dgs} enables high-quality real-time rendering through an explicit Gaussian representation, providing a practical foundation for the reconstruction and interaction of 3D digital assets. Complex scenes, however, often contain large numbers of Gaussians, whose storage, memory, and rendering costs limit asset transmission and deployment~\citep{lightgaussian2024,eagles2024}. Gaussian pruning reduces these costs by removing primitives. Methods such as LightGaussian~\citep{lightgaussian2024} and PUP 3D-GS~\citep{pup3dgs2025} combine importance estimation with post-pruning fine-tuning to balance model size and rendering quality.

Under a fixed budget and aggressive pruning, we observe that results with similar overall image-quality metrics can exhibit different patterns of local degradation: structures and textures remain clear in some regions, while other regions become visibly blurred or distorted. Existing methods also make different regional trade-offs. For example, PUP 3D-GS discusses the trade-off between foreground detail and background quality~\citep{pup3dgs2025}, whereas GaussianPOP emphasizes the preservation of detail-rich regions~\citep{gaussianpop2026}. These choices have their own motivations, but they raise a question: when the number of retained Gaussians is strictly limited, how can pruning preserve both overall rendering quality and reconstruction quality across different regions?

\begin{figure}[t]
\centering
\includegraphics[width=\linewidth]{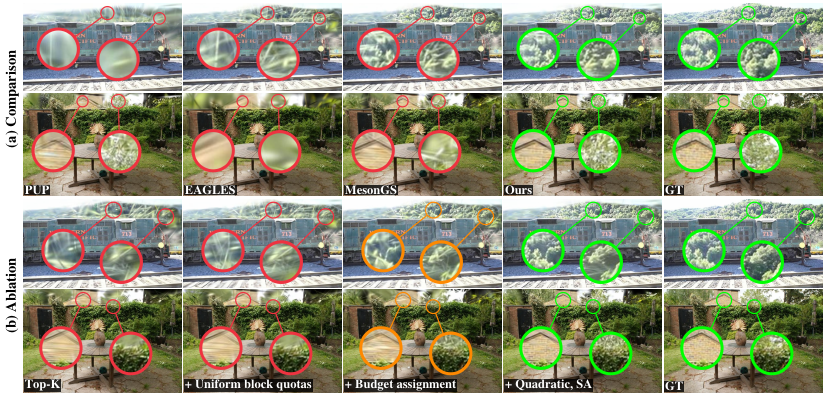}
\caption{\textbf{Addressing spatial quality imbalance in 3DGS pruning.}
(a) Comparisons with baselines illustrate uneven detail preservation across image regions, while SplatQ preserves the highlighted structures more consistently.
(b) Component ablations show how block quotas, budget assignment, and quadratic selection progressively improve spatial quality balance over global Top-$K$ pruning.}
\label{fig:teaser}
\end{figure}

We have two observations from previous global importance ranking works~\citep{lightgaussian2024,pup3dgs2025,refine2026}: 
\textbf{First, global ranking does not directly control how the retention budget is distributed across a scene.} Such methods score each Gaussian separately and retain the highest-scoring Gaussians globally~\citep{lightgaussian2024,pup3dgs2025,refine2026}. The number retained in each region therefore depends primarily on the score distribution. If Gaussians in certain regions consistently receive higher scores, the limited budget may concentrate there, leaving other regions with insufficient representational capacity. Thus,relying on a global ranking cannot guarantee that the rendering quality of all regions is well preserved under aggressive pruning
Figure~\ref{fig:teaser}(a) illustrates this spatial quality imbalance:
pruning baselines preserve some structures while losing detail in other
regions of the same scene.
\textbf{Second, Gaussian pruning under a fixed budget is a combinatorial optimization problem.} It requires selecting, subject to a retention constraint, a set of Gaussians that renders well together. Independent scoring followed by Top-$K$ ranking reduces this set-selection problem to a comparison of individual scores. Yet multiple Gaussians may jointly support the same pixel or make overlapping contributions during rendering~\citep{kerbl2023_3dgs}. 
Because Gaussian contributions overlap, the degradation caused by removing multiple Gaussians is not strictly additive and cannot be inferred from individual deletion costs.Highly overlapping Gaussians might all receive high individual scores and be redundantly retained, even though preserving just one would suffice.
Consequently, accurate individual scores alone cannot determine which set of Gaussians should be retained. Ignoring interactions may prevent the limited budget from being used for a combination that achieves better reconstruction quality.

To address these limitations, we propose \textbf{SplatQ}, a post-training Gaussian pruning framework that performs similarity-guided partitioning, budget assignment, and interaction-aware quadratic selection in sequence. SplatQ aims to reduce disparities in local reconstruction quality under aggressive pruning and use the retained Gaussian budget more effectively, improving overall and local reconstruction quality with the same number of Gaussians. \textbf{First,} SplatQ partitions Gaussians into local blocks according to spatial proximity and appearance similarity. These blocks provide units for budget assignment while keeping the selection problems at a specific size. \textbf{Second,} to control local retention counts, SplatQ derives an initial quota for each block from importance estimates and adjusts the quota distribution to reduce the risk of concentrating the budget in a few blocks while leaving others with insufficient capacity. \textbf{Finally,} to account for the joint effects of deleting multiple Gaussians, SplatQ uses rendering responses to changes in opacity multipliers to estimate individual deletion effects and pairwise interactions. 
Figure~\ref{fig:teaser}(b) illustrates how block quotas, budget assignment, and quadratic selection progressively improve detail preservation across the highlighted regions.
It then formulates and solves a QUBO~\citep{glover2018qubo} under each block's quota to select a retained set with a smaller estimated joint deletion effect. Because solving QUBOs with classical methods requires more computation than direct Top-$K$ ranking, we also examine a coherent Ising machine (CIM) as a hardware solver backend and assess its potential to improve solving efficiency. After pruning, SplatQ fine-tunes the retained Gaussians to recover rendering quality.

We conduct extensive experiments on 27 scenes from 5 benchmarks covering indoor, outdoor, and objects, comparing SplatQ with eight open-source pruning methods. Under a common 90\% pruning setting, SplatQ achieves the best overall image quality and rendering speed among the compared methods, while providing more uniform reconstruction quality across image regions. We further evaluate different pruning ratios, fine-tuning budgets, and combinations of pruning rounds and per-round fine-tuning iterations, facilitating comparisons with the reported settings of methods without released code. SplatQ also remains effective in a picture-free setting, where neither pruning nor fine-tuning uses the original training photographs and fine-tuning relies on images rendered by the fixed pretrained model. Even under this constraint, SplatQ surpasses the strongest baseline, MesonGS~\citep{mesongs2024}, in dataset-macro PSNR. Ablation studies further validate the effectiveness of partition-based budget assignment and interaction-aware combinatorial optimization, while experiments with CIM demonstrate that an alternative solver backend can also produce competitive selection results.

\section{Related work}
\label{sec:related}
\paragraph{Gaussian compression and pruning.}
Existing 3DGS pruning methods often rank or sample Gaussians by rendering contributions, using compositing weights, coverage, volume, opacity, or gradients~\citep{eagles2024,lightgaussian2024,mesongs2024,trimmingfat2024,coreset3dgs2026}. These criteria may overlook compensation by other Gaussians and undervalue sparsely observed regions. Other methods use parameter sensitivity or its approximations~\citep{pup3dgs2025,refine2026,c3dgs2024}, or estimate single-Gaussian deletion error relative to renderings of the current or original model~\citep{gaussianpop2026,potr2026}. However, sensitivity is not equivalent to actual deletion error, and individual deletion errors do not directly characterize joint removal. Learned scores, masks, or sparse representations enable reconstruction-driven adaptation but require additional learning~\citep{rap2026,compact3dgs2024,lp3dgs2024,gaussianspa2025,maskgaussian2025}. Further methods consider relationships among Gaussians through neighboring appearance differences or local overlap~\citep{omg2025,qirf2026}, or merge Gaussians into new representatives~\citep{ghap2025}. In contrast, SplatQ partitions pretrained Gaussians by spatial and appearance similarity, initializes and smooths blockwise retention quotas under an exact global budget, and jointly selects Gaussians within each block using a quadratic objective built from individual and pairwise linearized rendering responses.

\paragraph{Balancing Rendering Quality Across Image Regions.}
NeRF++~\citep{nerfpp2020}, Mip-NeRF 360~\citep{barron2022mipnerf360}, and HoGS~\citep{hogs2025} study how to represent near and distant content during scene reconstruction. For a pretrained 3DGS model, pruning additionally requires deciding how many primitives to retain in different parts of the scene under a limited Gaussian budget. PUP 3D-GS reports a trade-off between foreground detail and background quality under aggressive pruning~\citep{pup3dgs2025}. These regional differences motivate SplatQ to partition Gaussians with similar spatial positions and appearances into size-limited blocks, smooth the block quotas derived from importance estimates while preserving the exact global retention count, and perform joint selection within each block. In addition to overall reconstruction quality, we evaluate pruning results by measuring quality differences across image patches.

\begin{figure}[t]
    \centering
    \includegraphics[width=\linewidth]{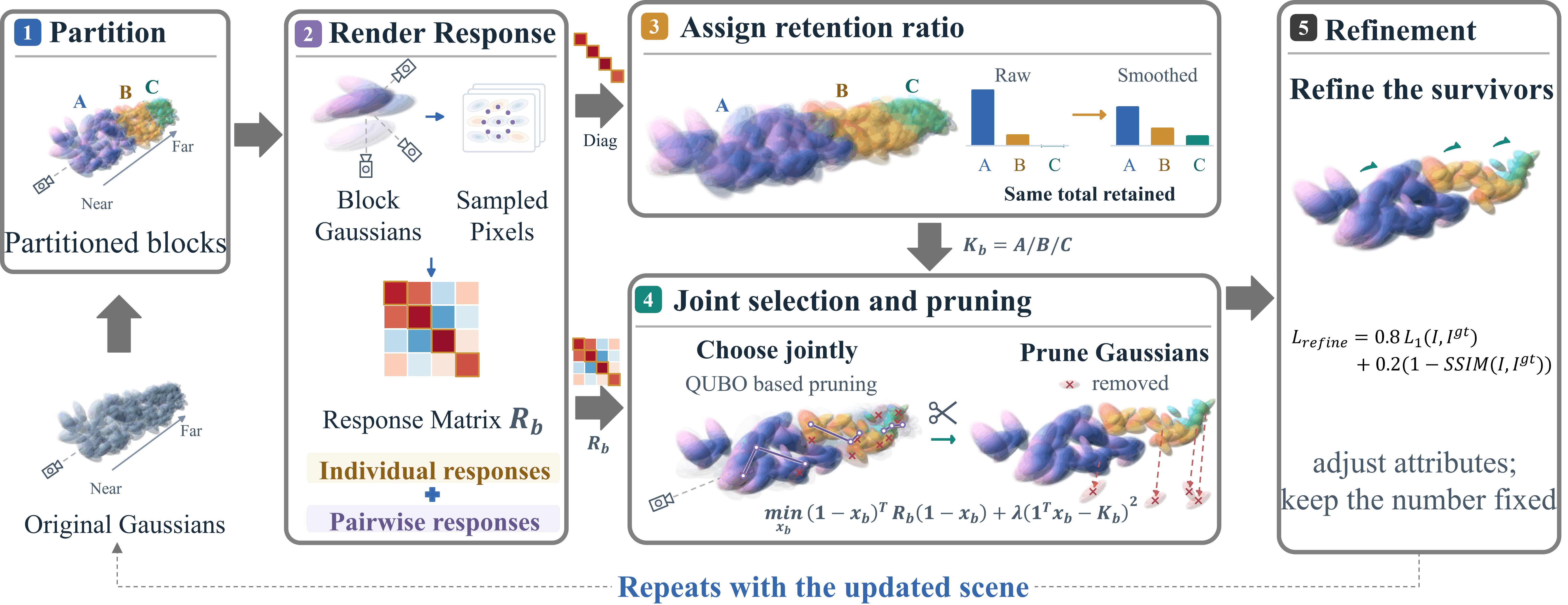}
    \caption{Overview of SplatQ.
    We partition Gaussians, compute a response matrix $R_b$ for each block using sampled pixels
    from multiple views.
    Diagonal scores determine block retention quotas, while the full matrix is used to formulate a blockwise QUBO that determines which Gaussians to delete.
    The retained Gaussians are then fine-tuned without changing their count.
    For iterative pruning, the process repeats on the updated model.}
    \label{fig:method}
\end{figure}

\section{Method}
\label{sec:method}

Given a pretrained 3D Gaussian Splatting model with $N$ Gaussians,
SplatQ aims to retain exactly \(K\) Gaussians while accounting for 
their joint effects on the rendered image and reducing disparities in retention across scene regions. 
We represent the model as $\mathcal G
=
\left\{
G_i=
\left(
\boldsymbol\mu_i,\mathbf s_i,\mathbf r_i,
\mathbf c_i,\mathbf h_i,o_i
\right)
\right\}_{i=1}^{N}$
where each Gaussian is parameterized by its center $\boldsymbol\mu_i$,
scale $\mathbf s_i$, rotation quaternion $\mathbf r_i$,
zeroth-order (DC) and higher-order spherical-harmonic coefficients
$\mathbf c_i$ and $\mathbf h_i$, and opacity $o_i$.

Previous methods suffer from two limitations: 1) They may produce imbalanced Gaussian retention across near and distant regions. 2) Moreover, independent Gaussian scores cannot
capture the joint rendering effects of simultaneous deletions. SplatQ
addresses these limitations through blockwise retention ratio allocation and
interaction-aware quadratic selection, respectively.

SplatQ decomposes this global selection problem into five stages.
Partitioning in Section~\ref{sec:partition} first converts Gaussians $\mathcal G$ into several size-limited blocks $\{\mathcal C_b\}$. 
Opacity response modeling in Section~\ref{sec:response} then produces a response matrix $R_b$ for each block.
Budget allocation in Section~\ref{sec:allocation} uses part of the matrix $R_b$ to assign a retention quota $K_b$ to every block, with $\sum_b K_b=K$. 
Given $R_b$ and $K_b$, quadratic selection in Section~\ref{sec:selection} determines which $K_b$ block members to retain. 
Finally, we fine-tune the retained attributes without changing the count of Gaussians in Section~\ref{sec:refinement}.

\subsection{Similarity Guided Partition}
\label{sec:partition}

Independent importance scores do not capture how the image changes when
several Gaussians are deleted together. To better describe these changes, we approximate their joint rendering effects using a quadratic objective over binary retention variables. Each variable equals 1 if the corresponding Gaussian is retained and 0 if it is deleted, as detailed in Section~\ref{sec:selection}.
Formulating this optimization problem over the entire 3D-GS model requires \(N\) binary variables and an \(N\times N\) matrix of pairwise coefficients. 
For models containing millions of Gaussians, the variable count and memory requirements make this approach impractical.
We therefore divide the Gaussians into blocks of at most \(M\) members and solve a smaller pruning problem within each block.

Simply dividing Gaussians into consecutive blocks in their stored order ignores their spatial and appearance relationships. 
Nearby Gaussians with similar appearance may fall into different blocks, preventing their joint effects from being considered in the same pruning problem.
We therefore use spatial proximity and appearance similarity to guide partitioning, so that potentially redundant Gaussians can be evaluated together.
Finding related Gaussians through all pairwise comparisons itself requires quadratic work. 
We therefore use a 3D grid and Morton ordering to efficiently generate candidate pairs, whose spatial and appearance similarity is measured by an affinity. 

\paragraph{Candidate generation.}Gaussian centers are assigned to grid cells, 
and the occupied cells are ordered by their Morton order to preserve spatial proximity. 
Then we divide the resulting Gaussian sequence into groups of approximately equal size, each within a size limit of M.
These groups are temporary search windows rather than the partition results $\mathcal C_b$.

\paragraph{Affinity graph.}To encourage spatially and visually related Gaussians to be assigned to the same block, we define the following pairwise affinity.
For Gaussian $i$, let $\boldsymbol\mu_i$ denote its center, 
$\mathbf s_i$ its activated scale vector, and $\mathbf c_i$ its three-channel DC spherical-harmonic descriptor.
For each candidate pair, we define the scale-aware affinity as
\begin{equation}
 a_{ij}=\exp\!\left[
 -\sum_{d=1}^{3}
 \frac{(\mu_{i,d}-\mu_{j,d})^2}
 {s_{i,d}^{2}+s_{j,d}^{2}+\epsilon_g}
 -\|\mathbf c_i-\mathbf c_j\|_2^2
 \right].
 \label{eq:affinity}
\end{equation}
The first term measures center separation relative to the Gaussian scales
and the second term reduces the affinity between spatially close
Gaussians with different appearance. 
Each Gaussian retains at most the $k$ candidates with the largest affinities, 
and the retained pairs form a sparse graph with \(a_{ij}\) as edge weights.
The candidate groups depend on the grid origin,
so one pass can miss pairs separated by a group boundary. 
We repeat candidate generation and affinity calculating with a shifted grid and merge both passes into one sparse graph. 
Repeated edges record their weights only once.
The merged sparse graph then guides the construction of the final partition.

\paragraph{Partition.}We next convert the sparse graph into the final size-limited blocks.
Using Morton order in the first pass, we select one seed from each interval containing at most \(M\) Gaussians 
and grow all blocks in parallel. Each unassigned Gaussian proposes
the block reached by its strongest edge to an assigned member, and proposals
are accepted only while that block has capacity. Gaussians not reached by
the graph fill the blocks in Morton order. This procedure yields
disjoint blocks $\{\mathcal C_b\}_{b=1}^{B}$ with
$n_b=|\mathcal C_b|\leq M$. 
These blocks serve as the units for response modeling, quota allocation, and quadratic selection in the following sections.

\subsection{Opacity Response Modeling}
\label{sec:response}

The partition defines the local Gaussian sets, 
but it does not determine how many Gaussians to retain in each block or which ones to retain.
For each block, we select pixels in rendered training views 
and calculate how their RGB values change when a block member's opacity changes.
These measurements provide both individual importance scores for
each block's budget allocation in Section~\ref{sec:allocation} 
and pairwise interactions for quadratic selection of Gaussians in ~\ref{sec:selection}.

For each block $b$ and camera view $v$, we use the projected Gaussian centers
as the initially selected pixel coordinates. We augment this set with neighboring pixels
obtained by offsetting each projected center. After removing duplicate
locations and discarding those outside the image, we collect the selected
pixels $p$ of this view $v$ as pairs $(v,p)$ into $\mathcal P_b$. These representative pixels are used to evaluate
the RGB changes as Gaussian opacity changes.

Let $L_{vp}$ denote the number of Gaussians covering pixel $p$ in view $v$.
We index these Gaussians in front-to-back order as
$g_{1},\ldots,g_{L_{vp}}$. Let $\alpha_{\ell}$ denote the original alpha of
$g_{\ell}$, given by its activated opacity times its projected Gaussian
value. We introduce a continuous multiplier $\eta_i\in[0,1]$ for each
Gaussian's activated opacity. The modified alpha and incoming transmittance are
\begin{equation}
 \alpha_{\ell}(\boldsymbol\eta)
 =\eta_{g_\ell}\alpha_{\ell},
 \qquad
 T_{\ell}(\boldsymbol\eta)
 =\prod_{r=1}^{\ell-1}\bigl(1-\alpha_{r}(\boldsymbol\eta)\bigr).
 \label{eq:opacity-alpha-transmittance}
\end{equation}
Using front-to-back alpha compositing, the rendered RGB channel
$q\in\{1,2,3\}$ at this pixel is
\begin{equation}
 \begin{aligned}
 I_{vpq}(\boldsymbol\eta)
 ={}&\sum_{\ell=1}^{L_{vp}}
 T_{\ell}(\boldsymbol\eta)
 \alpha_{\ell}(\boldsymbol\eta)c_{\ell}+T_{L_{vp}+1}(\boldsymbol\eta)c_q^{\mathrm{bg}}.
 \end{aligned}
 \label{eq:opacity-compositing}
\end{equation}
Here $c_{\ell}$ is the Gaussian's view-dependent color component,
$c_q^{\mathrm{bg}}$ is the background color component, and
$T_{L_{vp}+1}$ is the transmittance remaining after all contributors.

We differentiate each sampled RGB channel with respect to the opacity
multiplier of every block member. For $i\in\mathcal C_b$, the corresponding
entry of $J_b\in\mathbb R^{3|\mathcal P_b|\times n_b}$ is
\begin{equation}
 (J_b)_{(v,p,q),i}
 =
 \left.\frac{\partial I_{vpq}(\boldsymbol\eta)}{\partial\eta_i}
 \right|_{\boldsymbol\eta=\mathbf 1}.
 \label{eq:opacity-jacobian}
\end{equation}
Rows correspond to sampled view-pixel-channel triples $(v,p,q)$,
and columns correspond to the Gaussians in block $b$.
The derivatives define the opacity-response Gram matrix $R_b=J_b^{\mathsf T}J_b$ and diagonal score $D_i=(R_b)_{ii},\quad i\in\mathcal C_b$.
The diagonal score $D_i$ sums the squared color derivatives for opacity of Gaussian $i$
over all selected pixels, views, and RGB channels.
A larger $D_i$ therefore indicates that Gaussian $i$ has a greater influence on the rendered image.
Each off-diagonal entry is the signed inner product between two response columns. 
Positive entries indicate reinforcing linearized deletion effects, 
whereas negative entries indicate partial cancellation. 
It serves as the interactions used in Section~\ref{sec:selection}.

\subsection{Budget Assignment}
\label{sec:allocation}

This stage determines how many Gaussians each block retains.
We use the diagonal scores $D_i$ defined in Section~\ref{sec:response}
to assign each block a retention quota $K_b$, with $\sum_b K_b=K$.
We first rank all Gaussians by $D_i$ and count the global Top-$K$ members contained in each block: $K_b=\left|\mathcal C_b\cap\operatorname{TopK}\bigl(\{D_i\}_{i=1}^{N}\bigr)\right|$.
Top-$K$ returns exactly $K$ indices, with equal scores resolved by global
index. The resulting initial quotas therefore satisfy $\sum_bK_b=K$.
Global Top-$K$ selection may assign zero retained Gaussians to a block if none of its members ranks among the top $K$. 
Thus, some blocks may lose all their Gaussians.
These blocks may include distant scene regions. To reduce this imbalance, We smooth the initial quotas to reduce disparities in retention across blocks. 
For initial quotas $0<K<N$, the smoothed quotas are
\begin{equation}
 r_b=\frac{K_b+\epsilon_a}{(K/N)n_b+\epsilon_a},
 \qquad
 \widehat K_b=
 K\frac{n_b\log(1+\beta r_b)}
 {\sum_{b'=1}^{B}n_{b'}\log(1+\beta r_{b'})}.
 \label{eq:smooth-budget}
\end{equation}
Here $r_b$ compares the initial quota with the proportional quota
$(K/N)n_b$, while $n_b$ denotes block capacity. Increasing
$\beta>0$ reduces disparities among block retention ratios. The
pseudocount $\epsilon_a>0$ handles blocks with zero initial quota.
We round each continuous quota down and allocate the remaining units to
eligible blocks in descending order of their fractional remainders. The
final quotas satisfy $0\leq K_b\leq n_b$ and $\sum_bK_b=K$ exactly.

\subsection{Quadratic Selection}
\label{sec:selection}

Budget allocation fixes the retained count $K_b$ for each block. We now use
$R_b$ to determine which $K_b$ members should be retained while accounting
for their joint effects on the rendered image.
Let $\mathbf x_b\in\{0,1\}^{n_b}$ denote the block retention vector, 
where $x_{b,i}=1$ indicates that the corresponding Gaussian is retained. 
We define the QUBO problem as
\begin{equation}
\min_{\mathbf x_b\in\{0,1\}^{n_b}}
E_b(\mathbf x_b)=
(\mathbf 1-\mathbf x_b)^{\mathsf T}
R_b(\mathbf 1-\mathbf x_b)
+
\lambda
\left(\mathbf 1^{\mathsf T}\mathbf x_b-K_b\right)^2.
\label{eq:retention-qubo}
\end{equation}
The first term approximates the rendering change caused by jointly deleting the Gaussians with $x_{b,i}=0$. 
The second term penalizes deviations from the allocated quota $K_b$. 
Each block's QUBO can be solved independently using greedy selection, simulated annealing, or an Ising solver. 
After repairing infeasible solver outputs when necessary, we combine the block masks into a model containing exactly $K$ Gaussians.
Appendix~\ref{app:qubo} and Appendix~\ref{app:solver} provide the algebraic derivations and solver details.

\subsection{Fine-Tuning and Iterative Pruning}
\label{sec:refinement}

We combine the result binary masks $\mathbf x_b$ from all blockwise QUBOs and fine-tune
the retained Gaussians with densification disabled.
The refinement loss is
$\mathcal L_{\mathrm{refine}}
=0.8\,\mathcal L_1(I,I^{\mathrm{gt}})
+0.2\bigl(1-\operatorname{SSIM}(I,I^{\mathrm{gt}})\bigr)$,
where $I$ and $I^{\mathrm{gt}}$ are the rendered and ground-truth images.
For iterative pruning, we repeat partitioning, opacity response
computation, budget allocation, and quadratic selection on the refined
model until the target retention ratio is reached.

\section{Experiments}
\label{sec:experiments}

\subsection{Experimental Settings}

\paragraph{Datasets and metrics.}
We evaluate SplatQ on 27 scenes from 5 benchmarks: Mip-NeRF 360 (9), Deep Blending (2), Tanks and Temples (2), NeRF Synthetic (8), and Ref-NeRF/Shiny Blender (6). These benchmarks cover indoor scenes,  outdoor scenes, and objects. We compare against unpruned 3DGS and eight pruning baselines, reporting average PSNR, SSIM, LPIPS-VGG, and rendering FPS. To assess spatial quality balance, we divide each test image into a $16\times16$ grid of non-overlapping patches and compute PSNR against the corresponding ground-truth region for each patch. For each image, we calculate the variance and range of its 256 patch PSNR values. We then average these statistics over test views within each scene and equally over scenes within each dataset. Considered alongside full-image PSNR, lower patch variance and range indicate smaller quality differences across image regions. Macro scores average the five dataset means equally. FPS is measured on Ubuntu 22.04 servers equipped with RTX 5090 GPUs and Intel Xeon Gold 6459C processors.

\paragraph{Experimental protocol.}
All methods start from the same 3DGS model pretrained for 30k iterations for each scene and follow a shared two-round pruning and fine-tuning schedule. The first round removes 80\% of the original Gaussians, and the second removes 50\% of the survivors, leaving 10\% of the original Gaussians. Each round includes 10k fine-tuning iterations, for 20k in total. 
Baselines that originally did not use fine-tuning receive the same schedule.
All methods use the refinement loss defined in Section~\ref{sec:refinement}, with densification disabled. 
We preserve each baseline's Gaussian importance estimation procedure. 
For methods that do not natively enforce an exact retention count, we apply Top-$K$ selection using their importance scores to meet the target retention ratio. To facilitate comparison with reported results from methods without released code, we additionally evaluate pruning ratios from 10\% to 90\%, fine-tuning budgets from 0 to 30k iterations, and combinations of one to four pruning rounds with 5k to 30k fine-tuning iterations per round at 90\% final removal. We also evaluate a \emph{picture-free} setting mentioned in Section ~\ref{sec:introduction}.

\paragraph{Ablation settings.}
Global Top-$K$ retains the $K$ Gaussians with the highest individual importance scores $D_i$ across the entire model. Uniform block quotas assigns the same retention ratio to every block, whereas budget assignment uses our smoothed quotas. Both variants retain the highest-scoring Gaussians within each block according to $D_i$. The Quadratic variants use the full objective with smoothed quotas and are solved using greedy selection, simulated annealing (SA), or a coherent Ising machine (CIM). Quadratic, uniform quotas instead combines uniform block quotas with greedy selection. For each variant, we repeat the complete pruning and fine-tuning pipeline five times with random seeds 0 through 4, starting from the same pretrained models.

\subsection{Comparison with Strong Baseline}

\begin{figure}[ht]
\centering
\includegraphics[width=\linewidth]{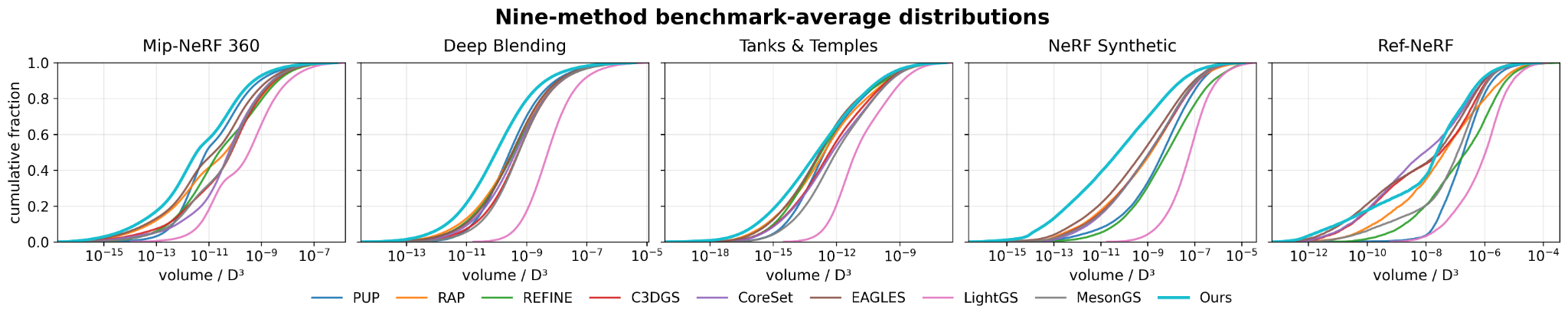}
\vspace{-15pt}
\caption{Normalized Gaussian-volume distributions across five benchmarks. Curves farther left indicate smaller volumes at the same cumulative fraction.}
\label{fig:volume_cdf}
\end{figure}

\paragraph{Quality and rendering speed.}
As shown in Table~\ref{tab:main_results}, SplatQ achieves a macro-average PSNR of 28.426\,dB, exceeding MesonGS and PUP by 0.224\,dB and 0.300\,dB, respectively.
Among the compared pruning methods, it achieves the highest PSNR on four of the five datasets (MesonGS leads on Tanks and Temples), as well as the highest SSIM and lowest LPIPS-VGG on all five.
The lower entry in each quality cell reports the picture-free result.
Under this setting, SplatQ achieves macro-average PSNR, SSIM, and LPIPS-VGG values of 28.250\,dB, 0.88216, and 0.17936, respectively, with its macro-average PSNR exceeding MesonGS by 0.238\,dB.
SplatQ also achieves the highest rendering FPS among the pruning methods on all five datasets.
Figure~\ref{fig:volume_cdf} compares the Gaussian volume distributions of the nine pruning methods.
At matched cumulative fractions, SplatQ generally yields smaller normalized Gaussian volumes.
Details of volume normalization and cross-scene aggregation are provided in Appendix~\ref{app:metric_methods}.
Prior work shows that, for a fixed number of Gaussians, rendering cost is related to the number of screen-space pixels processed per Gaussian~\citep{liu2026shortersplatting,hanson2025speedysplat,feng2025flashgs}.
Thus, the smaller Gaussian volumes in SplatQ may correspond to more localized screen-space footprints and hence a lower rendering workload, providing a possible explanation for the observed FPS gains. We attribute the smaller Gaussian sizes to our pruning's ability to maintain better spatial relationships among the Gaussians, which reduces the need for compensatory enlargement during fine-tuning.
We report direct rendering metrics in Appendix Table~\ref{tab:fps-workload}, including the blending workload, average number of covering layers per pixel, and actual number of blending operations per pixel. Overall, SplatQ performs favorably against the competing methods on these metrics, consistent with our analysis of rendering efficiency.

\begin{figure}[H]
\centering
\includegraphics[width=\linewidth]{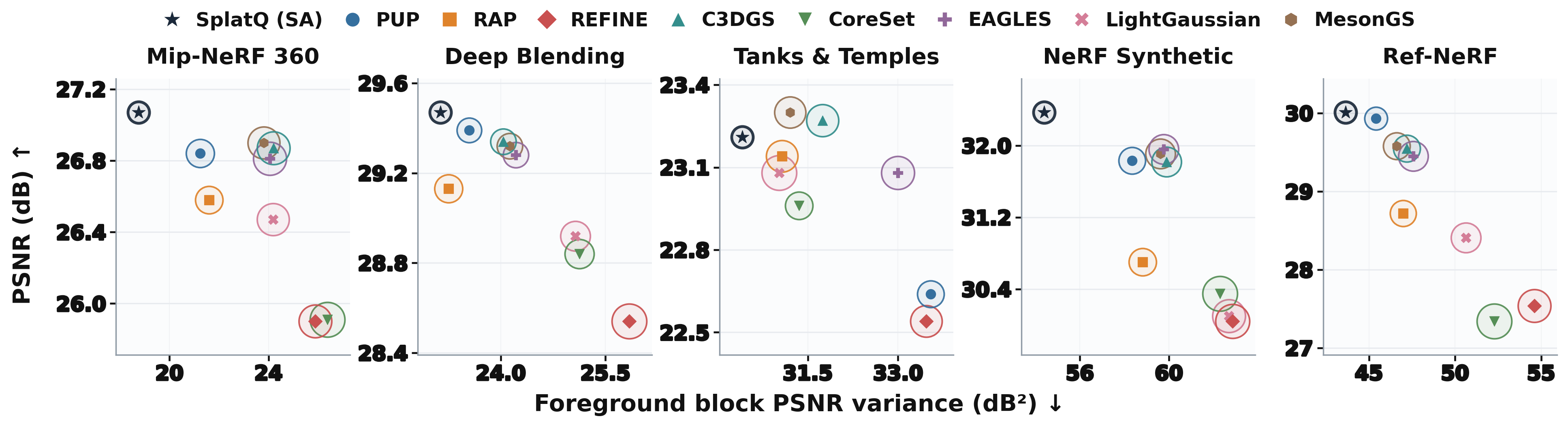}
\caption{Overall reconstruction quality and spatial quality variation across five benchmarks. The horizontal axis shows the variance of PSNR over 256 image patches, and the vertical axis shows full-image PSNR. Bubble area represents the patch-PSNR range and is comparable only within each dataset. Higher positions, smaller horizontal values, and smaller bubbles indicate better results.The numerical values used to generate the figure are reported in Table~\ref{tab:spatial-quality-balance}.}
\label{fig:spatial_quality_balance}
\end{figure}

\paragraph{Spatial quality balance.}
To evaluate how effectively SplatQ alleviates regional imbalance in reconstruction quality, Figure~\ref{fig:spatial_quality_balance} compares SplatQ (SA) with eight pruning baselines across five benchmarks.
Alongside full-image PSNR, we use the variance and range of patch PSNR introduced in Section~4.1 to measure spatial variation in reconstruction quality.
SplatQ (SA) achieves the lowest patch-PSNR variance and range on all five benchmarks, while attaining the highest full-image PSNR on four benchmarks and remaining close to the best result on Tanks and Temples.
These results indicate that SplatQ reduces quality differences across regions while maintaining high overall reconstruction quality, thereby alleviating regional quality imbalance.

\paragraph{Sensitivity to pruning and fine-tuning.}
The pruning-ratio study evaluates removal ratios from 10\% to 90\% in increments of 10 percentage points, together with seven fine-tuning budgets ranging from 0 to 30k steps.
The pruning-round study evaluates one to four rounds with 5k--30k fine-tuning steps per round, at a final removal ratio of 90\%.
The main paper presents the 2D slice of pruning-ratio results at 10k fine-tuning steps and the full 3D sweep over pruning rounds and fine-tuning steps (Figure~\ref{fig:pruning_ablations}).We also computed BD-rate in Appendix~\ref{sec:bd_rate}.
SplatQ outperforms the compared open-source pruning methods in reconstruction quality under the vast majority of evaluated configurations, demonstrating that its advantages extend across different pruning ratios, numbers of pruning rounds, and fine-tuning budgets.
Under a few settings on Tanks and Temples, some quality metrics of SplatQ are slightly below those of the best-performing method.
This suggests room for further improvement on datasets containing distant scene regions.
Notably, all methods retain the same number of Gaussians and use the same uncompressed attribute representation, resulting in the same storage size.

\begin{table}[!t]
\centering
\caption{Main and picture-free comparisons at 90\% pruning.  Best among pruned methods is \textbf{bold} and second best is \underline{underlined}.}
\label{tab:main_results}
\fontsize{7.5}{8.7}\selectfont
\setlength{\tabcolsep}{2.0pt}\renewcommand{\arraystretch}{0.90}
\newcommand{\paircell}[2]{\shortstack[r]{#1\\[-.15ex]#2}}
\newcommand{\pairmethod}[1]{\shortstack[l]{#1\\[-.15ex]\strut}}
\begin{tabular*}{\linewidth}{@{\extracolsep{\fill}}l*{12}{r}@{}}
\toprule
Method & \multicolumn{4}{c}{Mip-NeRF 360} & \multicolumn{4}{c}{Deep Blending} & \multicolumn{4}{c}{Tanks and Temples}\\
\cmidrule(lr){2-5}
\cmidrule(lr){6-9}
\cmidrule(lr){10-13}
 & PSNR$\uparrow$ & SSIM$\uparrow$ & LPIPS$\downarrow$ & FPS$\uparrow$ & PSNR$\uparrow$ & SSIM$\uparrow$ & LPIPS$\downarrow$ & FPS$\uparrow$ & PSNR$\uparrow$ & SSIM$\uparrow$ & LPIPS$\downarrow$ & FPS$\uparrow$\\
\midrule
3DGS & 27.47 & 0.8123 & 0.2216 & 214.73 & 29.43 & 0.8984 & 0.2460 & 216.03 & 23.36 & 0.8378 & 0.1860 & 285.66 \\
\midrule
\pairmethod{LightGaussian} & \paircell{26.47}{26.17} & \paircell{0.7711}{0.7671} & \paircell{0.2950}{0.2994} & \paircell{600.25}{\strut} & \paircell{28.92}{28.99} & \paircell{0.8918}{0.8926} & \paircell{0.2840}{0.2828} & \paircell{792.80}{\strut} & \paircell{23.08}{22.80} & \paircell{0.7992}{0.7973} & \paircell{0.2640}{0.2624} & \paircell{1063.42}{\strut}\\
\pairmethod{MesonGS} & \paircell{\underline{26.90}}{\underline{26.54}} & \paircell{\underline{0.7903}}{\underline{0.7841}} & \paircell{0.2665}{0.2755} & \paircell{716.36}{\strut} & \paircell{29.32}{29.30} & \paircell{0.8979}{0.8977} & \paircell{0.2644}{0.2662} & \paircell{973.20}{\strut} & \paircell{\textbf{23.30}}{\underline{22.92}} & \paircell{\underline{0.8154}}{\underline{0.8113}} & \paircell{0.2371}{0.2404} & \paircell{1286.88}{\strut}\\
\pairmethod{EAGLES} & \paircell{26.81}{26.43} & \paircell{\underline{0.7903}}{\underline{0.7841}} & \paircell{\underline{0.2663}}{\underline{0.2753}} & \paircell{707.75}{\strut} & \paircell{29.28}{29.29} & \paircell{0.8979}{0.8977} & \paircell{0.2642}{0.2664} & \paircell{1006.56}{\strut} & \paircell{23.08}{22.77} & \paircell{0.8107}{0.8068} & \paircell{0.2405}{0.2441} & \paircell{1254.02}{\strut}\\
\pairmethod{C3DGS} & \paircell{26.87}{26.47} & \paircell{0.7889}{0.7817} & \paircell{0.2698}{0.2797} & \paircell{747.66}{\strut} & \paircell{29.34}{29.34} & \paircell{0.8984}{0.8982} & \paircell{0.2640}{0.2658} & \paircell{1005.52}{\strut} & \paircell{\underline{23.27}}{22.90} & \paircell{0.8143}{0.8104} & \paircell{\underline{0.2369}}{\underline{0.2401}} & \paircell{1299.13}{\strut}\\
\pairmethod{PUP} & \paircell{26.84}{26.39} & \paircell{0.7883}{0.7837} & \paircell{0.2675}{0.2754} & \paircell{\underline{896.26}}{\strut} & \paircell{\underline{29.39}}{\underline{29.37}} & \paircell{\underline{0.9017}}{\underline{0.9022}} & \paircell{\underline{0.2588}}{\underline{0.2572}} & \paircell{\underline{1206.42}}{\strut} & \paircell{22.64}{22.39} & \paircell{0.8038}{0.7988} & \paircell{0.2416}{0.2466} & \paircell{\underline{1452.20}}{\strut}\\
\pairmethod{RAP} & \paircell{26.58}{26.05} & \paircell{0.7706}{0.7591} & \paircell{0.3026}{0.3163} & \paircell{781.16}{\strut} & \paircell{29.13}{29.00} & \paircell{0.8910}{0.8883} & \paircell{0.2867}{0.2931} & \paircell{1071.71}{\strut} & \paircell{23.14}{22.70} & \paircell{0.7969}{0.7872} & \paircell{0.2674}{0.2784} & \paircell{1392.78}{\strut}\\
\pairmethod{REFINE} & \paircell{25.90}{25.42} & \paircell{0.7536}{0.7416} & \paircell{0.3188}{0.3325} & \paircell{801.46}{\strut} & \paircell{28.54}{28.35} & \paircell{0.8789}{0.8762} & \paircell{0.3003}{0.3067} & \paircell{1078.79}{\strut} & \paircell{22.54}{22.13} & \paircell{0.7712}{0.7616} & \paircell{0.2878}{0.2982} & \paircell{1307.51}{\strut}\\
\pairmethod{CoreSet} & \paircell{25.91}{25.50} & \paircell{0.7705}{0.7608} & \paircell{0.2903}{0.3044} & \paircell{771.27}{\strut} & \paircell{28.84}{28.84} & \paircell{0.8882}{0.8878} & \paircell{0.2812}{0.2848} & \paircell{1005.39}{\strut} & \paircell{22.96}{22.58} & \paircell{0.7983}{0.7906} & \paircell{0.2569}{0.2649} & \paircell{1344.95}{\strut}\\
\pairmethod{\textbf{SplatQ (SA)}} & \paircell{\textbf{27.07}}{\textbf{26.69}} & \paircell{\textbf{0.7932}}{\textbf{0.7870}} & \paircell{\textbf{0.2547}}{\textbf{0.2721}} & \paircell{\textbf{1022.23}}{\strut} & \paircell{\textbf{29.47}}{\textbf{29.43}} & \paircell{\textbf{0.9027}}{\textbf{0.9028}} & \paircell{\textbf{0.2562}}{\textbf{0.2571}} & \paircell{\textbf{1232.28}}{\strut} & \paircell{23.21}{\textbf{22.93}} & \paircell{\textbf{0.8160}}{\textbf{0.8136}} & \paircell{\textbf{0.2322}}{\textbf{0.2349}} & \paircell{\textbf{1573.73}}{\strut}\\
\bottomrule
\end{tabular*}
\par\vspace{5pt}
\begin{tabular*}{\linewidth}{@{\extracolsep{\fill}}l*{11}{r}@{}}
\toprule
Method & \multicolumn{4}{c}{NeRF Synthetic} & \multicolumn{4}{c}{Ref-NeRF} & \multicolumn{3}{c}{Macro}\\
\cmidrule(lr){2-5}
\cmidrule(lr){6-9}
\cmidrule(lr){10-12}
 & PSNR$\uparrow$ & SSIM$\uparrow$ & LPIPS$\downarrow$ & FPS$\uparrow$ & PSNR$\uparrow$ & SSIM$\uparrow$ & LPIPS$\downarrow$ & FPS$\uparrow$ & PSNR$\uparrow$ & SSIM$\uparrow$ & LPIPS$\downarrow$\\
\midrule
3DGS & 33.32 & 0.9695 & 0.0304 & 743.34 & 30.36 & 0.9470 & 0.0837 & 919.88 & 28.788 & 0.89300 & 0.15354\\
\midrule
\pairmethod{LightGaussian} & \paircell{30.10}{30.18} & \paircell{0.9486}{0.9494} & \paircell{0.0624}{0.0613} & \paircell{2668.98}{\strut} & \paircell{28.41}{28.47} & \paircell{0.9363}{0.9369} & \paircell{0.1063}{0.1053} & \paircell{2392.75}{\strut} & \paircell{27.396}{27.322} & \paircell{0.86940}{0.86866} & \paircell{0.20234}{0.20224}\\
\pairmethod{MesonGS} & \paircell{31.91}{31.84} & \paircell{0.9614}{0.9613} & \paircell{0.0444}{\underline{0.0445}} & \paircell{3388.19}{\strut} & \paircell{29.58}{29.46} & \paircell{0.9430}{0.9430} & \paircell{0.0934}{0.0937} & \paircell{3002.43}{\strut} & \paircell{\underline{28.202}}{\underline{28.012}} & \paircell{\underline{0.88160}}{\underline{0.87948}} & \paircell{0.18116}{0.18406}\\
\pairmethod{EAGLES} & \paircell{\underline{31.96}}{\underline{31.89}} & \paircell{\underline{0.9617}}{\underline{0.9615}} & \paircell{\underline{0.0440}}{\underline{0.0445}} & \paircell{3377.08}{\strut} & \paircell{29.45}{29.41} & \paircell{0.9432}{0.9432} & \paircell{0.0927}{\underline{0.0932}} & \paircell{\underline{3060.99}}{\strut} & \paircell{28.116}{27.958} & \paircell{0.88076}{0.87866} & \paircell{0.18154}{0.18470}\\
\pairmethod{C3DGS} & \paircell{31.82}{31.75} & \paircell{0.9610}{0.9609} & \paircell{0.0449}{0.0452} & \paircell{3372.90}{\strut} & \paircell{29.55}{29.52} & \paircell{0.9430}{0.9431} & \paircell{0.0931}{0.0934} & \paircell{2984.86}{\strut} & \paircell{28.170}{27.996} & \paircell{0.88112}{0.87886} & \paircell{0.18174}{0.18484}\\
\pairmethod{PUP} & \paircell{31.83}{31.63} & \paircell{0.9607}{0.9595} & \paircell{\underline{0.0440}}{0.0456} & \paircell{\underline{3489.60}}{\strut} & \paircell{\underline{29.93}}{\underline{29.77}} & \paircell{\underline{0.9443}}{\underline{0.9434}} & \paircell{\underline{0.0915}}{0.0937} & \paircell{3026.85}{\strut} & \paircell{28.126}{27.910} & \paircell{0.87976}{0.87752} & \paircell{\underline{0.18068}}{\underline{0.18370}}\\
\pairmethod{RAP} & \paircell{30.70}{30.26} & \paircell{0.9506}{0.9476} & \paircell{0.0615}{0.0653} & \paircell{3346.09}{\strut} & \paircell{28.72}{28.40} & \paircell{0.9360}{0.9340} & \paircell{0.1067}{0.1098} & \paircell{2740.48}{\strut} & \paircell{27.654}{27.282} & \paircell{0.86902}{0.86324} & \paircell{0.20498}{0.21258}\\
\pairmethod{REFINE} & \paircell{30.04}{29.65} & \paircell{0.9469}{0.9439} & \paircell{0.0640}{0.0682} & \paircell{2985.89}{\strut} & \paircell{27.54}{27.07} & \paircell{0.9294}{0.9263} & \paircell{0.1194}{0.1247} & \paircell{2623.77}{\strut} & \paircell{26.912}{26.524} & \paircell{0.85600}{0.84992} & \paircell{0.21806}{0.22606}\\
\pairmethod{CoreSet} & \paircell{30.35}{29.82} & \paircell{0.9517}{0.9483} & \paircell{0.0562}{0.0603} & \paircell{3294.38}{\strut} & \paircell{27.34}{26.89} & \paircell{0.9270}{0.9249} & \paircell{0.1144}{0.1178} & \paircell{2861.67}{\strut} & \paircell{27.080}{26.726} & \paircell{0.86714}{0.86248} & \paircell{0.19980}{0.20644}\\
\pairmethod{\textbf{SplatQ (SA)}} & \paircell{\textbf{32.37}}{\textbf{32.30}} & \paircell{\textbf{0.9623}}{\textbf{0.9623}} & \paircell{\textbf{0.0424}}{\textbf{0.0426}} & \paircell{\textbf{3595.66}}{\strut} & \paircell{\textbf{30.01}}{\textbf{29.90}} & \paircell{\textbf{0.9453}}{\textbf{0.9451}} & \paircell{\textbf{0.0897}}{\textbf{0.0901}} & \paircell{\textbf{3180.37}}{\strut} & \paircell{\textbf{28.426}}{\textbf{28.250}} & \paircell{\textbf{0.88390}}{\textbf{0.88216}} & \paircell{\textbf{0.17504}}{\textbf{0.17936}}\\
\bottomrule
\end{tabular*}
\end{table}

\begin{figure}[ht]
    \centering
    \vspace{-16pt}
    \includegraphics[width=\linewidth]{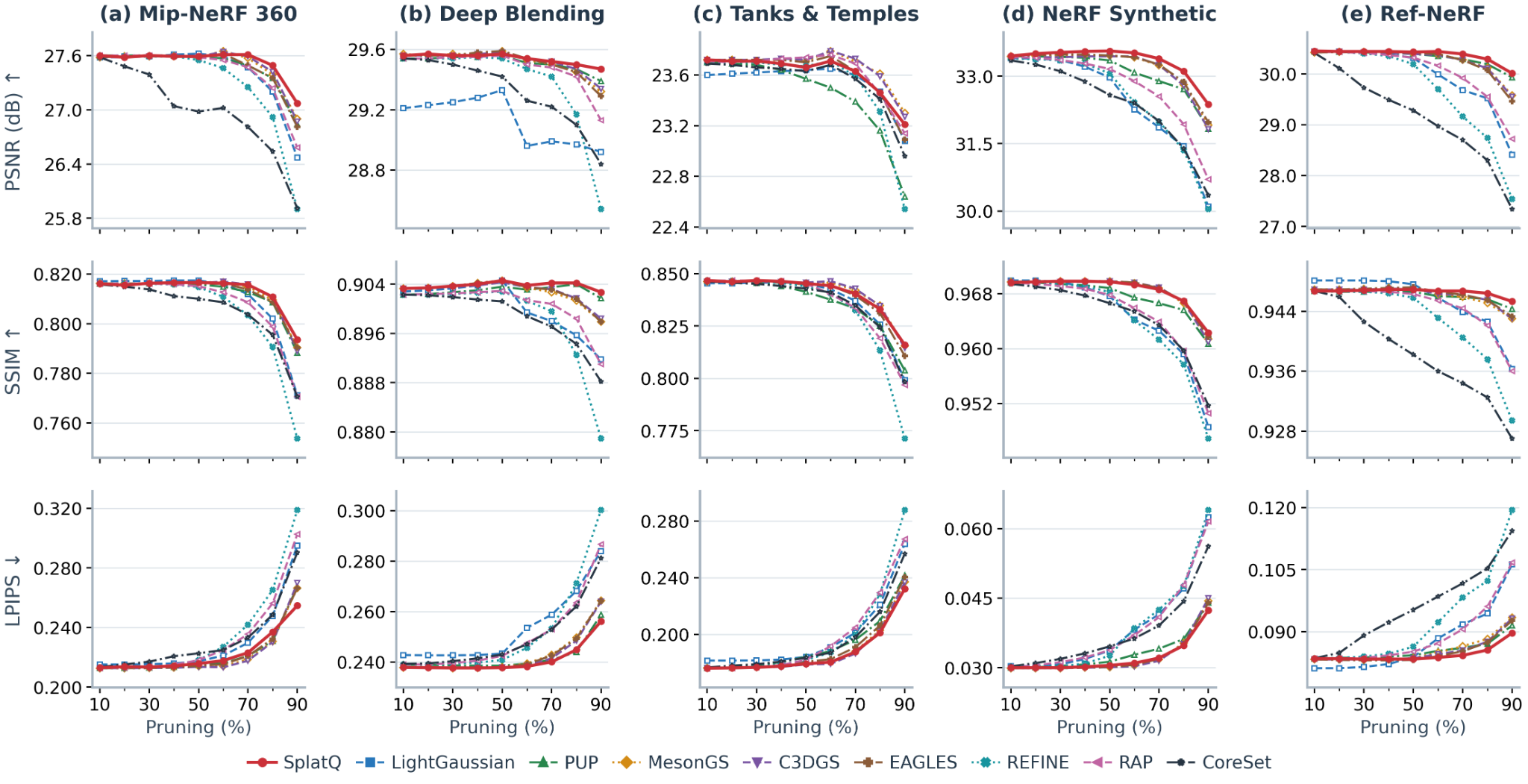}
    \par\vspace{-2pt}
    \includegraphics[width=\linewidth]{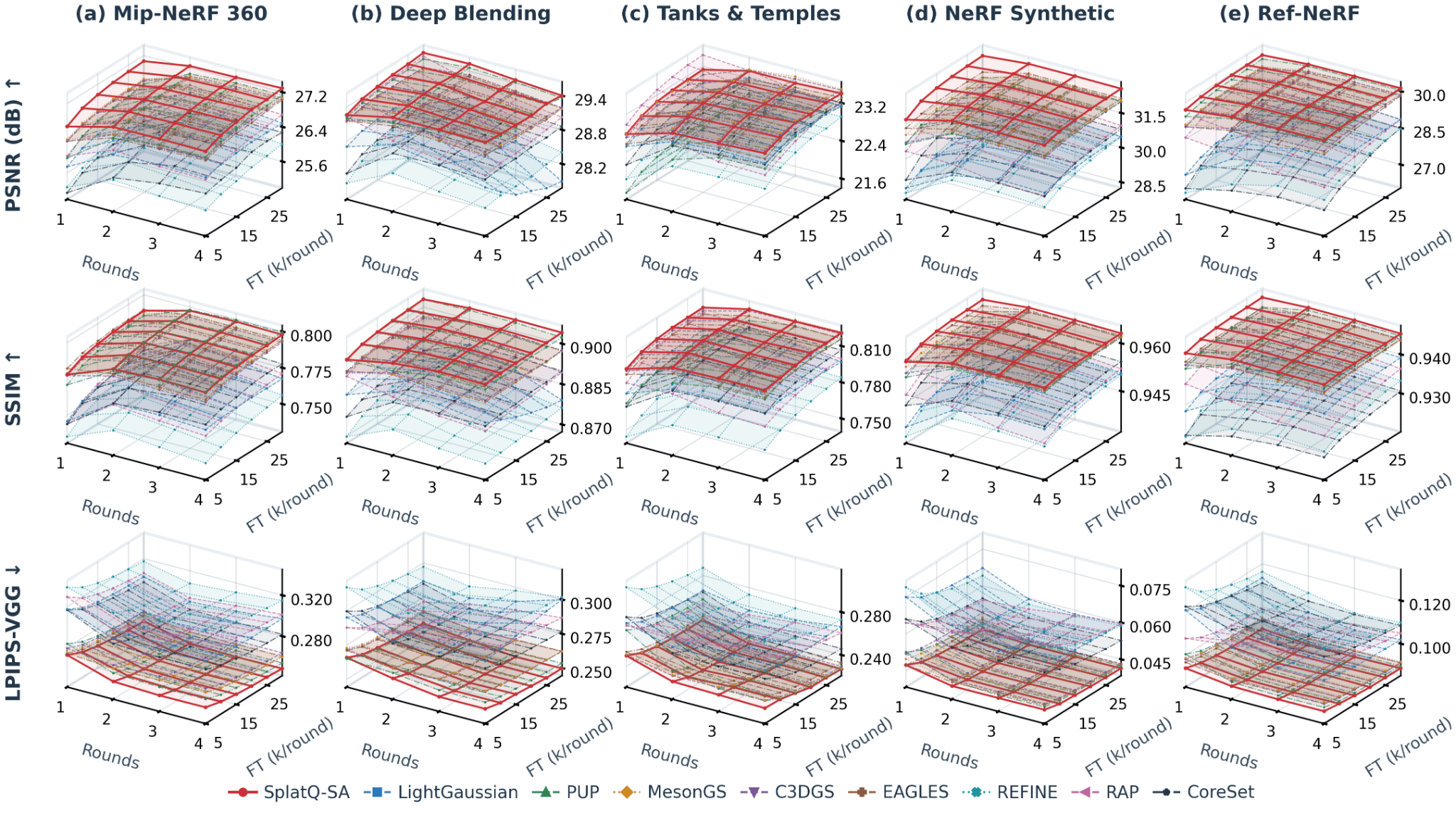}
    \caption{Top: 2D pruning-percentage figure at 10k fine-tuning iterations. Bottom: Pruning-round 3D figure at 90\% final removal. SplatQ denotes the SA variant.}
     \vspace{-5pt}
    \label{fig:pruning_ablations}
\end{figure}

\paragraph{Qualitative Comparison}

Based on the results in Table~\ref{tab:main_results}, we select PUP, MesonGS, and EAGLES as three strong baselines for qualitative comparison.
Figure~\ref{fig:teaser}(a) compares representative test views from SplatQ and these baselines.
In these examples, SplatQ preserves detail more consistently across the highlighted regions, while several baselines show uneven degradation.
\subsection{Ablation Experiments}
\vspace{-5pt}

\begin{table}[!t]
\centering
\caption{Component and solver ablations at 90\% pruning. Boldface marks SplatQ's default config.}
\label{tab:ablation_results_mean}
\fontsize{8.0}{9.2}\selectfont
\setlength{\tabcolsep}{0.5pt}\renewcommand{\arraystretch}{0.9}
\begin{tabular*}{\linewidth}{@{\extracolsep{\fill}}l*{9}{r}@{}}
\toprule
Variant & \multicolumn{3}{c}{Mip-NeRF 360} & \multicolumn{3}{c}{Deep Blending} & \multicolumn{3}{c}{Tanks and Temples}\\
\cmidrule(lr){2-4}\cmidrule(lr){5-7}\cmidrule(lr){8-10}
 & PSNR$\uparrow$ & SSIM$\uparrow$ & LPIPS$\downarrow$ & PSNR$\uparrow$ & SSIM$\uparrow$ & LPIPS$\downarrow$ & PSNR$\uparrow$ & SSIM$\uparrow$ & LPIPS$\downarrow$\\
\midrule
Top-$k$ & 26.68 & 0.7787 & 0.2640 & 29.09 & 0.8944 & 0.2649 & 22.72 & 0.8034 & 0.2469\\
+ Uniform block quotas & 26.52 & 0.7755 & 0.2685 & 28.98 & 0.8930 & 0.2668 & 22.64 & 0.8004 & 0.2505\\
+ Budget assignment & 26.74 & 0.7889 & 0.2626 & 29.15 & 0.8958 & 0.2638 & 22.81 & 0.8050 & 0.2455\\
+ Quadratic, greedy & 27.03 & 0.7928 & 0.2552 & 29.42 & 0.9020 & 0.2568 & 23.15 & 0.8135 & 0.2357\\
Quadratic, uniform quotas & 26.79 & 0.7811 & 0.2602 & 29.36 & 0.8985 & 0.2585 & 23.07 & 0.8093 & 0.2424\\
\textbf{+ Quadratic, SA} & 27.07 & 0.7932 & 0.2547 & 29.47 & 0.9027 & 0.2562 & 23.21 & 0.8160 & 0.2322\\
+ Quadratic, CIM & 27.14 & 0.7960 & 0.2507 & 29.54 & 0.9041 & 0.2532 & 23.21 & 0.8173 & 0.2296\\
\bottomrule
\end{tabular*}
\par\smallskip
\begin{tabular*}{\linewidth}{@{\extracolsep{\fill}}l*{9}{r}@{}}
\toprule
Variant & \multicolumn{3}{c}{NeRF Synthetic} & \multicolumn{3}{c}{Ref-NeRF} & \multicolumn{3}{c}{Macro}\\
\cmidrule(lr){2-4}\cmidrule(lr){5-7}\cmidrule(lr){8-10}
 & PSNR$\uparrow$ & SSIM$\uparrow$ & LPIPS$\downarrow$ & PSNR$\uparrow$ & SSIM$\uparrow$ & LPIPS$\downarrow$ & PSNR$\uparrow$ & SSIM$\uparrow$ & LPIPS$\downarrow$\\
\midrule
Top-$k$ & 31.85 & 0.9568 & 0.0503 & 29.55 & 0.9407 & 0.0972 & 27.978 & 0.87480 & 0.18466\\
+ Uniform block quotas & 31.72 & 0.9557 & 0.0520 & 29.43 & 0.9396 & 0.0993 & 27.858 & 0.87284 & 0.18742\\
+ Budget assignment & 31.96 & 0.9575 & 0.0496 & 29.64 & 0.9415 & 0.0961 & 28.060 & 0.87774 & 0.18352\\
+ Quadratic, greedy & 32.31 & 0.9618 & 0.0427 & 29.96 & 0.9449 & 0.0906 & 28.374 & 0.88300 & 0.17620\\
Quadratic, uniform quotas & 32.14 & 0.9596 & 0.0452 & 29.76 & 0.9431 & 0.0924 & 28.224 & 0.87832 & 0.17974\\
\textbf{+ Quadratic, SA} & 32.37 & 0.9623 & 0.0424 & 30.01 & 0.9453 & 0.0897 & 28.426 & 0.88390 & 0.17504\\
+ Quadratic, CIM & 32.48 & 0.9635 & 0.0409 & 30.07 & 0.9456 & 0.0887 & 28.488 & 0.88530 & 0.17262\\
\bottomrule
\end{tabular*}
\end{table}

\paragraph{Component and solver ablations.}
Table~\ref{tab:ablation_results_mean} compares seven variants across datasets and reports their equally weighted macro-averages.
Uniform block quotas lower macro PSNR from 27.978 to 27.858\,dB relative to global first-order top-$k$; budget allocation raises it to 28.060\,dB.
Adding quadratic interactions with greedy selection yields 28.374\,dB, versus 28.224\,dB without budget allocation.
The main SA variant achieves 28.426\,dB, while CIM, reported separately, reaches 28.488\,dB.
Appendix~\ref{app:additional_experiment:cost} compares CIM, SA, and greedy runtimes.

\paragraph{Qualitative ablations.}
Figure~\ref{fig:teaser}(b) illustrates the corresponding changes in local reconstruction quality.
Uniform block quotas alone do not consistently recover the highlighted details.
As budget allocation and quadratic selection with SA are progressively introduced, reconstruction quality improves in these examples.

\subsection{Composability Experiments}
\label{sec:composability-experiments}

We apply four attribute compressors to 90\%-pruned models from our method,
PUP, RAP, REFINE, and each compressor's own pruning method. Evaluating reloaded compressed models, our inputs
achieve the highest PSNR in 18 of 20 compressor--dataset combinations
(Table~\ref{tab:composability-quality}), with an additional
$3.14$--$13.89\times$ storage reduction on Mip-NeRF 360
(Table~\ref{tab:composability-performance}). These results support combining
our pruning with attribute compression. FPS measures decoded models with a
common renderer and excludes decoding times. Complete storage and runtime results are provided in Appendix~\ref{app:composability}.
\vspace{-3pt}
\newcommand{\ComposabilityQualityTable}{%
\begin{table}[t]
\vspace*{-3mm}
\centering
\caption{Attribute compression at 90\% pruning. Best and second-best displayed quality values within each compressor are \textbf{bold} and \underline{underlined}.}
\label{tab:composability-quality}
\begingroup
\setlength{\tabcolsep}{1.5pt}
\renewcommand{\arraystretch}{1.08}
\resizebox{\linewidth}{!}{%
\begin{tabular}{@{}ll*{15}{r}@{}}
\toprule
\multirow{2}{*}{Method} & \multirow{2}{*}{Input} & \multicolumn{3}{c}{Mip-NeRF 360} & \multicolumn{3}{c}{Deep Blending} & \multicolumn{3}{c}{Tanks \& Temples} & \multicolumn{3}{c}{NeRF Synthetic} & \multicolumn{3}{c}{Ref-NeRF} \\
\cmidrule(lr){3-5}\cmidrule(lr){6-8}\cmidrule(lr){9-11}\cmidrule(lr){12-14}\cmidrule(lr){15-17}
 & & PSNR$\uparrow$ & SSIM$\uparrow$ & LPIPS$\downarrow$ & PSNR$\uparrow$ & SSIM$\uparrow$ & LPIPS$\downarrow$ & PSNR$\uparrow$ & SSIM$\uparrow$ & LPIPS$\downarrow$ & PSNR$\uparrow$ & SSIM$\uparrow$ & LPIPS$\downarrow$ & PSNR$\uparrow$ & SSIM$\uparrow$ & LPIPS$\downarrow$ \\
\midrule
\multirow{5}{*}{LightGS} & Ours & \textbf{26.70} & \textbf{0.786} & \textbf{0.273} & \textbf{29.26} & \textbf{0.900} & \underline{0.262} & \textbf{22.89} & \textbf{0.808} & \textbf{0.242} & \textbf{31.59} & \textbf{0.959} & \textbf{0.047} & \textbf{29.36} & \textbf{0.942} & \textbf{0.096} \\
 & PUP & \underline{26.47} & \underline{0.784} & \underline{0.282} & \underline{29.14} & \textbf{0.900} & \textbf{0.260} & 22.38 & \underline{0.795} & \underline{0.253} & \underline{31.08} & \underline{0.956} & \underline{0.049} & \underline{29.24} & \underline{0.941} & \underline{0.099} \\
 & RAP & 26.18 & 0.761 & 0.314 & 28.85 & \underline{0.888} & 0.293 & \underline{22.80} & 0.786 & 0.279 & 30.10 & 0.947 & 0.066 & 28.16 & 0.932 & 0.113 \\
 & REFINE & 25.64 & 0.745 & 0.329 & 28.33 & 0.877 & 0.305 & 22.28 & 0.763 & 0.298 & 29.51 & 0.943 & 0.069 & 27.17 & 0.926 & 0.125 \\
 & LightGS & 26.00 & 0.758 & 0.309 & 28.53 & 0.884 & 0.294 & 22.72 & 0.786 & 0.279 & 29.51 & 0.943 & 0.068 & 27.85 & 0.932 & 0.114 \\
\midrule
\multirow{5}{*}{MesonGS} & Ours & \textbf{26.59} & \textbf{0.781} & \textbf{0.277} & \textbf{29.29} & \underline{0.900} & \textbf{0.260} & \underline{22.97} & \textbf{0.808} & \textbf{0.242} & \textbf{31.49} & \textbf{0.958} & \textbf{0.047} & \textbf{29.81} & \textbf{0.944} & \textbf{0.092} \\
 & PUP & 26.39 & \underline{0.780} & 0.287 & \underline{29.23} & \textbf{0.901} & \underline{0.263} & 22.41 & 0.795 & 0.252 & 31.14 & \underline{0.956} & \underline{0.048} & \underline{29.75} & \underline{0.943} & \underline{0.094} \\
 & RAP & 26.13 & 0.757 & 0.317 & 28.94 & 0.887 & 0.294 & 22.90 & 0.787 & 0.277 & 30.00 & 0.946 & 0.065 & 28.32 & 0.934 & 0.109 \\
 & REFINE & 25.47 & 0.739 & 0.334 & 28.34 & 0.875 & 0.308 & 22.33 & 0.761 & 0.298 & 29.38 & 0.942 & 0.067 & 27.16 & 0.927 & 0.121 \\
 & MesonGS & \underline{26.49} & 0.778 & \underline{0.281} & 29.16 & 0.895 & 0.271 & \textbf{23.06} & \underline{0.806} & \underline{0.247} & \underline{31.27} & \textbf{0.958} & \underline{0.048} & 29.38 & 0.942 & 0.096 \\
\midrule
\multirow{5}{*}{EAGLES} & Ours & \textbf{26.40} & \textbf{0.784} & \textbf{0.273} & \textbf{29.42} & \textbf{0.898} & \textbf{0.273} & \textbf{22.70} & \textbf{0.804} & \textbf{0.247} & \textbf{31.18} & \textbf{0.958} & \textbf{0.049} & \textbf{29.32} & \textbf{0.942} & \textbf{0.098} \\
 & PUP & 26.16 & 0.772 & 0.300 & \underline{29.39} & \underline{0.897} & 0.282 & 21.96 & 0.779 & 0.274 & 30.91 & \underline{0.957} & \textbf{0.049} & \underline{29.31} & \textbf{0.942} & \underline{0.099} \\
 & RAP & 25.82 & 0.754 & 0.324 & 29.38 & 0.893 & 0.290 & 22.14 & 0.769 & 0.303 & 30.21 & 0.950 & \underline{0.061} & 28.69 & \underline{0.937} & 0.109 \\
 & REFINE & 25.59 & 0.745 & 0.328 & 28.53 & 0.875 & 0.313 & 21.70 & 0.753 & 0.312 & 29.88 & 0.948 & 0.062 & 27.94 & 0.932 & 0.116 \\
 & EAGLES & \underline{26.19} & \underline{0.776} & \underline{0.282} & 29.24 & 0.895 & \underline{0.278} & \underline{22.37} & \underline{0.788} & \underline{0.268} & \underline{30.97} & \textbf{0.958} & \textbf{0.049} & 29.16 & \textbf{0.942} & 0.100 \\
\midrule
\multirow{5}{*}{C3DGS} & Ours & \textbf{26.69} & \textbf{0.783} & \textbf{0.276} & \textbf{29.17} & \textbf{0.899} & \underline{0.267} & \underline{22.93} & \textbf{0.809} & \textbf{0.240} & \textbf{32.26} & \textbf{0.962} & \textbf{0.043} & \textbf{29.94} & \textbf{0.945} & \textbf{0.090} \\
 & PUP & 26.48 & \textbf{0.783} & 0.285 & \underline{29.16} & \underline{0.898} & \textbf{0.263} & 22.43 & 0.796 & 0.250 & 31.78 & \underline{0.961} & \underline{0.044} & \underline{29.88} & \underline{0.944} & \underline{0.091} \\
 & RAP & 26.31 & 0.763 & 0.312 & 28.94 & 0.887 & 0.294 & 22.91 & 0.791 & 0.272 & 30.77 & 0.951 & 0.060 & 28.82 & 0.937 & 0.105 \\
 & REFINE & 25.71 & 0.747 & 0.328 & 28.43 & 0.877 & 0.306 & 22.31 & 0.764 & 0.294 & 30.17 & 0.948 & 0.062 & 27.70 & 0.931 & 0.117 \\
 & C3DGS & \underline{26.57} & \underline{0.780} & \underline{0.280} & 29.13 & 0.895 & 0.273 & \textbf{23.03} & \underline{0.808} & \underline{0.243} & \underline{31.79} & \underline{0.961} & 0.045 & 29.56 & 0.943 & 0.093 \\
\bottomrule
\end{tabular}%
}
\endgroup
\end{table}%
}

\newcommand{\ComposabilityCompactPerformanceTable}{%
\begin{table}[t]
\centering
\caption{Approximate total compression ratio (Rat., attribute ratio $\times 10$), decode time (dec, seconds), and decoded-model FPS. Best and second-best FPS within each compressor are \textbf{bold} and \underline{underlined}.}
\label{tab:composability-performance}
\begingroup
\setlength{\tabcolsep}{1.5pt}
\renewcommand{\arraystretch}{1.08}
\resizebox{\linewidth}{!}{%
\begin{tabular}{@{}ll*{15}{r}@{}}
\toprule
\multirow{2}{*}{Method} & \multirow{2}{*}{Input} & \multicolumn{3}{c}{Mip-NeRF 360} & \multicolumn{3}{c}{Deep Blending} & \multicolumn{3}{c}{Tanks \& Temples} & \multicolumn{3}{c}{NeRF Synthetic} & \multicolumn{3}{c}{Ref-NeRF} \\
\cmidrule(lr){3-5}\cmidrule(lr){6-8}\cmidrule(lr){9-11}\cmidrule(lr){12-14}\cmidrule(lr){15-17}
 & & Rat.$\uparrow$ & dec$\downarrow$ & FPS$\uparrow$ & Rat.$\uparrow$ & dec$\downarrow$ & FPS$\uparrow$ & Rat.$\uparrow$ & dec$\downarrow$ & FPS$\uparrow$ & Rat.$\uparrow$ & dec$\downarrow$ & FPS$\uparrow$ & Rat.$\uparrow$ & dec$\downarrow$ & FPS$\uparrow$ \\
\midrule
\multirow{5}{*}{LightGS} & Ours & 51.0 & 1.59 & \textbf{1039.4} & 51.4 & 1.41 & \textbf{1257.4} & 50.0 & 0.89 & \textbf{1518.8} & 47.3 & 0.21 & \textbf{2936.7} & 45.9 & 0.16 & \textbf{2472.1} \\
 & PUP & 51.2 & 1.58 & \underline{899.7} & 51.5 & 1.45 & \underline{1228.9} & 50.2 & 0.86 & \underline{1362.0} & 47.1 & 0.21 & \underline{2769.5} & 46.3 & 0.15 & \underline{2455.7} \\
 & RAP & 50.9 & 1.64 & 819.7 & 51.3 & 1.49 & 1046.2 & 50.1 & 0.88 & 1317.1 & 47.3 & 0.23 & 2544.9 & 45.9 & 0.15 & 2151.6 \\
 & REFINE & 51.0 & 1.61 & 799.4 & 51.3 & 1.43 & 1045.4 & 50.1 & 0.89 & 1246.8 & 47.3 & 0.21 & 2293.0 & 45.9 & 0.16 & 2077.9 \\
 & LightGS & 51.1 & 1.61 & 594.0 & 51.6 & 1.44 & 784.6 & 50.1 & 0.87 & 1019.4 & 46.9 & 0.23 & 2195.4 & 45.3 & 0.15 & 2085.7 \\
\midrule
\multirow{5}{*}{MesonGS} & Ours & 138.9 & 4.98 & \textbf{987.4} & 136.9 & 4.52 & \textbf{1163.9} & 115.5 & 2.80 & \textbf{1462.8} & 72.2 & 0.75 & \textbf{2993.8} & 39.3 & 0.59 & \textbf{2562.1} \\
 & PUP & 138.7 & 4.90 & \underline{857.6} & 136.9 & 4.70 & \underline{1158.7} & 117.1 & 2.76 & \underline{1355.6} & 72.2 & 0.73 & \underline{2769.7} & 39.3 & 0.56 & \underline{2491.7} \\
 & RAP & 140.1 & 4.98 & 790.1 & 138.4 & 4.58 & 1016.1 & 117.3 & 2.91 & 1312.0 & 72.2 & 0.67 & 2679.6 & 39.3 & 0.53 & 2231.8 \\
 & REFINE & 140.3 & 4.88 & 780.3 & 139.1 & 4.65 & 1008.9 & 118.7 & 2.81 & 1226.0 & 72.3 & 0.67 & 2396.9 & 39.3 & 0.52 & 2155.2 \\
 & MesonGS & 139.8 & 5.00 & 695.5 & 139.4 & 4.58 & 935.1 & 117.3 & 2.86 & 1220.4 & 72.2 & 0.70 & 2590.0 & 39.3 & 0.57 & 2451.8 \\
\midrule
\multirow{5}{*}{EAGLES} & Ours & 31.4 & 9.38 & \textbf{987.0} & 35.0 & 21.29 & \textbf{1330.1} & 29.9 & 16.17 & \textbf{1564.6} & 20.5 & 4.13 & \textbf{2881.6} & 15.9 & 2.40 & \textbf{2482.8} \\
 & PUP & 32.0 & 9.24 & 816.2 & 35.2 & 21.80 & 1169.5 & 31.1 & 16.30 & 1487.7 & 21.7 & 3.69 & \underline{2787.4} & 16.4 & 2.37 & \underline{2464.2} \\
 & RAP & 32.3 & 8.22 & \underline{874.6} & 35.5 & 17.59 & \underline{1207.5} & 31.0 & 12.91 & \underline{1498.9} & 22.3 & 3.11 & 2656.4 & 16.6 & 2.01 & 2307.5 \\
 & REFINE & 32.3 & 8.20 & 814.8 & 35.3 & 17.31 & 1031.7 & 31.2 & 12.91 & 1421.5 & 22.6 & 3.02 & 2451.2 & 17.8 & 2.09 & 2171.1 \\
 & EAGLES & 32.3 & 8.97 & 754.2 & 35.7 & 21.15 & 1090.3 & 31.0 & 16.21 & 1388.6 & 21.9 & 3.71 & 2649.3 & 16.8 & 2.31 & 2421.1 \\
\midrule
\multirow{5}{*}{C3DGS} & Ours & 103.3 & 1.15 & \textbf{1030.9} & 104.0 & 0.96 & \textbf{1279.6} & 79.5 & 0.69 & \textbf{1540.8} & 59.5 & 0.33 & \textbf{2986.0} & 51.5 & 0.30 & \textbf{2641.5} \\
 & PUP & 94.8 & 1.17 & \underline{906.8} & 100.9 & 0.97 & \underline{1261.8} & 72.2 & 0.67 & \underline{1453.2} & 55.4 & 0.33 & \underline{2803.9} & 48.8 & 0.29 & 2583.5 \\
 & RAP & 88.6 & 1.14 & 837.9 & 96.1 & 0.97 & 1112.7 & 72.3 & 0.63 & 1380.2 & 58.1 & 0.31 & 2680.3 & 51.8 & 0.29 & 2291.9 \\
 & REFINE & 98.1 & 1.18 & 823.1 & 102.3 & 0.96 & 1109.2 & 81.1 & 0.66 & 1293.4 & 60.1 & 0.31 & 2456.7 & 53.4 & 0.29 & 2263.8 \\
 & C3DGS & 74.2 & 1.23 & 789.1 & 73.6 & 1.02 & 1062.0 & 59.8 & 0.70 & 1303.3 & 49.0 & 0.33 & 2783.3 & 47.9 & 0.30 & \underline{2636.8} \\
\bottomrule
\end{tabular}%
}
\endgroup
\end{table}%
}

\ComposabilityQualityTable
\ComposabilityCompactPerformanceTable

\section{Conclusion}
\vspace{-3pt}
This work makes three main contributions. First, we propose SplatQ, which combines partitioning based on spatial and appearance similarity, regional budget allocation, and joint Gaussian selection to mitigate regional quality imbalance under a fixed budget and account for joint deletion effects overlooked by independent scoring. We also introduce metrics for assessing regional quality imbalance. Second, we construct a QUBO model from a quadratic approximation of joint deletion error and a retention-count constraint, and demonstrate its solvability through experiments on CIM quantum computing hardware. Finally, in comparisons against 8 baselines across 5 benchmarks, SplatQ achieves the best overall reconstruction metrics and rendering speed at 90\% pruning. We further evaluate different pruning ratios, numbers of pruning rounds, and fine-tuning budgets to facilitate comparisons with methods without released code or those that do not enforce a fixed retention budget.

\subsection*{AI use statement}
Generative AI tools assisted with manuscript drafting, language editing
and translation,
figure and table preparation, and LaTeX formatting.
All reported numerical results originate from actual experimental
records, not AI-generated data.
The authors take responsibility for the final content and accuracy
of the paper.

\subsection*{Ethics statement}

Our experiments use only existing public benchmark datasets.
Potential downstream concerns include unauthorized sharing of
reconstructed private spaces and loss of important visual details
under aggressive pruning.
Neither compression nor the picture-free setting provides a privacy
guarantee.
Use and redistribution of scene assets should respect applicable
licenses and privacy requirements.

\subsection*{Reproducibility statement}

Section~\ref{sec:method} and Appendix~A describe the method,
derivations, pseudocode, and solver postprocessing.
Appendix~B.1 specifies the datasets, baseline adaptations,
hyperparameters, random seeds, fine-tuning and evaluation protocols,
and hardware.
Table~2 reports means and standard deviations across five
complete-pipeline runs.
Appendix~B.2 provides computational costs and additional ablations.
We plan to publicly release the source code.

\FloatBarrier
\bibliography{references}

@inproceedings{liu2026shortersplatting,
  title     = {Speeding Up the Learning of {3D} Gaussians with Much Shorter Gaussian Lists},
  author    = {Liu, Jiaqi and Han, Zhizhong},
  booktitle = {CVPR},
  year      = {2026}
}

@inproceedings{hanson2025speedysplat,
  title     = {{Speedy-Splat}: Fast {3D Gaussian Splatting} with Sparse Pixels and Sparse Primitives},
  author    = {Hanson, Alex and Tu, Allen and Lin, Geng and
               Singla, Vasu and Zwicker, Matthias and Goldstein, Tom},
  booktitle = {CVPR},
  pages     = {21537--21546},
  year      = {2025}
}

@inproceedings{feng2025flashgs,
  title     = {{FlashGS}: Efficient {3D Gaussian Splatting} for Large-scale and High-resolution Rendering},
  author    = {Feng, Guofeng and Chen, Siyan and Fu, Rong and
               Liao, Zimu and Wang, Yi and Liu, Tao and
               Hu, Boni and Xu, Linning and Pei, Zhilin and
               Li, Hengjie and Li, Xiuhong and Sun, Ninghui and
               Zhang, Xingcheng and Dai, Bo},
  booktitle = {CVPR},
  pages     = {26652--26662},
  year      = {2025}
}

@article{kerbl2023_3dgs,
  author = {Bernhard Kerbl and Georgios Kopanas and Thomas Leimk{\"u}hler and George Drettakis},
  title = {{3D Gaussian Splatting} for Real-Time Radiance Field Rendering},
  journal = {{ACM TOG}},
  volume = {42},
  number = {4},
  year = {2023},
  doi = {10.1145/3592433},
  url = {https://repo-sam.inria.fr/fungraph/3d-gaussian-splatting/}
}

@inproceedings{barron2022mipnerf360,
  author = {Jonathan T. Barron and Ben Mildenhall and Dor Verbin and Pratul P. Srinivasan and Peter Hedman},
  title = {{Mip-NeRF 360}: Unbounded Anti-Aliased Neural Radiance Fields},
  booktitle = {{CVPR}},
  pages     = {5470--5479},
  year = {2022},
  url = {https://jonbarron.info/mipnerf360/}
}

@article{hedman2018deepblending,
  author = {Peter Hedman and Julien Philip and True Price and Jan-Michael Frahm and George Drettakis and Gabriel Brostow},
  title = {Deep Blending for Free-Viewpoint Image-Based Rendering},
  journal = {{ACM TOG}},
  volume = {37},
  number = {6},
  year = {2018},
  doi = {10.1145/3272127.3275084},
  url = {http://visual.cs.ucl.ac.uk/pubs/deepblending/}
}

@article{knapitsch2017tanks,
  author = {Arno Knapitsch and Jaesik Park and Qian-Yi Zhou and Vladlen Koltun},
  title = {Tanks and Temples: Benchmarking Large-Scale Scene Reconstruction},
  journal = {{ACM TOG}},
  volume = {36},
  number = {4},
  year = {2017},
  doi = {10.1145/3072959.3073599},
  url = {https://www.tanksandtemples.org/}
}

@inproceedings{mildenhall2020nerf,
  author = {Ben Mildenhall and Pratul P. Srinivasan and Matthew Tancik and Jonathan T. Barron and Ravi Ramamoorthi and Ren Ng},
  title = {{NeRF}: Representing Scenes as Neural Radiance Fields for View Synthesis},
  booktitle = {{ECCV}},
  year = {2020},
  url = {https://arxiv.org/abs/2003.08934}
}

@inproceedings{verbin2022refnerf,
  author = {Dor Verbin and Peter Hedman and Ben Mildenhall and Todd Zickler and Jonathan T. Barron and Pratul P. Srinivasan},
  title = {{Ref-NeRF}: Structured View-Dependent Appearance for Neural Radiance Fields},
  booktitle = {{CVPR}},
  year = {2022},
  url = {https://arxiv.org/abs/2112.03907}
}

@article{glover2018qubo,
  author = {Fred Glover and Gary Kochenberger and Yu Du},
  title = {A Tutorial on Formulating and Using {QUBO} Models},
  journal = {arXiv preprint arXiv:1811.11538},
  year = {2018},
  url = {https://arxiv.org/abs/1811.11538}
}

@inproceedings{eagles2024,
  author    = {Girish, Sharath and Gupta, Kamal and Shrivastava, Abhinav},
  title     = {{EAGLES}: Efficient Accelerated {3D} Gaussians with Lightweight EncodingS},
  booktitle = {{ECCV}},
  series    = {{LNCS}},
  volume    = {15121},
  pages     = {54--71},
  year      = {2024},
  doi       = {10.1007/978-3-031-73036-8_4}
}

@inproceedings{lightgaussian2024,
  author    = {Fan, Zhiwen and Wang, Kevin and Wen, Kairun and Zhu, Zehao and Xu, Dejia and Wang, Zhangyang},
  title     = {{LightGaussian}: Unbounded {3D} Gaussian Compression with 15x Reduction and 200+ {FPS}},
  booktitle = {{NeurIPS}},
  volume    = {37},
  pages     = {140138--140158},
  year      = {2024}
}

@inproceedings{mesongs2024,
  author    = {Xie, Shuzhao and Zhang, Weixiang and Tang, Chen and Bai, Yunpeng and Lu, Rongwei and Ge, Shijia and Wang, Zhi},
  title     = {{MesonGS}: Post-training Compression of {3D} Gaussians via Efficient Attribute Transformation},
  booktitle = {{ECCV}},
  pages     = {434--452},
  year      = {2024},
  doi       = {10.1007/978-3-031-73414-4_25}
}

@inproceedings{trimmingfat2024,
  author    = {Ali, Muhammad Salman and Qamar, Maryam and Bae, Sung-Ho and Tartaglione, Enzo},
  title     = {Trimming the Fat: Efficient Compression of {3D} Gaussian Splats through Pruning},
  booktitle = {{BMVC}},
  year      = {2024},
  url       = {https://bmva-archive.org.uk/bmvc/2024/papers/Paper_358/paper.pdf}
}

@misc{coreset3dgs2026,
  author        = {Mousa, Waseem and Maalouf, Alaa},
  title         = {Provable Pruning for Efficient {3D} Gaussian Splatting via Coresets},
  year          = {2026},
  eprint        = {2607.02721},
  archivePrefix = {arXiv},
  primaryClass  = {cs.CV},
  url           = {https://arxiv.org/abs/2607.02721}
}

@inproceedings{pup3dgs2025,
  author    = {Hanson, Alex and Tu, Allen and Singla, Vasu and Jayawardhana, Mayuka and Zwicker, Matthias and Goldstein, Tom},
  title     = {{PUP} {3D-GS}: Principled Uncertainty Pruning for {3D} Gaussian Splatting},
  booktitle = {{CVPR}},
  pages     = {5949--5958},
  year      = {2025}
}

@inproceedings{refine2026,
  author    = {Chen, Zhang and Wan, Shuai and Yu, Mengting and Yang, Fuzheng and Hou, Junhui},
  title     = {{REFINE}: Super-Efficient {3D} Gaussian Splatting Pruning via Rendering-Free Primitive Importance},
  booktitle = {{ECCV}},
  series    = {{LNCS}},
  volume    = {17027},
  pages     = {131--147},
  year      = {2026},
  publisher = {Springer},
  doi       = {10.1007/978-3-032-37602-2_8}
}

@inproceedings{c3dgs2024,
  author    = {Niedermayr, Simon and Stumpfegger, Josef and Westermann, R{\"u}diger},
  title     = {Compressed {3D} Gaussian Splatting for Accelerated Novel View Synthesis},
  booktitle = {{CVPR}},
  pages     = {10349--10358},
  year      = {2024}
}

@misc{gaussianpop2026,
  author        = {Lee, Soonbin and Kim, Yeong-Gyu and Sasse, Simon and Borges, Tom{\'a}s M. and S{\'a}nchez, Yago and Ryu, Eun-Seok and Schierl, Thomas and Hellge, Cornelius},
  title         = {{GaussianPOP}: Principled Simplification Framework for Compact {3D} Gaussian Splatting via Error Quantification},
  year          = {2026},
  eprint        = {2602.06830},
  archivePrefix = {arXiv},
  primaryClass  = {cs.CV},
  url           = {https://arxiv.org/abs/2602.06830}
}

@article{potr2026,
  author  = {Ramlot, Bert and Courteaux, Martijn and Lambert, Peter and Van Wallendael, Glenn},
  title   = {{POTR}: Post-Training {3DGS} Compression},
  journal = {{IEEE TCSVT}},
  volume  = {36},
  number  = {8},
  pages   = {12168--12181},
  year    = {2026},
  doi     = {10.1109/TCSVT.2026.3685779}
}

@inproceedings{rap2026,
  author    = {Yang, Kaifa and Yang, Qi and Xu, Yiling and Li, Zhu},
  title     = {{RAP}: Fast Feedforward Rendering-Free Attribute-Guided Primitive Importance Score Prediction for Efficient {3D} Gaussian Splatting Processing},
  booktitle = {{CVPR}},
  pages     = {33323--33332},
  year      = {2026}
}

@inproceedings{compact3dgs2024,
  author    = {Lee, Joo Chan and Rho, Daniel and Sun, Xiangyu and Ko, Jong Hwan and Park, Eunbyung},
  title     = {Compact {3D} Gaussian Representation for Radiance Field},
  booktitle = {{CVPR}},
  pages     = {21719--21728},
  year      = {2024}
}

@inproceedings{lp3dgs2024,
  author    = {Zhang, Zhaoliang and Song, Tianchen and Lee, Yongjae and Yang, Li and Peng, Cheng and Chellappa, Rama and Fan, Deliang},
  title     = {{LP-3DGS}: Learning to Prune {3D} Gaussian Splatting},
  booktitle = {{NeurIPS}},
  volume    = {37},
  year      = {2024},
  doi       = {10.52202/079017-3891}
}

@inproceedings{gaussianspa2025,
  author    = {Zhang, Yangming and Jia, Wenqi and Niu, Wei and Yin, Miao},
  title     = {{GaussianSpa}: An {``Optimizing-Sparsifying''} Simplification Framework for Compact and High-Quality {3D} Gaussian Splatting},
  booktitle = {{CVPR}},
  pages     = {26673--26682},
  year      = {2025}
}

@inproceedings{maskgaussian2025,
  author    = {Liu, Yifei and Zhong, Zhihang and Zhan, Yifan and Xu, Sheng and Sun, Xiao},
  title     = {{MaskGaussian}: Adaptive {3D} Gaussian Representation from Probabilistic Masks},
  booktitle = {{CVPR}},
  pages     = {681--690},
  year      = {2025}
}

@inproceedings{omg2025,
  author    = {Lee, Joo Chan and Ko, Jong Hwan and Park, Eunbyung},
  title     = {Optimized Minimal {3D} Gaussian Splatting},
  booktitle = {{NeurIPS}},
  volume    = {38},
  year      = {2025},
  doi       = {10.52202/085713-4534}
}

@misc{qirf2026,
  author        = {Jiang, Shizeng and Zhang, Hao and Ma, Xuerui and Hu, Ying and Zhang, Tao},
  title         = {{QIRF}: Quantum-Inspired Non-Orthogonal Function-Space Compression for {3D} Gaussian Splatting},
  year          = {2026},
  eprint        = {2607.18067},
  archivePrefix = {arXiv},
  primaryClass  = {cs.CV},
  url           = {https://arxiv.org/abs/2607.18067}
}

@inproceedings{ghap2025,
  author    = {Wang, Tao and Li, Mengyu and Zeng, Geduo and Meng, Cheng and Zhang, Qiong},
  title     = {Gaussian Herding across Pens: An Optimal Transport Perspective on Global Gaussian Reduction for {3DGS}},
  booktitle = {{NeurIPS}},
  volume    = {38},
  year      = {2025},
  doi       = {10.52202/085713-5277}
}

@article{nerfpp2020,
  author  = {Zhang, Kai and Riegler, Gernot and Snavely, Noah and Koltun, Vladlen},
  title   = {{NeRF++}: Analyzing and Improving Neural Radiance Fields},
  journal = {arXiv preprint arXiv:2010.07492},
  year    = {2020},
  url     = {https://arxiv.org/abs/2010.07492}
}

@inproceedings{hogs2025,
  author    = {Liu, Xinpeng and Huang, Zeyi and Okura, Fumio and Matsushita, Yasuyuki},
  title     = {{HoGS}: Unified Near and Far Object Reconstruction via Homogeneous Gaussian Splatting},
  booktitle = {{CVPR}},
  pages     = {26714--26722},
  year      = {2025},
  url       = {https://arxiv.org/abs/2503.19232}
}
\bibliographystyle{iclr2027_conference}

\clearpage
\appendix

\section{Method derivations and implementation details}
\label{app:method}

\subsection{Notation}
\label{app:notation}

We use the notation of Section~\ref{sec:method}. The current model has
$N$ Gaussians, partitioned into disjoint blocks $\mathcal C_b$ of size
$n_b\leq M$. The target retained count is $K$, and the final block
budgets satisfy $\sum_b K_b=K$. During allocation, $K_b$ is updated
from its initial Top-$K$ count to its final integer budget.

For each block, $\mathcal P_b$ contains selected view-pixel pairs,
and $m_b=3|\mathcal P_b|$ denotes the number of sampled color channels.
Thus $J_b\in\mathbb R^{m_b\times n_b}$ and
$R_b=J_b^{\mathsf T}J_b\in\mathbb R^{n_b\times n_b}$.

The depth index $\ell$ is distinct from Gaussian identity $i$.
For a fixed view and pixel, $g_\ell$ denotes the global ID of the
Gaussian at depth rank $\ell$.
In particular, $\alpha_\ell(\boldsymbol\eta)=\eta_{g_\ell}\alpha_\ell$.
The reference for response evaluation is the current model at each
pruning round, not a fixed model shared by all rounds.

Table~\ref{tab:method-notation} collects these main symbols.
\begin{table}[htbp]
\caption{Principal method notation. Other scalar parameters are defined at their first use.}
\label{tab:method-notation}
\centering
\small
\begin{tabular}{@{}p{0.23\linewidth}p{0.71\linewidth}@{}}
\toprule
Symbol & Meaning \\
\midrule
$\mathcal G,N,K$ & Current Gaussian model, its Gaussian count, and target retained count for the current pruning round. \\
$\mathbf x_b,\mathbf z_b$ & Block retention and deletion vectors in $\{0,1\}^{n_b}$, with $\mathbf z_b=\mathbf1-\mathbf x_b$. Subscripts are omitted in local derivations. \\
$\mathcal C_b,n_b,M$ & Set of Gaussian indices in block $b$, its size $n_b=|\mathcal C_b|$, and maximum allowed block size. \\
$B,K_b$ & Number of blocks and block retention budget, updated from the initial Top-$K$ count to the final integer budget. \\
$\boldsymbol\eta$ & Continuous opacity multipliers in $[0,1]^N$. \\
$v,p,q$ & Camera index, pixel index, and RGB channel index. \\
$L_{vp},g_\ell$ & Number of Gaussians covering pixel $p$ in view $v$, and the global Gaussian ID at depth rank $\ell$ for that pixel. \\
$I_{vpq}(\boldsymbol\eta)$ & Rendered RGB channel $q$ at pixel $p$ in view $v$, evaluated with opacity multipliers $\boldsymbol\eta$ in the full scene. \\
$\mathcal P_b,m_b$ & Selected view-pixel pairs for block $b$, and the sampled channel count $m_b=3|\mathcal P_b|$. \\
$J_b,R_b$ & Sampled opacity Jacobian in $\mathbb R^{m_b\times n_b}$ and its Gram matrix $R_b=J_b^{\mathsf T}J_b\in\mathbb R^{n_b\times n_b}$. \\
$D_i$ & Allocation score $(R_b)_{ii}$ for $i\in\mathcal C_b$, equal to the summed squared opacity responses over sampled channels. \\
$\widehat K_b$ & Continuous block quota from logarithmic smoothing. \\
$\epsilon_a,\beta$ & Pseudocount for handling zero initial budgets, and logarithmic smoothing parameter. \\
$\lambda$ & Coefficient of the count penalty $(\mathbf1^{\mathsf T}\mathbf x_b-K_b)^2$. \\
$I,I^{\mathrm{gt}}$ & Rendered image and supervision image for a training view, used in the refinement loss. \\
\bottomrule
\end{tabular}
\end{table}

\begin{algorithm}[t]
\caption{Similarity guided partitioning}
\label{alg:splatq}
\small
\begin{algorithmic}[1]
\Require Gaussians $\mathcal G$; block limit $M$; neighbors $k$; comparison-group size limit
\Ensure Disjoint blocks $\{\mathcal C_b\}_{b=1}^{B}$ covering all $N$ Gaussians
\Statex \textbf{1. Build the graph}
\For{two spatial passes}
    \State Assign centers $\boldsymbol\mu_i$ to grid cells and sort cells by Morton code
    \State Split the sequence into the fewest balanced groups satisfying the comparison limit
    \For{each comparison group}
        \State Compute pairwise affinities $a_{ij}$
        \State Keep each Gaussian's top-$k$ edges
    \EndFor
    \State Shift the grid origin and enlarge the cell size for the second pass
\EndFor
\State Merge edges from both passes, keep weight of repeated edges once
\Statex \textbf{2. Grow blocks in parallel}
\State Split the first-pass Morton sequence into $\lceil N/M\rceil$ balanced intervals, 
\State seed each block with its first Gaussian, and leave the others unassigned.
\For{at most 16 iterations}
    \State Each unassigned Gaussian proposes the block of its strongest assigned neighbor
    \For{each block $\mathcal C_b$, in parallel}
        \State Rank proposals by descending edge weight, then ascending Gaussian index
        \State Add the first $\min(\text{number of proposals},\,n_b-|\mathcal C_b|)$ Gaussians
    \EndFor
    \State Stop if no Gaussian was assigned in this iteration
\EndFor
\Statex \textbf{3. Fill unused capacity}
\State Assign remaining Gaussians to fill each block to $n_b$ in the first-pass Morton order.
\State \Return $\{\mathcal C_b\}_{b=1}^{B}$
\end{algorithmic}
\end{algorithm}
\subsection{Sampling and Gram accumulation}
\label{app:sampling}

For each block and selected training view, we generate candidate pixels only from Gaussians 
whose projected radii reported by the rasterizer are positive.
Let $\mathbf u_{vi}$ be a projected center and $h_{vi}>0$ its radius
in pixels. The candidate locations are
\begin{gather}
 \mathbf u_{vi}+\max(h_{vi},1)\boldsymbol\delta,
 \qquad \boldsymbol\delta\in\mathcal D,\nonumber\\
 \mathcal D=\{(0,0),(\pm0.5,0),(0,\pm0.5),(\pm0.35,\pm0.35)\}.
 \label{eq:sampling-offsets}
\end{gather}
This yields the projected center and eight neighboring locations for each Gaussian.
We round their coordinates to the nearest integer pixels and discard locations outside the image boundaries.
For each block, we remove duplicate pixels within each view and collect the selected pairs $(v,p)$ across views into $\mathcal P_b$.

Responses are evaluated through the full scene rendering, 
with the Jacobian columns restricted to block members. Gaussians outside
the block and the background therefore remain part of each response.
Partitioning the sampled rows into batches gives
\begin{equation}
 R_b=\sum_{t=1}^{T_b}(J_b^{[t]})^{\mathsf T}J_b^{[t]},
 \qquad
 D_i=(R_b)_{ii}
 =\sum_{(v,p)\in\mathcal P_b}\sum_{q=1}^{3}
 (J_b)_{(v,p,q),i}^{2}.
 \label{eq:gram-batches}
\end{equation}
Here $T_b$ is the number of batches, and each batch contains all
three channels of its sampled pixels.
Batch accumulation equals the full Gram product up to floating-point
rounding and does not require storing the full Jacobian.
If no valid pixels are sampled across the selected views, the
zero-initialized matrix and its diagonal scores remain zero.

\begin{algorithm}[t]
\caption{Representative pixel selection}
\label{alg:pixel-selection}
\small
\begin{algorithmic}[1]
\Require Block $\mathcal C_b$ and selected training views
\Ensure Sampled view-pixel pairs $\mathcal P_b$
\State $\mathcal P_b\gets\varnothing$
\State $\mathcal D\gets\{(0,0),(\pm0.5,0),(0,\pm0.5),(\pm0.35,\pm0.35)\}$
\For{each selected view $v$}
    \State Obtain projected centers $\mathbf u_{vi}$ and screen radii $h_{vi}$ for $i\in\mathcal C_b$
    \State Generate $p=\operatorname{round}\!\left(\mathbf u_{vi}+\max(h_{vi},1)\boldsymbol\delta\right)$
    for $h_{vi}>0$ and $\boldsymbol\delta\in\mathcal D$
    \State Remove duplicate pixels and discard those outside the image
    \State Add the remaining view-pixel pairs $(v,p)$ to $\mathcal P_b$
\EndFor
\State \Return $\mathcal P_b$
\end{algorithmic}
\end{algorithm}
\subsection{Derivation of the opacity response}
\label{app:derivative}

For a fixed view $v$, pixel $p$, and color channel $q$, consider Gaussian $g_\ell$ at depth rank $\ell$. 
We keep the projected footprints, depth order, and all other opacity multipliers fixed. 
We then rewrite Equation~\eqref{eq:opacity-compositing} as the sum of contributions from $g_\ell$, the Gaussians in front of it, and the content behind it:
\begin{equation}
 I_{vpq}(\boldsymbol\eta)
 =A^{<\ell}+T_\ell\left[
 \alpha_\ell(\boldsymbol\eta)c_\ell
 +(1-\alpha_\ell(\boldsymbol\eta))C_\ell^{\mathrm{back}}
 \right].
 \label{eq:composite-split}
\end{equation}
Here $A^{<\ell}$ is the foreground contribution and
$C_\ell^{\mathrm{back}}$ is the suffix color composited with unit
incoming transmittance. Neither term, nor $T_\ell$, depends on
$\eta_{g_\ell}$. At the reference model, the suffix is
\begin{align}
 C_\ell^{\mathrm{back}}
 &=\sum_{r=\ell+1}^{L_{vp}}
 \left[\prod_{u=\ell+1}^{r-1}(1-\alpha_u)\right]
 \alpha_r c_r\nonumber\\
 &\quad+\left[\prod_{u=\ell+1}^{L_{vp}}(1-\alpha_u)\right]
 c_q^{\mathrm{bg}}.
 \label{eq:suffix-color}
\end{align}
Empty products equal one, so the suffix after the last Gaussian is
the background color. Since
$\partial\alpha_\ell(\boldsymbol\eta)/\partial\eta_{g_\ell}
=\alpha_\ell$, for $i=g_\ell\in\mathcal C_b$ we obtain
\begin{equation}
 (J_b)_{(v,p,q),i}
 =\left.\frac{\partial I_{vpq}(\boldsymbol\eta)}{\partial\eta_i}
 \right|_{\boldsymbol\eta=\mathbf1}
 =T_\ell\alpha_\ell(c_\ell-C_\ell^{\mathrm{back}}).
 \label{eq:derivative-proof}
\end{equation}
All quantities on the right are evaluated at the reference model.
The first color term is the Gaussian's direct response; the suffix
term captures the change in visibility of everything behind it.

When the projected footprints remain fixed and alpha values are not clamped, 
the rendered channel is an affine function of each opacity multiplier, 
with all others held fixed.
A single deletion is therefore represented exactly by its derivative
under these assumptions. Multiple deletions introduce products of
multiplier changes through transmittance, so their combined first-order
response is generally approximate. 
In practice, alpha clamping, contribution thresholds, and early ray termination 
can invalidate the exact correspondence between 
the derivative and the effect of deleting one Gaussian.

\subsection{Capacity handling}
\label{app:allocation}

Equation~\eqref{eq:smooth-budget} produces continuous quotas summing
to $K$. To enforce capacities during integer conversion, the
implementation first clips them and initializes integer budgets:
\begin{equation}
 \widetilde K_b=\min\{n_b,\max(0,\widehat K_b)\},
 \qquad K_b\leftarrow\lfloor\widetilde K_b\rfloor.
 \label{eq:capped-quotas}
\end{equation}
It then recomputes the remaining budget and residuals as
\begin{equation}
 H=K-\sum_bK_b,
 \qquad f_b=\widetilde K_b-K_b.
 \label{eq:quota-remainder}
\end{equation}
While $H>0$, eligible blocks are those with $K_b<n_b$.
They are sorted by descending $f_b$, with ties broken by block index.
Each of the first $\min(H,\text{number of eligible blocks})$ blocks
receives one unit. The remaining budget and residuals are then updated,
and the procedure repeats until $H=0$.

If no quota is clipped, one pass reduces to largest-remainder rounding.
Clipping can create a larger deficit, requiring multiple passes.
Each increment respects capacity and reduces the remaining deficit.
The procedure therefore terminates with $0\leq K_b\leq n_b$ and $\sum_bK_b=K$.
For $K=0$ or disabled smoothing, the initial budgets are returned unchanged.

\subsection{Joint deletion}
\label{app:taylor}

Let $\mathbf f_b(\boldsymbol\xi)\in\mathbb R^{m_b}$ stack the sampled
RGB channels when block-member opacity multipliers are
$\boldsymbol\xi\in\mathbb R^{n_b}$. All multipliers outside the block
remain one. For a small perturbation $\mathbf u$,
\begin{equation}
 \mathbf f_b(\mathbf1-\mathbf u)-\mathbf f_b(\mathbf1)
 =-J_b\mathbf u+O(\|\mathbf u\|_2^2).
 \label{eq:rgb-taylor}
\end{equation}
The squared magnitude of its linear term is
\begin{equation}
 \|J_b\mathbf u\|_2^2=\mathbf u^{\mathsf T}R_b\mathbf u.
 \label{eq:quadratic-taylor}
\end{equation}
With binary deletion variables $\mathbf z_b=\mathbf1-\mathbf x_b$,
the approximate deletion cost is:
\begin{equation}
 E_{\mathrm{del}}(\mathbf z_b)
 =\mathbf z_b^{\mathsf T}R_b\mathbf z_b
 =\sum_i(R_b)_{ii}z_{b,i}
 +2\sum_{i<j}(R_b)_{ij}z_{b,i}z_{b,j}.
 \label{eq:deletion-energy}
\end{equation}
The diagonal is the individual squared response. The off-diagonal
terms measure reinforcement or cancellation between response columns.
Substituting $\mathbf z_b=\mathbf1-\mathbf x_b$ and adding the count
penalty yields Equation~\eqref{eq:retention-qubo} in the main text.

Our objective approximates changes from the current rendering rather than error relative to ground truth. 
This approximation may become less accurate when many Gaussians are deleted together. 
We therefore recompute the responses after model refinement in each subsequent pruning round.

\subsection{QUBO coefficients and the Ising mapping}
\label{app:qubo}

Within this subsection, write $n=n_b$, $K=K_b$ and $R=R_b$ for local quantities.
Expanding Equation~\eqref{eq:retention-qubo} and using $x_i^2=x_i$ gives
\begin{align}
 E(\mathbf x)&=C+\sum_i a_ix_i+\sum_{i<j}b_{ij}x_ix_j,
 \nonumber\\
 a_i&=R_{ii}-2\sum_jR_{ij}+\lambda(1-2K),\nonumber\\
 b_{ij}&=2R_{ij}+2\lambda,
 \qquad C=\mathbf1^{\mathsf T}R\mathbf1+\lambda K^2.
 \label{eq:qubo-coefficients}
\end{align}
The constant does not affect selection. 
We store the rendering objective as a vector of linear coefficients and a sparse list of quadratic coefficients. 
Each pair $(i,j)$ is stored once with $i<j$. 

For spins $\sigma_i\in\{-1,+1\}$, substitute $x_i=(1+\sigma_i)/2$:
\begin{align}
 E(\boldsymbol\sigma)
 &=C'+\sum_i h_i\sigma_i
 +\sum_{i<j}\mathcal J_{ij}\sigma_i\sigma_j,\nonumber\\
 \mathcal J_{ij}&=b_{ij}/4,
 \qquad h_i=a_i/2+\tfrac14\sum_{j\ne i}b_{ij},\nonumber\\
 C'&=C+\tfrac12\sum_i a_i+\tfrac14\sum_{i<j}b_{ij}.
 \label{eq:ising-map}
\end{align}
Here $b_{ji}=b_{ij}$ when evaluating the field sum.
For the backend convention $H(\boldsymbol\sigma)=-\boldsymbol\sigma^{\mathsf T}W\boldsymbol\sigma$,
the symmetric pair entries are $W_{ij}=-\mathcal J_{ij}/2$.
Fields are encoded using one reference spin $\sigma_0$, with
$W_{i0}=W_{0i}=-h_i/2$ and zero diagonal entries.
Decoding uses $x_i=(1+\sigma_i\sigma_0)/2$, so an $n$-variable
block requires $n+1$ spins in this encoding.

\subsection{Greedy selection, Simulated Annealing and CIM Solvers}
\label{app:solver}

Again use local $R,n,K$ and deletion cost
$E_{\mathrm{del}}(\mathbf z)=\mathbf z^{\mathsf T}R\mathbf z$.

\paragraph{Greedy selection.}
Starting with $\mathbf z=\mathbf0$, deleting a retained member $j$
changes the cost by
\begin{equation}
 \Delta_{\mathrm{add}}(j)=2(R\mathbf z)_j+R_{jj}.
 \label{eq:greedy-marginal}
\end{equation}
Greedy selection takes the smallest increment at each step until
exactly $n-K$ members are deleted. It satisfies the count constraint
by construction and does not require a count penalty.

\paragraph{Simulated annealing.}
Simulated annealing (SA) minimizes the penalized objective with the count-penalty coefficient $\lambda=1$, using the Ising encoding in Appendix~\ref{app:qubo}.
We initialize multiple random spin configurations and apply Metropolis
updates while reducing the temperature geometrically.
For a proposed spin flip with energy change $\Delta H$, the acceptance
probability at temperature $\theta>0$ is
\begin{equation}
 p_{\mathrm{accept}}=\min\{1,\exp(-\Delta H/\theta)\}.
 \label{eq:sa-acceptance}
\end{equation}
Energy decreases are always accepted, whereas temporary increases allow
the search to leave local minima.
Lower temperatures reduce the probability of accepting such increases.
By default, the initial and final temperatures are 100 and $10^{-3}$,
respectively, with a cooling factor of 0.99.
We use 10 iterations per temperature and a solution limit of 100 per block.
We select the returned state with the lowest Ising energy and decode
its retention variables using the reference spin.

\paragraph{Coherent Ising Machine (CIM).}
The coherent Ising machine (CIM) uses the same Ising formulation and reference-spin encoding as SA, but with a different count-penalty coefficient, $\lambda=0.0625$.
Before submission, we rescale and truncate the matrix coefficients to a signed 14-bit representation.
We request multiple samples and select the returned state with the lowest energy under the submitted matrix.
The selected state is then decoded into binary retention variables.
Neither SA nor CIM guarantees an exact retained count, we therefore apply the following postprocessing.

\subsection{Postprocessing}
\label{app:method-config}

\paragraph{Capacity repairing.}
For an infeasible solver output, count repair changes one bit at a time.
If too many Gaussians are retained, it deletes the retained member
with the smallest increase in unpenalized energy. If too few are
retained, it restores the deleted member with the smallest increase.
The coefficients and current state determine each increment exactly,
and the process stops when the retained count equals $K$.
Repair can increase the unpenalized cost because feasibility takes priority.

\paragraph{One-swap refinement.}
For a deleted member $i$ and retained member $j$, a one-swap move
restores $i$ and deletes $j$. Its energy change is
\begin{equation}
 \Delta_{\mathrm{swap}}(i,j)
 =2\bigl[(R\mathbf z)_j-(R\mathbf z)_i\bigr]
 +R_{ii}+R_{jj}-2R_{ij}.
 \label{eq:swap-marginal}
\end{equation}
When enabled, refinement repeatedly selects the best exchange and
accepts it only when $\Delta_{\mathrm{swap}}<-\tau$, for a numerical
tolerance $\tau\geq0$. It stops when no qualifying exchange remains
or the configured swap limit is reached. Every accepted swap preserves
the count and decreases the objective. Without an active iteration
limit, termination gives a one-swap local optimum within tolerance. 
For $K=0$ or $K=n$, the feasible state is unique and no search is needed.

We use one-swap refinement as a shared local-search step rather than a separate methodological contribution.
The same refinement rule and stopping criteria are applied after greedy selection, SA, and CIM.
We therefore keep this step fixed in the solver comparisons, which evaluate each solver together with the shared postprocessing.
\section{Experiment}
\label{app:experiment_details}

This section supplements the experimental results in Section~\ref{sec:experiments}.

\begin{table}[t]
    \caption{
        Scoring time, solver time, and peak GPU memory across three datasets.
        Times are reported in seconds and memory in GB.
        For SplatQ, scoring denotes $R_b$ calculation.
        A dash indicates that no separate QUBO solver is used.
    }
    \label{tab:pruning_cost}
    \centering
    \footnotesize
    \setlength{\tabcolsep}{3pt}
    \begin{tabular*}{\linewidth}{
        @{\extracolsep{\fill}}l*{9}{r}@{}
    }
        \toprule
        & \multicolumn{3}{c}{Mip-NeRF 360}
        & \multicolumn{3}{c}{Deep Blending}
        & \multicolumn{3}{c}{Tanks \& Temples} \\
        \cmidrule(lr){2-4}
        \cmidrule(lr){5-7}
        \cmidrule(lr){8-10}
        Method
        & Scoring & Solving & Memory
        & Scoring & Solving & Memory
        & Scoring & Solving & Memory \\
        \midrule
        LightGS
        & 4.63 & -- & 9.91
        & 5.26 & -- & 7.74
        & 6.12 & -- & 5.52 \\
        MesonGS
        & 6.17 & -- & 7.64
        & 20.34 & -- & 6.50
        & 8.59 & -- & 4.02 \\
        EAGLES
        & 1.10 & -- & 8.36
        & 1.12 & -- & 7.80
        & 1.01 & -- & 5.00 \\
        C3DGS
        & 4.37 & -- & 10.05
        & 4.31 & -- & 8.48
        & 3.24 & -- & 5.47 \\
        PUP
        & 17.5 & -- & 10.1
        & 16.3 & -- & 8.7
        & 20.1 & -- & 5.5 \\
        REFINE
        & 0.26 & -- & 1.5
        & 0.25 & -- & 1.4
        & 0.27 & -- & 1.3 \\
        RAP
        & 89.4 & -- & 16.4
        & 88.5 & -- & 14.6
        & 41.7 & -- & 9.2 \\
        CoreSet
        & 1240.4 & -- & 4.0
        & 876.3 & -- & 4.4
        & 638.9 & -- & 4.1 \\
        \midrule
        SplatQ (greedy)
        & 1005.0 & 101.1 & 10.41
        & 1237.1 & 99.1 & 8.56
        & 3456.3 & 58.9 & 5.45 \\
        SplatQ (SA)
        & 1005.0 & 845.9 & 10.41
        & 1237.1 & 748.4 & 8.56
        & 3456.3 & 448.9 & 5.45 \\
        SplatQ (CIM)
        & 1005.0 & 581.4 & 10.41
        & 1237.1 & 514.5 & 8.56
        & 3456.3 & 309.6 & 5.45 \\
        \bottomrule
    \end{tabular*}
\end{table}

\subsection{Experiment Settings}
\label{app:experiment_details:settings}

\paragraph{Datasets and Metrics}
We evaluate 27 scenes from five benchmarks: Mip-NeRF 360 (9), Deep Blending (2), Tanks and Temples (2), NeRF Synthetic (8), and Ref-NeRF/Shiny Blender (6)~\citep{barron2022mipnerf360, hedman2018deepblending,knapitsch2017tanks,mildenhall2020nerf,verbin2022refnerf}.
We compare against the unpruned 3DGS reference and eight pruning baselines, reporting dataset-average PSNR, SSIM, LPIPS-VGG, and measured rendering FPS.
Macro scores average the five dataset means equally.

\paragraph{Evaluation settings.}
FPS measurements were conducted on an Ubuntu 22.04 server equipped with NVIDIA GeForce RTX 5090 GPUs and Intel Xeon Gold 6459C processors.
The inspected evaluation pipeline saves test-view renderings and computes PSNR, SSIM, and LPIPS-VGG.
The FPS testing procedure warms up with ten test views and times three passes over all test views.

\paragraph{Implementation Details.}
For each scene, all pruning methods start from the same 3DGS model pretrained for 30k iterations and retain 10\% of its original Gaussians.
All methods follow the same two-round schedule: removing 80\% of the original Gaussians, then 50\% of the survivors.
Each round includes 10k fine-tuning iterations, for 20k in total, using the same loss $0.8\mathcal L_1+0.2(1-\mathrm{SSIM})$ with densification disabled.
We adapt the baselines by retaining their Gaussian selection mechanisms within this shared pruning and fine-tuning pipeline.
Methods originally evaluated without fine-tuning also receive the same fine-tuning schedule, and attribute compression is disabled for all methods.
SplatQ uses simulated annealing (SA) in the main comparison, with greedy and CIM evaluated as solver ablations.
Further details are provided in Appendix~\ref{app:notation} and Table~\ref{tab:splatq_settings}.

\paragraph{Ablation settings.}
Table~\ref{tab:ablation_results} compares budget assignment and selection variants.
Top-k globally retains the $K$ Gaussians with the largest diagonal scores $D_i$.
Uniform block quotas assigns the same retention fraction to all blocks, whereas Budget allocation uses our logarithmically smoothed quotas.
Both variants select the highest-scoring Gaussians within each block using $D_i$.
The Quadratic variants use the full objective with smoothed quotas and are solved using greedy selection, SA, or CIM.
Quadratic, no allocation variant instead uses uniform block quotas and greedy selection.
For each variant in this table, we repeat the complete pruning and fine-tuning pipeline five times using seeds 0 through 4, starting from the same pretrained models.

\begin{table}[t]
\caption{Partitioning, response, and budget settings. Parameters without dedicated symbols in Section~\ref{sec:method} are described by name.}
\label{tab:splatq_settings}
\centering
\small
\begin{tabular}{@{}ll@{}}
\toprule
Parameter & Setting \\
\midrule
Maximum block size $M$ & 512 \\
Neighbors per Gaussian per pass $k$ & 4 \\
Comparison group size limit & 512 \\
Spatial passes & 2 \\
maximum growth iterations & 16 \\
Response views & All training views \\
Response image width and height & Input dimensions divided by 4 \\
Pixel cap per block per view & 256 \\
Budget smoothing $\beta$ & 3 \\
Budget pseudocount $\epsilon_a$ & 1 \\
Spherical-harmonic degree & 3 \\
\bottomrule
\end{tabular}
\end{table}

\paragraph{Algorithm settings.}
Table~\ref{tab:splatq_settings} lists the partitioning, sampling, and allocation settings.
We use $M=512$ and $k=4$, with at most 512 Gaussians in each temporary comparison group.
Successive spatial passes change the grid origin using half-cell offsets.
Graph construction uses two passes, and parallel block growth runs for at most 16 iterations.
For each block, we use all training views with image width and height reduced to one quarter of their input dimensions.
At this resolution, we generate candidate pixels using the projected-center offsets in Algorithm~\ref{alg:pixel-selection}.
After removing duplicate locations and those outside the image, We evaluate $J_b$ at the selected pixels.
We use $\beta=3$ and $\epsilon_a=1$ for budget assignment.

\paragraph{QUBO solver settings.}
Greedy selection enforces $K_b$ directly and therefore does not use a count penalty.
The SA backend uses Kaiwu's simulated annealing optimizer with an initial temperature of 100, a cutoff of $10^{-3}$, and a cooling factor of 0.99.
It uses 10 iterations per temperature, eight worker processes, and a solution limit of 100 per block.
The SA random seed is the base seed 0 plus the block index.
After decoding the lowest-energy returned state, the implementation repairs its count and applies up to $n_b$ improving one-swap exchanges.
The CIM backend additionally truncates its Ising coefficients to signed 14-bit precision and applies the same repair and refinement.

\paragraph{Fine-tuning settings.}
Fine-tuning uses Adam with numerical stabilizer $10^{-15}$ and one randomly selected training view per update.
The initial learning rates for DC coefficients, higher-order coefficients, opacity, scale, and rotation are $0.0025$, $0.000125$, $0.05$, $0.005$, and $0.001$, respectively.
These rates decay by a factor of 0.95 every 400 refinement steps.
Position updates use the inherited exponential schedule, with endpoints $1.6\times10^{-4}$ and $1.6\times10^{-6}$ multiplied by the scene scale.
Refinement and final evaluation preserve input resolution when image width is at most 1600 pixels.
Wider images are resized to approximately 1600 pixels in width while preserving aspect ratio.
In the picture-free setting, neither pruning nor fine-tuning uses the original training images.
We retain the training camera parameters and supervise fine-tuning with images rendered by the fixed, original pretrained model. Ground-truth test images are used only for evaluation.

\paragraph{Metrics methods.}
\label{app:metric_methods}
For Figure~\ref{fig:spatial_quality_balance}, we divide each test image
into a $16\times16$ grid of 256 non-overlapping patches.
We compute patch PSNR against the corresponding ground truth using
all pixels for Mip-NeRF 360, Deep Blending, and Tanks \& Temples.
For NeRF Synthetic and Ref-NeRF, we exclude white background pixels
connected to the image boundary using a four-connected flood fill
on the ground truth (all RGB channels $\geq 250$).
PSNR is computed over the remaining pixels in each patch and patches with no remaining pixels are omitted.
For each image, we calculate the variance and range
(maximum minus minimum) of these patch PSNR values.
We average each statistic over test views within each scene,
then equally over scenes within each benchmark.
Full-image PSNR is reported alongside these statistics to distinguish
overall reconstruction quality from spatial quality variation.

For Figure~\ref{fig:volume_cdf}, we normalize Gaussian volumes using
a shared spatial reference for each scene.
We construct an axis-aligned bounding box from the coordinate-wise
minima and maxima of Gaussian centers across all nine pruned models.
Its diagonal length is
\[
D=\sqrt{(x_{\max}-x_{\min})^2+
        (y_{\max}-y_{\min})^2+
        (z_{\max}-z_{\min})^2}.
\]
Each Gaussian's volume is divided by $D^3$, so all methods use
the same normalization within a scene.
For each method, we compute normalized volume quantiles separately
for each scene.
At each cumulative fraction, we average the corresponding quantiles
equally across scenes within each benchmark.
The resulting curves summarize scene-level volume distributions
without pooling Gaussians across scenes.

\subsection{Additional Experimental Results}

\subsubsection{Ablations with Standard Deviations}

Table~\ref{tab:ablation_results} supplements the mean results in
Table~\ref{tab:ablation_results_mean} with standard deviations.
For each variant, we repeat the complete pruning and fine-tuning
pipeline five times using random seeds 0 through 4, starting from
the same pretrained models.
We first average scene-level metrics within each dataset, then report
the mean and standard deviation of these dataset averages across runs.

\begin{table}[!t]
\centering
\caption{Component and solver ablations at 90\% pruning. Each dataset cell reports the mean above its $\pm$ standard deviation across five runs of the complete pruning and fine-tuning pipeline. }
\label{tab:ablation_results}
\fontsize{8.0}{9.2}\selectfont
\setlength{\tabcolsep}{0.5pt}\renewcommand{\arraystretch}{1.10}
\newcommand{\abcell}[2]{\shortstack[c]{#1\\[-.4ex]{\fontsize{7.0}{7.8}\selectfont$\pm$\mbox{#2}}}}
\begin{tabular*}{\linewidth}{@{\extracolsep{\fill}}l*{9}{r}@{}}
\toprule
Variant & \multicolumn{3}{c}{Mip-NeRF 360} & \multicolumn{3}{c}{Deep Blending} & \multicolumn{3}{c}{Tanks and Temples}\\
\cmidrule(lr){2-4}\cmidrule(lr){5-7}\cmidrule(lr){8-10}
 & PSNR$\uparrow$ & SSIM$\uparrow$ & LPIPS$\downarrow$ & PSNR$\uparrow$ & SSIM$\uparrow$ & LPIPS$\downarrow$ & PSNR$\uparrow$ & SSIM$\uparrow$ & LPIPS$\downarrow$\\
\midrule
Top-$k$ & \abcell{26.68}{4.1e-3} & \abcell{0.7787}{2.50e-5} & \abcell{0.2640}{7.40e-5} & \abcell{29.09}{1.31e-2} & \abcell{0.8944}{5.85e-5} & \abcell{0.2649}{7.42e-5} & \abcell{22.72}{7.6e-3} & \abcell{0.8034}{8.67e-5} & \abcell{0.2469}{1.158e-4}\\
+ Uniform block quotas & \abcell{26.52}{5.5e-3} & \abcell{0.7755}{2.75e-5} & \abcell{0.2685}{8.31e-5} & \abcell{28.98}{1.42e-2} & \abcell{0.8930}{5.37e-5} & \abcell{0.2668}{8.32e-5} & \abcell{22.64}{8.1e-3} & \abcell{0.8004}{9.12e-5} & \abcell{0.2505}{1.269e-4}\\
+ Budget assignment & \abcell{26.74}{4.8e-3} & \abcell{0.7889}{2.79e-5} & \abcell{0.2626}{8.07e-5} & \abcell{29.15}{1.35e-2} & \abcell{0.8958}{5.87e-5} & \abcell{0.2638}{7.92e-5} & \abcell{22.81}{8.3e-3} & \abcell{0.8050}{9.03e-5} & \abcell{0.2455}{1.240e-4}\\
+ Quadratic, greedy & \abcell{27.03}{5.2e-3} & \abcell{0.7928}{2.82e-5} & \abcell{0.2552}{8.27e-5} & \abcell{29.42}{1.38e-2} & \abcell{0.9020}{5.26e-5} & \abcell{0.2568}{7.05e-5} & \abcell{23.15}{7.5e-3} & \abcell{0.8135}{8.77e-5} & \abcell{0.2357}{1.141e-4}\\
Quadratic, uniform quotas & \abcell{26.79}{4.7e-3} & \abcell{0.7811}{2.65e-5} & \abcell{0.2602}{8.01e-5} & \abcell{29.36}{1.53e-2} & \abcell{0.8985}{5.65e-5} & \abcell{0.2585}{7.81e-5} & \abcell{23.07}{8.6e-3} & \abcell{0.8093}{8.45e-5} & \abcell{0.2424}{1.374e-4}\\
\textbf{+ Quadratic, SA} & \abcell{27.07}{4.4e-3} & \abcell{0.7932}{2.56e-5} & \abcell{0.2547}{7.90e-5} & \abcell{29.47}{1.37e-2} & \abcell{0.9027}{5.34e-5} & \abcell{0.2562}{7.08e-5} & \abcell{23.21}{7.3e-3} & \abcell{0.8160}{8.41e-5} & \abcell{0.2322}{1.393e-4}\\
+ Quadratic, CIM & \abcell{27.14}{2.0e-3} & \abcell{0.7960}{1.12e-4} & \abcell{0.2507}{9.12e-5} & \abcell{29.54}{1.24e-2} & \abcell{0.9041}{1.22e-4} & \abcell{0.2532}{1.20e-4} & \abcell{23.21}{1.27e-2} & \abcell{0.8173}{3.5e-5} & \abcell{0.2296}{1.25e-4}\\
\bottomrule
\end{tabular*}
\par\smallskip
\begin{tabular*}{\linewidth}{@{\extracolsep{\fill}}l*{9}{r}@{}}
\toprule
Variant & \multicolumn{3}{c}{NeRF Synthetic} & \multicolumn{3}{c}{Ref-NeRF} & \multicolumn{3}{c}{Macro}\\
\cmidrule(lr){2-4}\cmidrule(lr){5-7}\cmidrule(lr){8-10}
 & PSNR$\uparrow$ & SSIM$\uparrow$ & LPIPS$\downarrow$ & PSNR$\uparrow$ & SSIM$\uparrow$ & LPIPS$\downarrow$ & PSNR$\uparrow$ & SSIM$\uparrow$ & LPIPS$\downarrow$\\
\midrule
Top-$k$ & \abcell{31.85}{9.9e-3} & \abcell{0.9568}{1.43e-5} & \abcell{0.0503}{6.86e-5} & \abcell{29.55}{8.7e-3} & \abcell{0.9407}{4.31e-5} & \abcell{0.0972}{5.88e-5} & 27.978 & 0.87480 & 0.18466\\
+ Uniform block quotas & \abcell{31.72}{8.5e-3} & \abcell{0.9557}{1.48e-5} & \abcell{0.0520}{7.12e-5} & \abcell{29.43}{7.8e-3} & \abcell{0.9396}{3.77e-5} & \abcell{0.0993}{6.60e-5} & 27.858 & 0.87284 & 0.18742\\
+ Budget assignment & \abcell{31.96}{8.9e-3} & \abcell{0.9575}{1.29e-5} & \abcell{0.0496}{7.51e-5} & \abcell{29.64}{9.0e-3} & \abcell{0.9415}{3.97e-5} & \abcell{0.0961}{6.79e-5} & 28.060 & 0.87774 & 0.18352\\
+ Quadratic, greedy & \abcell{32.31}{9.1e-3} & \abcell{0.9618}{1.45e-5} & \abcell{0.0427}{6.62e-5} & \abcell{29.96}{9.4e-3} & \abcell{0.9449}{3.73e-5} & \abcell{0.0906}{7.08e-5} & 28.374 & 0.88300 & 0.17620\\
Quadratic, uniform quotas & \abcell{32.14}{9.1e-3} & \abcell{0.9596}{1.40e-5} & \abcell{0.0452}{6.97e-5} & \abcell{29.76}{8.9e-3} & \abcell{0.9431}{4.22e-5} & \abcell{0.0924}{6.95e-5} & 28.224 & 0.87832 & 0.17974\\
\textbf{+ Quadratic, SA} & \abcell{32.37}{8.3e-3} & \abcell{0.9623}{1.44e-5} & \abcell{0.0424}{7.42e-5} & \abcell{30.01}{8.7e-3} & \abcell{0.9453}{4.13e-5} & \abcell{0.0897}{6.30e-5} & 28.426 & 0.88390 & 0.17504\\
+ Quadratic, CIM & \abcell{32.48}{7.0e-3} & \abcell{0.9635}{3.2e-5} & \abcell{0.0409}{6.5e-5} & \abcell{30.07}{3.6e-3} & \abcell{0.9456}{3.3e-5} & \abcell{0.0887}{1.27e-4} & 28.488 & 0.88530 & 0.17262\\
\bottomrule
\end{tabular*}
\end{table}

\subsubsection{Ablation on Gaussian Partitioning Strategies}
\label{app:partition-ablation}

We examine whether more elaborate grouping criteria improve the quality--cost
trade-off of blockwise Gaussian pruning. Tables~\ref{tab:partition-quality}
and~\ref{tab:partition-cost} summarize reconstruction quality and preprocessing
cost, respectively, for our default partitioner and six alternatives across
five datasets. The pruning protocol retains $20\%$ of the original Gaussians
in the first round and $10\%$ in the second, with 10,000 ground-truth-supervised
fine-tuning iterations per round. Blocks contain at most 512 Gaussians, and
the second round repartitions the fine-tuned first-round model. Quality is
evaluated after the second round and averaged equally over scenes within
each dataset.

\paragraph{Partitioning variants.}
Our default method constructs a local position-, scale-, and color-based
similarity graph and applies capacity-constrained graph growth.
\emph{Random} replaces graph-based partitioning with balanced splitting of a
random permutation. \emph{Morton} instead splits the voxel-based Morton order
directly, without graph growth.
\emph{Rotation} replaces the position distance in the edge
weights with quaternion angular distance, retaining spatial candidate
generation and color similarity.
\emph{Full covariance} replaces the same distance with a Mahalanobis distance
using full covariances, likewise retaining spatial candidates
and color similarity.
\emph{Shared ray} replaces the local similarity graph with a graph weighted
by joint alpha-compositing contributions to sampled pixels across all training
views. \emph{Screen tile} uses products of contributions averaged within
$16\times16$ tiles instead of same-pixel contributions.
All graph-based variants retain top-four neighbor selection and
capacity-constrained graph growth.

\paragraph{Reconstruction quality and cost.}
Table~\ref{tab:partition-quality} shows reconstruction quality on all five datasets.
Table~\ref{tab:partition-cost} separates graph construction and partitioning
from first-round Gate Fisher computation.

\begin{table}[t]
\centering
\caption{Partitioning ablation using the opacity-response Gram matrix $R_b$:
reconstruction quality after two-round pruning to 10\% retention and 10,000
fine-tuning iterations per round. Macro averages the five dataset-level
means equally. Higher SSIM and PSNR and lower LPIPS are better.
Bold denotes the best displayed value, including ties.}
\label{tab:partition-quality}
\begingroup
\setlength{\tabcolsep}{3pt}
\renewcommand{\arraystretch}{1.12}
\resizebox{\linewidth}{!}{%
\begin{tabular}{@{}l*{9}{r}@{}}
\toprule
 & \multicolumn{3}{c}{Mip-NeRF 360} & \multicolumn{3}{c}{Deep Blending} & \multicolumn{3}{c}{Tanks \& Temples} \\
\cmidrule(lr){2-4}\cmidrule(lr){5-7}\cmidrule(lr){8-10}
Partitioner & SSIM$\uparrow$ & PSNR$\uparrow$ & LPIPS$\downarrow$ & SSIM$\uparrow$ & PSNR$\uparrow$ & LPIPS$\downarrow$ & SSIM$\uparrow$ & PSNR$\uparrow$ & LPIPS$\downarrow$ \\
\midrule
Ours & 0.7932 & \textbf{27.07} & \textbf{0.2547} & \textbf{0.9027} & \textbf{29.47} & \textbf{0.2562} & \textbf{0.8160} & \textbf{23.21} & \textbf{0.2322} \\
Random & 0.7881 & 26.49 & 0.2565 & 0.9013 & 29.24 & 0.2565 & 0.8095 & 22.86 & 0.2362 \\
Morton & 0.7928 & 27.05 & 0.2551 & 0.9026 & 29.45 & 0.2565 & 0.8145 & 23.13 & 0.2345 \\
Rotation-aware & 0.7928 & 27.03 & 0.2549 & 0.9025 & 29.40 & 0.2565 & 0.8145 & 23.15 & 0.2352 \\
Full covariance & 0.7927 & 27.05 & 0.2552 & \textbf{0.9027} & 29.46 & 0.2563 & 0.8145 & 23.19 & 0.2352 \\
Shared ray & \textbf{0.7935} & 27.06 & 0.2548 & 0.9024 & 29.44 & 0.2570 & 0.8137 & 23.13 & 0.2358 \\
Screen tile & 0.7927 & 27.05 & 0.2559 & 0.9026 & 29.45 & 0.2564 & 0.8145 & 23.15 & 0.2348 \\
\bottomrule
\end{tabular}%
}

\par\vspace{0.8em}

\resizebox{\linewidth}{!}{%
\begin{tabular}{@{}l*{9}{r}@{}}
\toprule
 & \multicolumn{3}{c}{NeRF Synthetic} & \multicolumn{3}{c}{Ref-NeRF} & \multicolumn{3}{c}{Macro} \\
\cmidrule(lr){2-4}\cmidrule(lr){5-7}\cmidrule(lr){8-10}
Partitioner & SSIM$\uparrow$ & PSNR$\uparrow$ & LPIPS$\downarrow$ & SSIM$\uparrow$ & PSNR$\uparrow$ & LPIPS$\downarrow$ & SSIM$\uparrow$ & PSNR$\uparrow$ & LPIPS$\downarrow$ \\
\midrule
Ours & \textbf{0.9623} & \textbf{32.37} & 0.0424 & \textbf{0.9453} & \textbf{30.01} & \textbf{0.0897} & \textbf{0.8839} & \textbf{28.43} & \textbf{0.1750} \\
Random & 0.9621 & 32.24 & 0.0428 & 0.9449 & 30.00 & 0.0901 & 0.8812 & 28.17 & 0.1764 \\
Morton & 0.9619 & 32.30 & 0.0427 & 0.9450 & 29.98 & 0.0900 & 0.8834 & 28.38 & 0.1758 \\
Rotation-aware & \textbf{0.9623} & 32.36 & 0.0429 & \textbf{0.9453} & \textbf{30.01} & \textbf{0.0897} & 0.8835 & 28.39 & 0.1758 \\
Full covariance & 0.9622 & 32.34 & 0.0424 & 0.9452 & 30.00 & \textbf{0.0897} & 0.8835 & 28.41 & 0.1758 \\
Shared ray & 0.9620 & 32.32 & 0.0426 & 0.9450 & 29.98 & 0.0899 & 0.8833 & 28.39 & 0.1760 \\
Screen tile & \textbf{0.9623} & 32.35 & \textbf{0.0420} & 0.9451 & 29.99 & \textbf{0.0897} & 0.8834 & 28.40 & 0.1758 \\
\bottomrule
\end{tabular}%
}
\endgroup
\end{table}

\begin{table}[t]
\centering
\caption{Partitioning ablation: recorded preprocessing costs, averaged per
scene within each dataset. $T_{\mathrm{part}}$ denotes graph construction plus
partitioning in seconds; $T_{\mathrm{F}}$ denotes first-round
opacity-response Gram matrix $R_b$ computation in minutes.}
\label{tab:partition-cost}
\begingroup
\setlength{\tabcolsep}{4pt}
\renewcommand{\arraystretch}{1.12}
\resizebox{\linewidth}{!}{%
\begin{tabular}{@{}l*{10}{r}@{}}
\toprule
& \multicolumn{2}{c}{Mip-NeRF 360}
& \multicolumn{2}{c}{Deep Blending}
& \multicolumn{2}{c}{Tanks \& Temples}
& \multicolumn{2}{c}{NeRF Synthetic}
& \multicolumn{2}{c}{Ref-NeRF} \\
\cmidrule(lr){2-3}\cmidrule(lr){4-5}\cmidrule(lr){6-7}\cmidrule(lr){8-9}\cmidrule(lr){10-11}
Partitioner & $T_{\mathrm{part}}$ (s) & $T_{\mathrm{F}}$ (min)
& $T_{\mathrm{part}}$ (s) & $T_{\mathrm{F}}$ (min)
& $T_{\mathrm{part}}$ (s) & $T_{\mathrm{F}}$ (min)
& $T_{\mathrm{part}}$ (s) & $T_{\mathrm{F}}$ (min)
& $T_{\mathrm{part}}$ (s) & $T_{\mathrm{F}}$ (min) \\
\midrule
Ours & 0.86 & 16.74 & 0.80 & 20.61 & 0.66 & 57.60 & 0.48 & 5.77 & 0.46 & 3.44 \\
Random & 0.22 & 92.09 & 0.21 & 160.68 & 0.21 & 250.78 & 0.20 & 9.75 & 0.20 & 7.29 \\
Morton & 0.31 & 16.47 & 0.29 & 18.97 & 0.27 & 50.43 & 0.29 & 5.26 & 0.29 & 3.09 \\
Rotation-aware & 0.98 & 20.69 & 0.87 & 24.24 & 0.78 & 73.74 & 0.60 & 6.51 & 0.58 & 3.66 \\
Full covariance & 155.36 & 17.64 & 135.72 & 20.75 & 83.25 & 57.07 & 13.65 & 5.74 & 8.03 & 3.42 \\
Shared ray & 149.18 & 21.44 & 269.64 & 20.90 & 295.11 & 62.27 & 49.53 & 5.58 & 48.06 & 3.24 \\
Screen tile & 258.64 & 23.18 & 1819.88 & 21.70 & 3703.23 & 66.21 & 525.83 & 6.29 & 203.68 & 3.40 \\
\bottomrule
\end{tabular}%
}
\endgroup
\end{table}

\subsubsection{Pruning Time and Memory}
\label{app:additional_experiment:cost}
Table~\ref{tab:pruning_cost} reports mean per-scene scoring time, QUBO solving time, and peak GPU memory. We average scenes equally within each dataset: nine for Mip-NeRF 360 and two each for Deep Blending and Tanks and Temples. We perform one pruning pass, removing 80\% of the Gaussians from the original 3D-GS. These measurements use RTX 5090 GPUs on the same servers.
For SplatQ, scoring includes partitioning and $R_b$ calculation, taking 1005.0, 1237.2, and 3456.3\,s on the three datasets, respectively. Selection includes budget allocation and solving QUBOs including necessary postprocessing. 
Greedy selection takes 101.1, 99.1, and 58.9\,s, whereas SA takes 845.9, 748.4, and 448.9\,s.
Greedy therefore reduces solver time by approximately $7.6$--$8.4\times$, although $R_b$ calculation remains the main computational cost.
The reported peak GPU memory is the memory peak over scoring and QUBO solving stages. We average these per-scene peaks. SplatQ’s peak GPU memory is similar to that of LightGS and C3DGS.
These measurements show the substantial offline cost of SplatQ’s interaction-aware selection.

Pruning latency should be considered within the complete compression pipeline, which also includes 20k fine-tuning iterations in our main setting.
Our refinement sweeps show that longer fine-tuning generally improves reconstruction quality within the tested range.
For a fixed pruning workload, longer fine-tuning reduces the relative contribution of pruning to total preparation time.
Moreover, this cost is incurred offline and is not repeated during rendering.
For applications that prioritize final reconstruction quality, the observed quality gains can therefore justify additional pruning time.
Building on the scene-level pruning cost measurements in
Table~\ref{tab:pruning_cost}, we measured QUBO solver runtimes across
backends, submission modes, and timing intervals. The first two rows
of Table~\ref{tab:qubo-solver-runtimes} report the runtimes of greedy
selection and simulated annealing (SA) on Mip-NeRF 360, with each solver
processing 59,112 matrices. For each solver, we tested several
concurrency levels on a single NVIDIA RTX~5090 and used the level that
achieved the highest measured throughput. The selected concurrency
levels were 64 for greedy selection and 8 for SA.

\begin{table}[t]
\centering
\caption{QUBO solver runtimes across backends, submission modes, and
timing intervals. The mean time per matrix is the elapsed time divided
by the batch size. Wall-clock timings that include matrix transfer may
vary with network conditions.}
\label{tab:qubo-solver-runtimes}
\small
\setlength{\tabcolsep}{3.5pt}
\renewcommand{\arraystretch}{1.12}
\begin{tabularx}{\linewidth}{@{}lXrrrrr@{}}
\toprule
Solver &
Mode &
\shortstack[r]{Batch\\Size} &
\shortstack[r]{Concurrent\\Solves} &
\shortstack[r]{Elapsed\\Time (s)} &
\shortstack[r]{Mean Time\\(s/matrix)} &
\shortstack[r]{Single-matrix\\Time (s)} \\
\midrule
RTX 5090 & Greedy selection                & 59,112 & 64 & 910    & 0.0154 & ---    \\
RTX 5090 & Simulated annealing             & 59,112 &  8 & 7,631  & 0.1291 & ---    \\
\addlinespace
CIM & Public cloud, cluster site          & 2,396  & 15 & 496    & 0.2070 & ---    \\
CIM & Public cloud, Hong Kong VPN         & 2,396  & 15 & 1,170  & 0.4883 & ---    \\
\addlinespace
CIM & Direct access, end-to-end           & 15     & 15 & 1.3275 & 0.0885 & 1.3275 \\
CIM & Direct access, on-device processing & 15     & 15 & 0.700  & 0.0467 & 0.700  \\
CIM & Direct access, optical evolution    & 15     & 15 & 0.075  & 0.0050 & 0.075  \\
\bottomrule
\end{tabularx}
\end{table}

The provider's official cloud platform offers public access to a
cluster of 15 coherent Ising machines (CIMs). The platform schedules
the matrices for solving after the entire batch has been uploaded.
We measure wall-clock time from the start of the upload until all
matrices in the batch have been solved. To avoid resource contention,
the benchmark required an exclusive reservation during which all other
compute jobs on the cluster were suspended. The available reservation
window limited the benchmark to the 2,396 matrices required for the
second pruning round of the Bicycle scene. Each matrix occupied
approximately 0.1~MB when serialized as a \texttt{.pkl} file.
Rows~3 and~4 of Table~\ref{tab:qubo-solver-runtimes} use the same batch.
Submission from the cluster site took 496~s, and submission through a
VPN endpoint in Hong Kong, China, took 1,170~s. The observed difference
of 674~s provides an estimate of the additional overhead associated
with cross-region submission under the tested conditions.

The CIM timings in Table~\ref{tab:pruning_cost} were obtained through
direct access, a non-public channel that the provider makes available
for academic testing. In this mode, the client sends matrices directly
to the 15 CIMs, bypassing the public cloud relay and its matrix
validation step. Row~5 of Table~\ref{tab:qubo-solver-runtimes} reports
an end-to-end wall-clock time of 1.3275~s for solving 15 matrices in
parallel on 15 CIMs. This interval includes matrix transfer, waiting
for automatic optical tuning, and on-device processing. Row~6 reports
an on-device processing time of 0.700~s, which includes matrix retrieval
from the scheduler, compilation, and optical evolution. Row~7 reports
an optical evolution time of 75~ms for a single matrix.
Optimizing matrix transfer from the client to the CIMs could reduce
end-to-end wall-clock time. Faster matrix retrieval from the scheduler
could reduce both on-device processing time and end-to-end wall-clock
time. Table~\ref{tab:qubo-solver-runtimes} reports timings at the finest
granularity permitted by the provider's disclosure policy.

\subsubsection{Effect of Pairwise Interactions without Fine-tuning.}
\begin{figure}[!t]
\centering
\includegraphics[width=\linewidth]{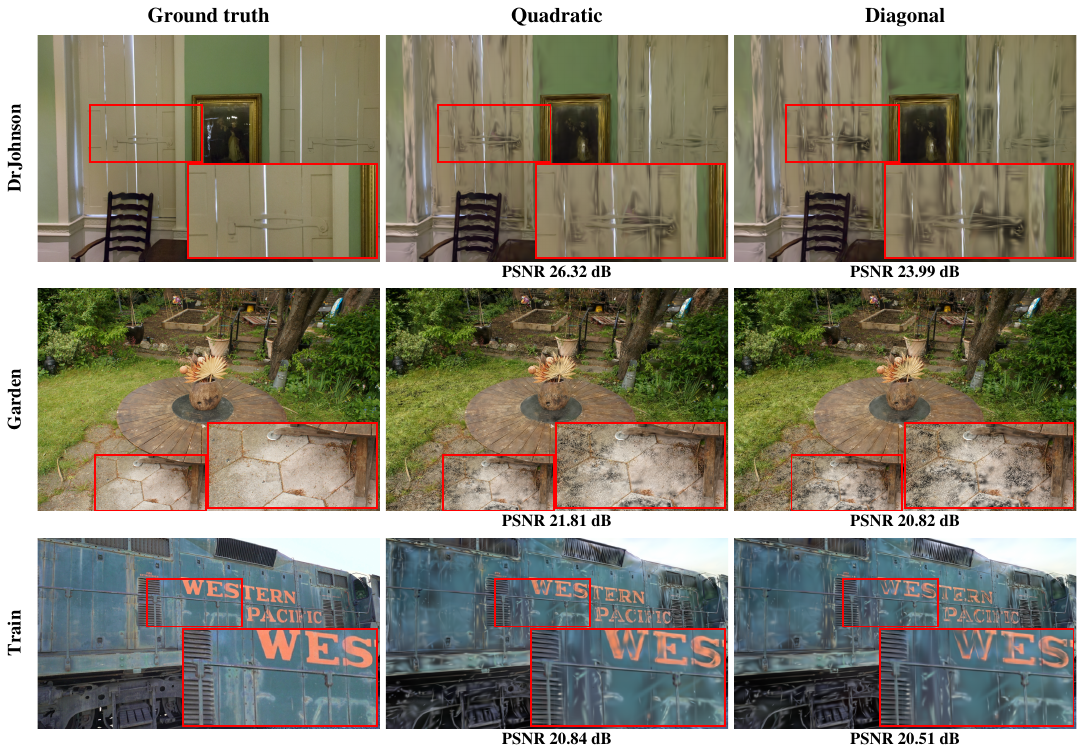}
\caption{Visual comparison after a single round of 80\% pruning without fine-tuning.}
\label{fig:pairwise_no_ft}
\end{figure}

To isolate the contribution of pairwise interactions, we compare a diagonal-only objective with the full quadratic objective.
We solve the diagonal-only problem exactly by retaining the $K_b$ Gaussians with the largest $D_i$ in each block.
The full quadratic problem is solved using SA with the settings described above.
Both variants use the same pretrained models, partitions, and block quotas.
We perform one run per variant per scene, removing 80\% of the Gaussians in a single round without fine-tuning.
Table~\ref{tab:pairwise_no_ft} reports dataset-average metrics from these runs.
Including pairwise interactions improves PSNR by 0.34, 0.30, and 0.28\,dB on Mip-NeRF 360, Deep Blending, and Tanks and Temples, respectively.
Figure~\ref{fig:pairwise_no_ft} provides corresponding visual comparisons.

\begin{table}[t]
    \caption{
        Effect of pairwise interactions after a single round of 80\% pruning without fine-tuning.
        Both variants use identical pretrained models, partitions, and block quotas.
        Values are dataset averages; PSNR is measured in dB and LPIPS uses the VGG backbone.
        Better results are shown in bold.
    }
    \label{tab:pairwise_no_ft}
    \centering
    \small
    \setlength{\tabcolsep}{3pt}
    \begin{tabular*}{\linewidth}{
        @{\extracolsep{\fill}}l*{9}{r}@{}
    }
        \toprule
        & \multicolumn{3}{c}{Mip-NeRF 360}
        & \multicolumn{3}{c}{Deep Blending}
        & \multicolumn{3}{c}{Tanks \& Temples} \\
        \cmidrule(lr){2-4}
        \cmidrule(lr){5-7}
        \cmidrule(lr){8-10}
        Objective
        & PSNR$\uparrow$ & SSIM$\uparrow$ & LPIPS$\downarrow$
        & PSNR$\uparrow$ & SSIM$\uparrow$ & LPIPS$\downarrow$
        & PSNR$\uparrow$ & SSIM$\uparrow$ & LPIPS$\downarrow$ \\
        \midrule
        Diagonal
        & 21.57 & 0.6848 & 0.3330
        & 26.51 & 0.8543 & 0.3018
        & 18.71 & 0.7195 & 0.2871 \\
        Quadratic
        & \textbf{21.91} & \textbf{0.6890} & \textbf{0.3272}
        & \textbf{26.81} & \textbf{0.8572} & \textbf{0.2984}
        & \textbf{18.99} & \textbf{0.7215} & \textbf{0.2844} \\
        \bottomrule
    \end{tabular*}
\end{table}

\FloatBarrier

\subsubsection{Pruning without Fine-Tuning}
\label{sec:zero-shot}

\begin{figure}[!t]
\centering
\includegraphics[width=\linewidth]{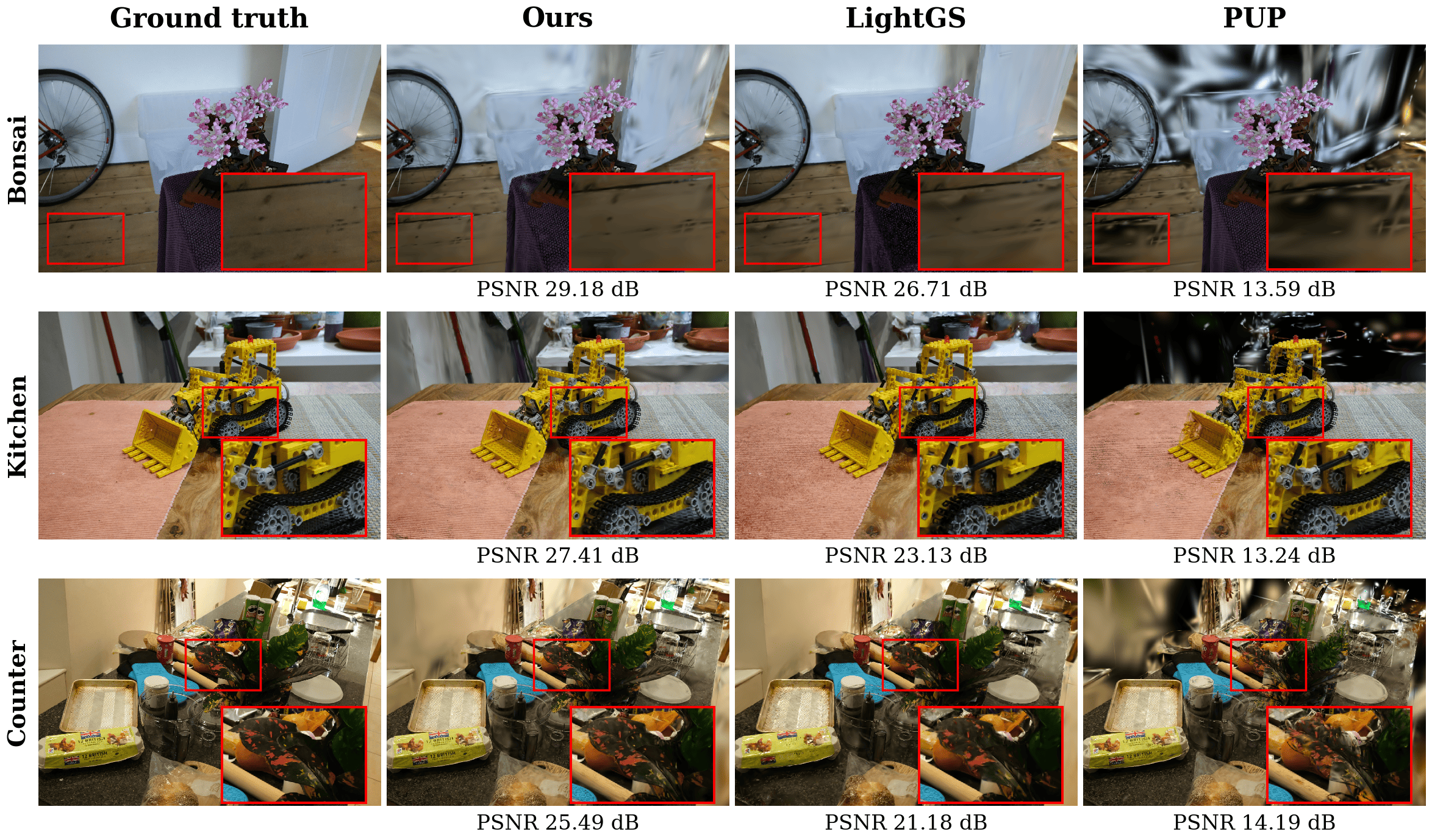}
\caption{Selected visual comparisons at 80\% pruning without fine-tuning.}
\label{fig:zero_shot_comparison}
\end{figure}

Table~\ref{tab:zero-shot-quality} reports reconstruction quality at 80\%
pruning without subsequent fine-tuning. Our method achieves the highest
PSNR and SSIM and the lowest LPIPS across the five reported datasets.
Figure~\ref{fig:zero_shot_comparison} shows selected views with matched detail
crops. 

\begin{table}[t]
\centering
\caption{Quality without fine-tuning at 80\% pruning. Best values are \textbf{bold}; second-best values are \underline{underlined}.}
\label{tab:zero-shot-quality}
\begingroup
\setlength{\tabcolsep}{2pt}
\renewcommand{\arraystretch}{1.08}
\resizebox{\linewidth}{!}{%
\begin{tabular}{@{}l*{15}{r}@{}}
\toprule
Method & \multicolumn{3}{c}{Mip-NeRF 360} & \multicolumn{3}{c}{Deep Blending} & \multicolumn{3}{c}{Tanks \& Temples} & \multicolumn{3}{c}{NeRF Synthetic} & \multicolumn{3}{c}{Ref-NeRF} \\
\cmidrule(lr){2-4}\cmidrule(lr){5-7}\cmidrule(lr){8-10}\cmidrule(lr){11-13}\cmidrule(lr){14-16}
 & PSNR$\uparrow$ & SSIM$\uparrow$ & LPIPS$\downarrow$ & PSNR$\uparrow$ & SSIM$\uparrow$ & LPIPS$\downarrow$ & PSNR$\uparrow$ & SSIM$\uparrow$ & LPIPS$\downarrow$ & PSNR$\uparrow$ & SSIM$\uparrow$ & LPIPS$\downarrow$ & PSNR$\uparrow$ & SSIM$\uparrow$ & LPIPS$\downarrow$ \\
\midrule
Ours & \textbf{21.91} & \textbf{0.6890} & \textbf{0.3272} & \textbf{26.81} & \textbf{0.8572} & \textbf{0.2984} & \textbf{18.99} & \textbf{0.7215} & \textbf{0.2844} & \textbf{25.89} & \textbf{0.9205} & \textbf{0.0698} & \textbf{26.80} & \textbf{0.9142} & \textbf{0.1206} \\
LightGS & \underline{21.50} & \underline{0.6582} & 0.3475 & \underline{26.63} & \underline{0.8469} & \underline{0.3119} & \underline{18.67} & \underline{0.7147} & \underline{0.2957} & 24.08 & 0.9106 & 0.0793 & 23.08 & \underline{0.8999} & 0.1368 \\
PUP & 19.03 & 0.6560 & \underline{0.3364} & 25.01 & 0.8403 & 0.3173 & 16.22 & 0.6913 & 0.3050 & \underline{25.18} & \underline{0.9192} & \underline{0.0760} & \underline{24.81} & 0.8992 & \underline{0.1352} \\
\bottomrule
\end{tabular}%
}
\endgroup
\end{table}

\subsubsection{Sensitivity to Block Size and Smoothing.}
Table~\ref{tab:parameter_sensitivity} evaluates block size $M$ and smoothing strength $\beta$ on Mip-NeRF 360 using SA.
We vary one parameter at a time, fixing $\beta=3$ for the block-size sweep and $M=512$ for the smoothing sweep.
Across $M\in\{256,512,1024,2048\}$, PSNR varies by only 0.07\,dB, while SSIM increases and LPIPS decreases as blocks grow.
The largest tested size, $M=2048$, achieves the best quality across all three metrics.
Increasing $M$ from 256 to 2048 also reduces scoring time from 1636.8 to 310.2\,s and solver time from 940.4 to 536.3\,s under the tested configuration.
We use \(M=512\) by default to maintain compatibility with the CIM backend used in our experiments, which supports Ising matrices with fewer than 1,000 variables.
Across $\beta\in\{1,2,3,4,5\}$, PSNR varies by only 0.05\,dB.
The default $\beta=3$ achieves the highest PSNR and SSIM, and ties with $\beta=2$ for the lowest LPIPS at the reported precision.

\begin{table}[t]
    \caption{
        Sensitivity to block size $M$ and smoothing strength $\beta$ on Mip-NeRF 360 using SA.
        We fix $\beta=3$ when varying $M$ and $M=512$ when varying $\beta$.
        PSNR is measured in dB, and scoring and solver times are reported in seconds.
        Bold indicates the best quality result within each sweep, including ties at the reported precision. 
        For the $beta$ part, we report quality only and omit timing results.
    }
    \label{tab:parameter_sensitivity}
    \centering
    \small
    \setlength{\tabcolsep}{3pt}
    \begin{tabular*}{\linewidth}{
        @{\extracolsep{\fill}}l*{9}{r}@{}
    }
        \toprule
        & \multicolumn{4}{c}{Block size $M$}
        & \multicolumn{5}{c}{Smoothing strength $\beta$} \\
        \cmidrule(lr){2-5}
        \cmidrule(lr){6-10}
        Metric
        & 256 & 512 & 1024 & 2048
        & 1 & 2 & 3 & 4 & 5 \\
        \midrule
        PSNR$\uparrow$
        & 27.02 & 27.07 & 27.05 & \textbf{27.09}
        & 27.02 & 27.04 & \textbf{27.07} & 27.06 & 27.05 \\
        SSIM$\uparrow$
        & 0.7927 & 0.7932 & 0.7934 & \textbf{0.7937}
        & 0.7926 & 0.7929 & \textbf{0.7932} & 0.7929 & 0.7925 \\
        LPIPS$\downarrow$
        & 0.2556 & 0.2547 & 0.2539 & \textbf{0.2528}
        & 0.2549 & \textbf{0.2547} & \textbf{0.2547}
        & 0.2550 & 0.2553 \\
        \midrule
        Scoring time
        & 1636.8 & 1005.0 & 541.6 & 310.2
        & -- & -- & -- & -- & -- \\
        Solver time
        & 940.4 & 845.9 & 784.0 & 536.3
        & -- & -- & -- & -- & -- \\
        \bottomrule
    \end{tabular*}
\end{table}

\subsubsection{BD-Rate Computation}
\label{sec:bd_rate}

\begin{figure*}[t]
    \centering
    \includegraphics[width=\textwidth]
        {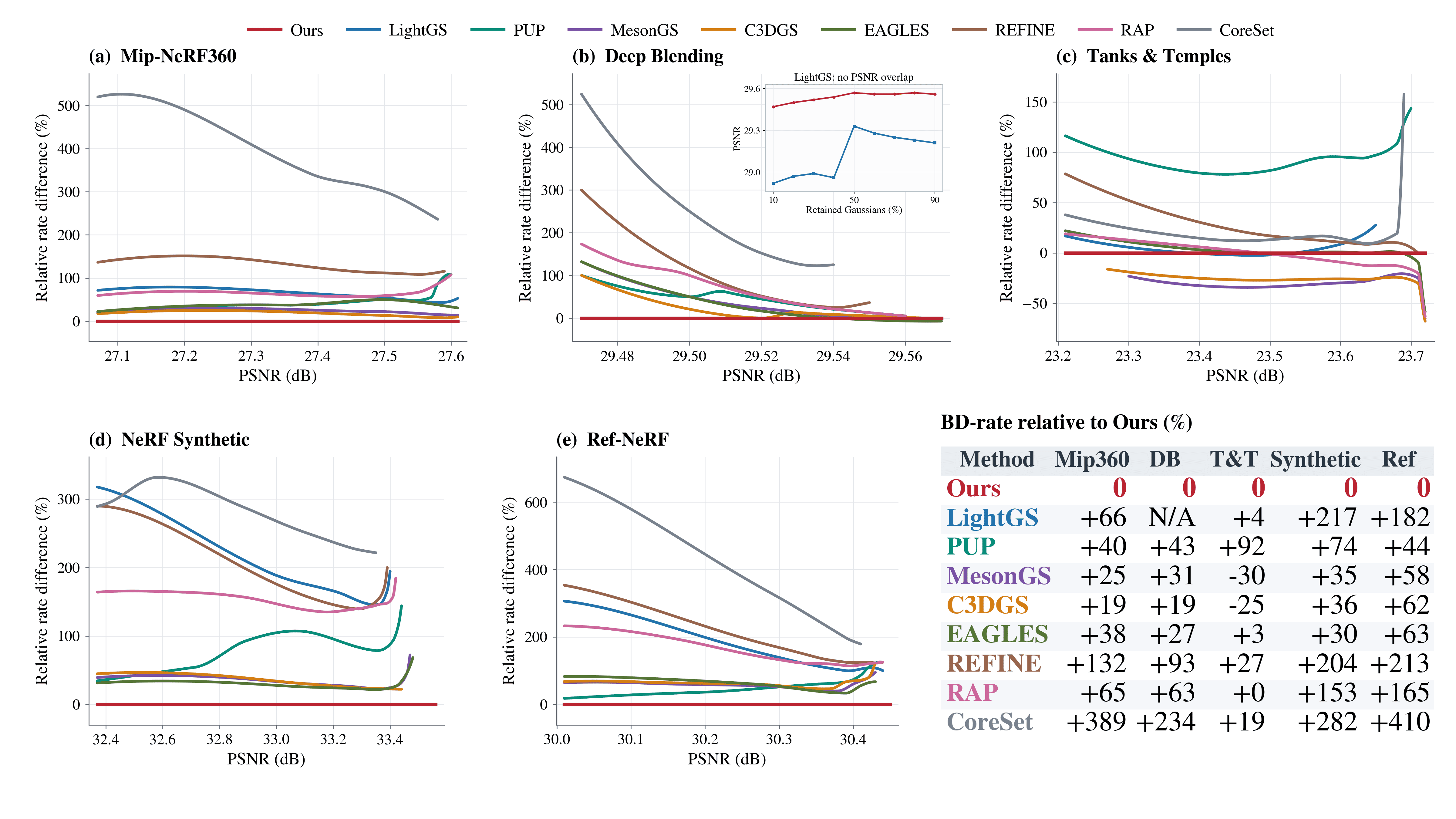}
    \caption{
        Relative Gaussian budget differences with respect to Ours.
        Panels (a)--(e) show the pointwise differences over each
        method's common PSNR range with Ours.
        Positive values indicate a larger Gaussian budget.
        The lower-right table reports BD-rates rounded to integer
        percentages.
        The inset in (b) shows the retention-ratio--PSNR measurements
        for LightGaussian and Ours on Deep Blending, where their
        PSNR ranges do not overlap.
    }
    \label{fig:relative_gaussian_budget}
\end{figure*}

To evaluate the trade-off between Gaussian count and reconstruction
quality across pruning methods, we use the Gaussian retention ratio
as a normalized budget and adopt the \mbox{Bj{\o}ntegaard} Delta Rate
(BD-rate) to quantify each method's Gaussian budget relative to ours
over a common PSNR range.
Figure~\ref{fig:relative_gaussian_budget} presents the relative budget
curves and the corresponding BD-rate results.

\paragraph{Input data and budget definition.}
For each dataset, we extract the results of each method at nine
pruning ratios ranging from 10\% to 90\% in increments of 10\%.
Let $i\in\{1,\ldots,9\}$ index the budget settings and let $p_i$
denote the corresponding pruning ratio.
The normalized Gaussian budget is defined as
\begin{equation}
    R_i = 1-p_i.
    \label{eq:gaussian_budget}
\end{equation}
Let $m$ denote the method being evaluated, and let $q_{m,i}$
denote its dataset-averaged PSNR, measured in dB, at budget
setting $i$.
The input data for method $m$ are
\begin{equation}
    \begin{aligned}
        \mathcal{D}_m
        &= \{(R_i,q_{m,i})\}_{i=1}^{9},\\
        R_i
        &\in \{0.1,0.2,\ldots,0.9\}.
    \end{aligned}
    \label{eq:budget_quality_data}
\end{equation}
For each dataset, we construct the budget--quality curves from
the dataset-averaged PSNR values and compute BD-rate independently.
All notation and calculations below refer to a single dataset.

\paragraph{Pareto-front extraction.}
We extract the Pareto front of each method to characterize its
effective budget--quality trade-off.
The experimental points are sorted in ascending order of $R_i$,
with the associated pruning ratios $p_i$ and quality values
$q_{m,i}$ reordered accordingly, and are reindexed as
$i=1,\ldots,9$.
We retain the first point and, for $i=2,\ldots,9$, retain a
point if
\begin{equation}
    q_{m,i} > \max_{1\leq j<i} q_{m,j},
    \label{eq:pareto_selection}
\end{equation}
where $j$ indexes all preceding points with smaller budgets.
This rule retains points that achieve higher PSNR as the budget
increases and selects the smallest budget among points with
identical PSNR.
We denote the retained index set by $\mathcal{I}_m$, with indices
referring to the sorted points, and use these points for
interpolation.

\paragraph{Log-budget interpolation and common quality range.}
We use the shape-preserving piecewise cubic Hermite interpolating
polynomial (PCHIP) to construct $f_m(q)$, with PSNR as the
independent variable and the natural logarithm of the budget as
the dependent variable.
Here, $q$ denotes a continuous PSNR level, and the interpolant
satisfies
\begin{equation}
    f_m(q_{m,i}) = \ln R_i,
    \qquad i\in\mathcal{I}_m.
    \label{eq:log_budget_interpolation}
\end{equation}
The estimated Gaussian budget required by method $m$ at quality
level $q$ is
\begin{equation}
    \widehat{R}_m(q) = \exp\bigl(f_m(q)\bigr),
    \label{eq:estimated_budget}
\end{equation}
where $\widehat{R}_m(q)$ denotes the interpolated budget estimate.

Let $o$ denote our method, which serves as the reference.
Its interpolant $f_o(q)$ and budget estimate $\widehat{R}_o(q)$
are constructed using the same procedure.
The common quality interval for methods $m$ and $o$ is
$[q_L,q_H]$, with boundaries
\begin{equation}
    \begin{aligned}
        q_L &= \max(q_{m,\min},q_{o,\min}),\\
        q_H &= \min(q_{m,\max},q_{o,\max}).
    \end{aligned}
    \label{eq:common_psnr_range}
\end{equation}
The subscripts $L$ and $H$ denote the lower and upper bounds of
the common interval.
The quantities $q_{m,\min}$ and $q_{m,\max}$ denote the minimum
and maximum PSNR covered by the Pareto front of method $m$;
the corresponding quantities for $o$ are defined analogously.
We compute BD-rate when both methods have at least two
Pareto-front points and $q_H>q_L$.
Other cases are reported as N/A (unavailable).

\paragraph{BD-rate and relative budget curves.}
The mean log-budget difference between method $m$ and reference
method $o$ is
\begin{equation}
    \overline{\Delta}_m
    =
    \frac{1}{q_H-q_L}
    \int_{q_L}^{q_H}
    \bigl[f_m(q)-f_o(q)\bigr]\,\mathrm{d}q.
    \label{eq:mean_log_budget_difference}
\end{equation}
Here, $\overline{\Delta}_m$ is the uniformly weighted average
of the log-budget difference over the common PSNR interval.
The corresponding BD-rate is
\begin{equation}
    \boxed{
    \mathrm{BD\text{-}rate}_{m/o}
    =
    100\left[
        \exp\bigl(\overline{\Delta}_m\bigr)-1
    \right]\%.
    }
    \label{eq:bd_rate}
\end{equation}
The subscript $m/o$ denotes method $m$ relative to reference
method $o$.
We evaluate the integral directly from the piecewise polynomials
of the PCHIP interpolants.

BD-rate is the percentage deviation from unity of the geometric
mean of the budget ratio
$\widehat{R}_m(q)/\widehat{R}_o(q)$
over the common quality interval.
Positive values indicate that method $m$ requires a larger
Gaussian budget, whereas negative values indicate a smaller budget.
The reference method has a BD-rate of zero relative to itself.

To visualize budget differences at individual quality levels,
we define the relative budget curve as
\begin{equation}
    \delta_m(q)
    =
    100\left[
        \frac{\widehat{R}_m(q)}{\widehat{R}_o(q)}-1
    \right]\%,
    \qquad q\in[q_L,q_H].
    \label{eq:relative_budget_curve}
\end{equation}
Here, $\delta_m(q)$ is the percentage difference in Gaussian
budget between method $m$ and the reference at quality level $q$.
The relative budget curves characterize local differences across
quality levels, while BD-rate summarizes the entire common
quality interval.
Each method is compared with the reference over its respective
common quality interval.
Calculations retain full numerical precision, and the summary
table reports BD-rates rounded to integer percentages.

\subsubsection{Sensitivity to Pruning and Fine-Tuning }
\label{app:sensitivity_surfaces}
\begin{figure}[!htbp]
\centering
\includegraphics[width=\linewidth]{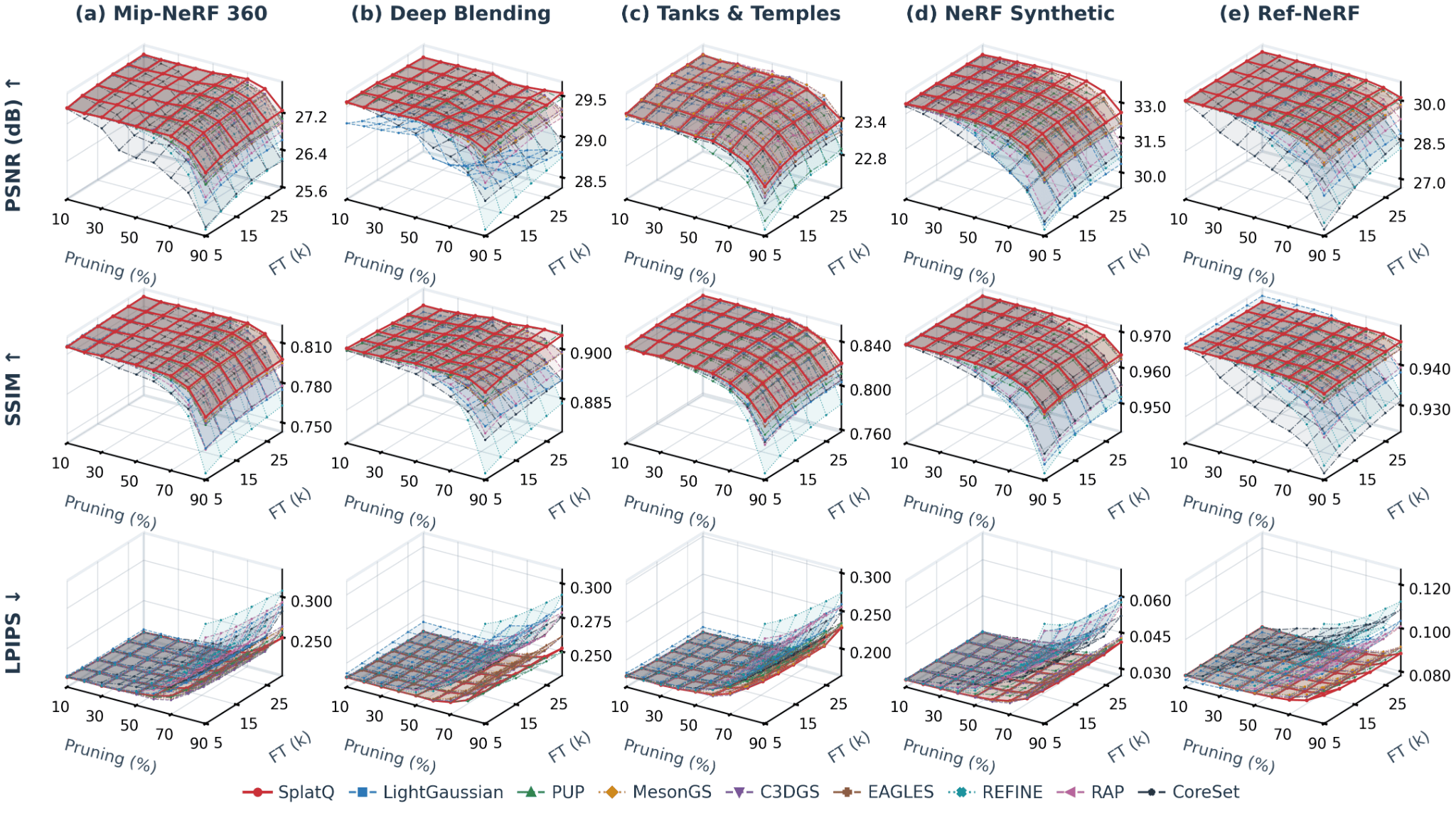}
\caption{Reconstruction quality across pruning percentages and fine-tuning budgets. Columns correspond to Mip-NeRF 360, Deep Blending, Tanks and Temples, NeRF Synthetic, and Ref-NeRF; rows show PSNR, SSIM, and LPIPS-VGG. Axes indicate removal percentage, fine-tuning iterations in thousands, and the corresponding metric. SplatQ uses SA.}
\label{fig:ratio_main_3d}
\end{figure}

\begin{figure}[H]
\centering
\includegraphics[width=\linewidth]{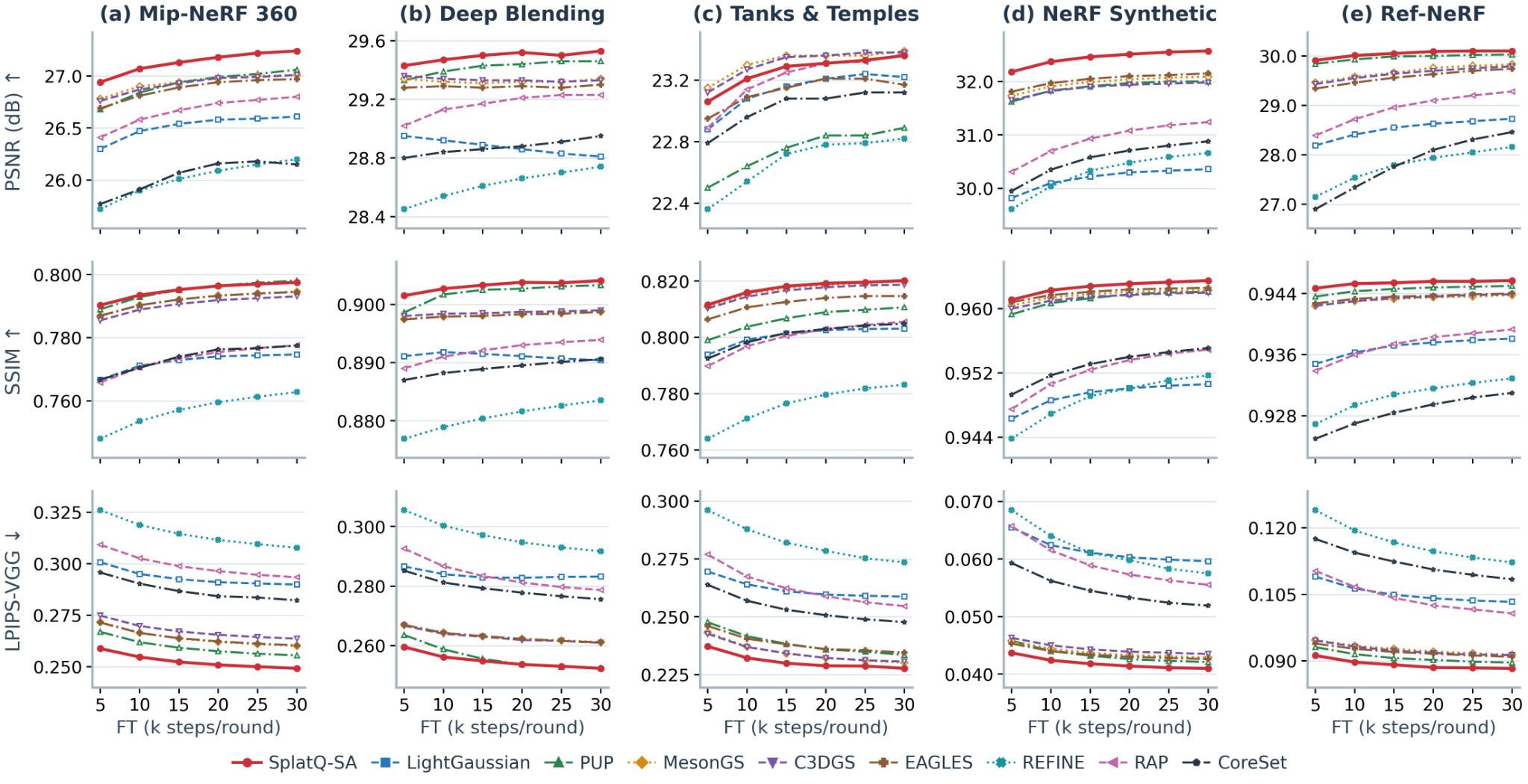}
\caption{Effect of fine-tuning duration with two pruning rounds and 90\% final removal. Columns correspond to the five datasets, and rows show PSNR, SSIM, and LPIPS-VGG. Fine-tuning steps on the horizontal axis are per round. The SplatQ curves use SA.}
\label{fig:rounds_main_2d}
\end{figure}

Figure~\ref{fig:ratio_main_3d} extends the pruning-ratio comparison
at 10k fine-tuning steps in the top panel of Figure~\ref{fig:pruning_ablations} to a broader range of pruning
ratios and nonzero fine-tuning budgets.
Figure~\ref{fig:rounds_main_2d} presents the two-round results
from the bottom panel of Figure~\ref{fig:pruning_ablations}.
In the pruning-round study, fine-tuning budgets are specified
per round: $R$ rounds with $T$ updates each entail $RT$ updates
in total.

\clearpage
\FloatBarrier

\subsubsection{Composability Experiments}
\label{app:composability}

Table~\ref{tab:composability-performance-full} extends the main-text
comparison with complete storage and runtime results across five datasets.
Original 3DGS and pruned-input sizes are listed once per dataset, since
all methods prune 90\% of the Gaussians and use the same PLY format,
yielding identical input sizes at the reported precision. We report
compressed sizes, compression ratios, decoding times, and rendering FPS
for each compressor--input combination. FPS is measured after decoding
with a common renderer and excludes decoding time. 

\begin{table}[!t]
\centering
\caption{Full compression storage and runtime performance. Original 3DGS and pruned-input sizes are
reported once per dataset; input sizes coincide at the displayed precision.
$S_{\mathrm{cmp}}$ is compressed size (MB), Rat.\ is the approximate total
compression ratio (scene-wise attribute ratio $\times 10$), and dec is decoding
time (s). FPS excludes decoding. Best and second-best FPS within each compressor
and dataset are \textbf{bold} and \underline{underlined}.}
\label{tab:composability-performance-full}
\begingroup
\small
\setlength{\tabcolsep}{2.5pt}
\renewcommand{\arraystretch}{1.0}
\begin{adjustbox}{max width=\linewidth}
\begin{tabular}{@{}ll*{12}{r}@{}}
\toprule
 & & \multicolumn{4}{c}{Mip-NeRF 360} & \multicolumn{4}{c}{Deep Blending} & \multicolumn{4}{c}{Tanks \& Temples} \\
\cmidrule(lr){3-6}\cmidrule(lr){7-10}\cmidrule(lr){11-14}
\multicolumn{2}{@{}l}{Original 3DGS / Pruned (MB)} & \multicolumn{4}{c}{833.89 / 83.39} & \multicolumn{4}{c}{737.96 / 73.80} & \multicolumn{4}{c}{442.40 / 44.24} \\
\midrule
Method & Input & $S_{\mathrm{cmp}}\downarrow$ & Rat.$\uparrow$ & dec$\downarrow$ & FPS$\uparrow$ & $S_{\mathrm{cmp}}\downarrow$ & Rat.$\uparrow$ & dec$\downarrow$ & FPS$\uparrow$ & $S_{\mathrm{cmp}}\downarrow$ & Rat.$\uparrow$ & dec$\downarrow$ & FPS$\uparrow$ \\
\midrule
\multirow{5}{*}{LightGS} & Ours & 16.28 & 51.0 & 1.59 & \textbf{1039.4} & 14.35 & 51.4 & 1.41 & \textbf{1257.4} & 8.81 & 50.0 & 0.89 & \textbf{1518.8} \\
 & PUP & 16.23 & 51.2 & 1.58 & \underline{899.7} & 14.32 & 51.5 & 1.45 & \underline{1228.9} & 8.78 & 50.2 & 0.86 & \underline{1362.0} \\
 & RAP & 16.29 & 50.9 & 1.64 & 819.7 & 14.38 & 51.3 & 1.49 & 1046.2 & 8.79 & 50.1 & 0.88 & 1317.1 \\
 & REFINE & 16.27 & 51.0 & 1.61 & 799.4 & 14.39 & 51.3 & 1.43 & 1045.4 & 8.79 & 50.1 & 0.89 & 1246.8 \\
 & LightGS & 16.25 & 51.1 & 1.61 & 594.0 & 14.31 & 51.6 & 1.44 & 784.6 & 8.79 & 50.1 & 0.87 & 1019.4 \\
\midrule
\multirow{5}{*}{MesonGS} & Ours & 5.78 & 138.9 & 4.98 & \textbf{987.4} & 5.38 & 136.9 & 4.52 & \textbf{1163.9} & 3.71 & 115.5 & 2.80 & \textbf{1462.8} \\
 & PUP & 5.78 & 138.7 & 4.90 & \underline{857.6} & 5.38 & 136.9 & 4.70 & \underline{1158.7} & 3.64 & 117.1 & 2.76 & \underline{1355.6} \\
 & RAP & 5.73 & 140.1 & 4.98 & 790.1 & 5.32 & 138.4 & 4.58 & 1016.1 & 3.64 & 117.3 & 2.91 & 1312.0 \\
 & REFINE & 5.72 & 140.3 & 4.88 & 780.3 & 5.29 & 139.1 & 4.65 & 1008.9 & 3.60 & 118.7 & 2.81 & 1226.0 \\
 & MesonGS & 5.74 & 139.8 & 5.00 & 695.5 & 5.28 & 139.4 & 4.58 & 935.1 & 3.64 & 117.3 & 2.86 & 1220.4 \\
\midrule
\multirow{5}{*}{EAGLES} & Ours & 26.30 & 31.4 & 9.38 & \textbf{987.0} & 21.09 & 35.0 & 21.29 & \textbf{1330.1} & 14.52 & 29.9 & 16.17 & \textbf{1564.6} \\
 & PUP & 25.80 & 32.0 & 9.24 & 816.2 & 20.94 & 35.2 & 21.80 & 1169.5 & 14.03 & 31.1 & 16.30 & 1487.7 \\
 & RAP & 25.60 & 32.3 & 8.22 & \underline{874.6} & 20.79 & 35.5 & 17.59 & \underline{1207.5} & 14.07 & 31.0 & 12.91 & \underline{1498.9} \\
 & REFINE & 25.50 & 32.3 & 8.20 & 814.8 & 20.88 & 35.3 & 17.31 & 1031.7 & 13.96 & 31.2 & 12.91 & 1421.5 \\
 & EAGLES & 25.59 & 32.3 & 8.97 & 754.2 & 20.64 & 35.7 & 21.15 & 1090.3 & 14.06 & 31.0 & 16.21 & 1388.6 \\
\midrule
\multirow{5}{*}{C3DGS} & Ours & 7.48 & 103.3 & 1.15 & \textbf{1030.9} & 7.08 & 104.0 & 0.96 & \textbf{1279.6} & 5.32 & 79.5 & 0.69 & \textbf{1540.8} \\
 & PUP & 8.17 & 94.8 & 1.17 & \underline{906.8} & 7.28 & 100.9 & 0.97 & \underline{1261.8} & 5.86 & 72.2 & 0.67 & \underline{1453.2} \\
 & RAP & 8.80 & 88.6 & 1.14 & 837.9 & 7.68 & 96.1 & 0.97 & 1112.7 & 5.85 & 72.3 & 0.63 & 1380.2 \\
 & REFINE & 8.08 & 98.1 & 1.18 & 823.1 & 7.22 & 102.3 & 0.96 & 1109.2 & 5.24 & 81.1 & 0.66 & 1293.4 \\
 & C3DGS & 10.38 & 74.2 & 1.23 & 789.1 & 10.03 & 73.6 & 1.02 & 1062.0 & 7.10 & 59.8 & 0.70 & 1303.3 \\
\bottomrule
\end{tabular}
\end{adjustbox}

\par\vspace{0.8em}

\begin{adjustbox}{max width=\linewidth}
\begin{tabular}{@{}ll*{8}{r}@{}}
\toprule
 & & \multicolumn{4}{c}{NeRF Synthetic} & \multicolumn{4}{c}{Ref-NeRF} \\
\cmidrule(lr){3-6}\cmidrule(lr){7-10}
\multicolumn{2}{@{}l}{Original 3DGS / Pruned (MB)} & \multicolumn{4}{c}{71.61 / 7.16} & \multicolumn{4}{c}{41.32 / 4.13} \\
\midrule
Method & Input & $S_{\mathrm{cmp}}\downarrow$ & Rat.$\uparrow$ & dec$\downarrow$ & FPS$\uparrow$ & $S_{\mathrm{cmp}}\downarrow$ & Rat.$\uparrow$ & dec$\downarrow$ & FPS$\uparrow$ \\
\midrule
\multirow{5}{*}{LightGS} & Ours & 1.51 & 47.3 & 0.21 & \textbf{2936.7} & 0.89 & 45.9 & 0.16 & \textbf{2472.1} \\
 & PUP & 1.52 & 47.1 & 0.21 & \underline{2769.5} & 0.89 & 46.3 & 0.15 & \underline{2455.7} \\
 & RAP & 1.51 & 47.3 & 0.23 & 2544.9 & 0.89 & 45.9 & 0.15 & 2151.6 \\
 & REFINE & 1.51 & 47.3 & 0.21 & 2293.0 & 0.89 & 45.9 & 0.16 & 2077.9 \\
 & LightGS & 1.52 & 46.9 & 0.23 & 2195.4 & 0.90 & 45.3 & 0.15 & 2085.7 \\
\midrule
\multirow{5}{*}{MesonGS} & Ours & 1.06 & 72.2 & 0.75 & \textbf{2993.8} & 0.95 & 39.3 & 0.59 & \textbf{2562.1} \\
 & PUP & 1.06 & 72.2 & 0.73 & \underline{2769.7} & 0.95 & 39.3 & 0.56 & \underline{2491.7} \\
 & RAP & 1.06 & 72.2 & 0.67 & 2679.6 & 0.95 & 39.3 & 0.53 & 2231.8 \\
 & REFINE & 1.06 & 72.3 & 0.67 & 2396.9 & 0.95 & 39.3 & 0.52 & 2155.2 \\
 & MesonGS & 1.06 & 72.2 & 0.70 & 2590.0 & 0.95 & 39.3 & 0.57 & 2451.8 \\
\midrule
\multirow{5}{*}{EAGLES} & Ours & 3.48 & 20.5 & 4.13 & \textbf{2881.6} & 2.41 & 15.9 & 2.40 & \textbf{2482.8} \\
 & PUP & 3.30 & 21.7 & 3.69 & \underline{2787.4} & 2.32 & 16.4 & 2.37 & \underline{2464.2} \\
 & RAP & 3.22 & 22.3 & 3.11 & 2656.4 & 2.29 & 16.6 & 2.01 & 2307.5 \\
 & REFINE & 3.18 & 22.6 & 3.02 & 2451.2 & 2.16 & 17.8 & 2.09 & 2171.1 \\
 & EAGLES & 3.28 & 21.9 & 3.71 & 2649.3 & 2.30 & 16.8 & 2.31 & 2421.1 \\
\midrule
\multirow{5}{*}{C3DGS} & Ours & 1.21 & 59.5 & 0.33 & \textbf{2986.0} & 0.79 & 51.5 & 0.30 & \textbf{2641.5} \\
 & PUP & 1.31 & 55.4 & 0.33 & \underline{2803.9} & 0.83 & 48.8 & 0.29 & 2583.5 \\
 & RAP & 1.23 & 58.1 & 0.31 & 2680.3 & 0.77 & 51.8 & 0.29 & 2291.9 \\
 & REFINE & 1.20 & 60.1 & 0.31 & 2456.7 & 0.75 & 53.4 & 0.29 & 2263.8 \\
 & C3DGS & 1.46 & 49.0 & 0.33 & 2783.3 & 0.87 & 47.9 & 0.30 & \underline{2636.8} \\
\bottomrule
\end{tabular}
\end{adjustbox}
\endgroup
\end{table}

\subsubsection{Workload Analysis}
\label{app:fps-workload}

Smaller 3D volumes and Gaussian count do not imply less rendering cost. We measure projected overlap and compositing workload by replaying rasterization for all test views of each scene. We first average equally over views within each scene, then equally over scenes within each dataset (Table~\ref{tab:fps-workload}).

Let $P=WH$, $\mathcal G$ be the retained projection-cache Gaussians, and
$\mathcal L_t$ the depth-sorted candidate list of a $16\times16$-pixel tile.
For pixel $p$, let $h_i(p)$ indicate valid support within Gaussian $i$'s
tile bounds, with nonpositive Gaussian exponent and
$\alpha_i(p)=\min(0.99,o_i\exp(q_i(p)))\ge1/255$, ignoring occlusion.
Let $e(p)$ count inspected candidates, including skipped candidates and
the item triggering early termination; $b_i(p)$ indicates an executed
color accumulation. Accumulation stops when the next transmittance is
below $10^{-4}$, excluding the triggering item from $b_i(p)$.
Writing $E=\sum_p e(p)$, $E_0=\sum_p|\mathcal L_{t(p)}|$, and
$B=\sum_{p,i}b_i(p)$, the reported quantities are
\[
\begin{aligned}
\mathrm{Cov.}&=\frac{1}{P}\sum_{p,i}h_i(p), &
\mathrm{ET}&=100\!\left(1-\frac{E}{\max(1,E_0)}\right), &
\mathrm{Checks}&=\frac{E}{P},\\
\mathrm{Pairs}&=10^{-6}\sum_t|\mathcal L_t|, &
\mathrm{Active}&=100\frac{|\{i:\sum_p b_i(p)>0\}|}{\max(1,|\mathcal G|)}, &
\mathrm{Blends}&=\frac{B}{P}.
\end{aligned}
\]
ET and Active are percentages; Active counts any executed accumulation,
not perceptual visibility. Checks includes candidates that never blend,
whereas Blends counts only executed accumulations ($B\le E\le E_0$).
Pairs is measured over all test views.

Ours has the lowest projected coverage on all five datasets and the
fewest candidate checks and Gaussian--tile pairs on four, with PUP slightly
lower on Deep Blending. Ours also has the fewest blending operations on
four datasets; RAP is slightly lower on Tanks \& Temples. On Mip-NeRF 360,
relative to PUP, Ours reduces projected coverage, candidate checks,
Gaussian--tile pairs, and blending operations by 19.2\%, 16.6\%, 23.4\%,
and 10.1\%, respectively.

These observations support reduced screen-space overlap and rendering
workload as an explanation for higher FPS.

\begingroup
\small
\setlength{\tabcolsep}{3pt}
\renewcommand{\arraystretch}{1.05}
\setlength{\LTleft}{0pt}
\setlength{\LTright}{0pt}
\setlength{\LTcapwidth}{\linewidth}
\begin{longtable}{@{}p{0.21\linewidth}p{\dimexpr0.13\linewidth-2\tabcolsep\relax}*{6}{>{\raggedleft\arraybackslash}p{\dimexpr0.11\linewidth-2\tabcolsep\relax}}@{}}
\caption{Dataset-level rendering workload. Ours generally reduces projected
overlap and rendering operations. \textbf{Bold} and \underline{underline} mark the lowest and
second-lowest displayed workload values, including ties.}
\label{tab:fps-workload}\\
\toprule
Dataset & Method & Cov.$\downarrow$ & \shortstack{ET\\(\%)} & Checks$\downarrow$ &
\shortstack{Pairs$\downarrow$\\(M)} & \shortstack{Active\\(\%)} & Blends$\downarrow$ \\
\midrule
\endfirsthead
\multicolumn{8}{c}{Table~\thetable{}(continued)}\\
\toprule
Dataset & Method & Cov.$\downarrow$ & \shortstack{ET\\(\%)} & Checks$\downarrow$ &
\shortstack{Pairs$\downarrow$\\(M)} & \shortstack{Active\\(\%)} & Blends$\downarrow$ \\
\midrule
\endhead
\midrule
\multicolumn{8}{r}{Continued on next page}\\
\endfoot
\bottomrule
\endlastfoot
\multirow{10}{*}{Mip-NeRF 360} & Ours & \textbf{44.88} & 18.44 & \textbf{315.93} & \textbf{2.581} & 88.63 & \textbf{34.36} \\*
 & PUP & \underline{55.57} & 25.30 & \underline{378.93} & \underline{3.368} & 82.70 & \underline{38.21} \\*
 & RAP & 67.79 & 33.65 & 395.58 & 3.967 & 82.53 & 39.08 \\*
 & REFINE & 69.13 & 33.82 & 402.87 & 4.056 & 81.39 & 39.84 \\*
 & C3DGS & 69.04 & 30.43 & 469.11 & 4.475 & 83.98 & 42.56 \\*
 & CoreSet & 65.05 & 30.29 & 436.14 & 4.157 & 83.69 & 40.21 \\*
 & EAGLES & 68.26 & 29.50 & 517.93 & 4.861 & 83.27 & 42.63 \\*
 & LightGS & 95.54 & 33.58 & 593.98 & 5.923 & 82.26 & 55.12 \\*
 & MesonGS & 69.39 & 29.86 & 511.07 & 4.820 & 83.67 & 43.00 \\*
 & 3DGS & 86.67 & 32.41 & 820.68 & 7.954 & 75.96 & 49.52 \\
\midrule
\multirow{10}{*}{Deep Blending} & Ours & \textbf{46.27} & 11.95 & \underline{339.51} & \underline{1.692} & 95.03 & \textbf{38.30} \\*
 & PUP & \underline{49.80} & 13.26 & \textbf{329.41} & \textbf{1.672} & 94.83 & \underline{39.57} \\*
 & RAP & 70.15 & 27.23 & 402.71 & 2.426 & 93.29 & 44.22 \\*
 & REFINE & 74.63 & 28.40 & 424.17 & 2.593 & 91.96 & 45.83 \\*
 & C3DGS & 69.24 & 20.81 & 451.11 & 2.505 & 92.96 & 48.36 \\*
 & CoreSet & 67.01 & 21.99 & 450.77 & 2.533 & 92.00 & 46.57 \\*
 & EAGLES & 63.02 & 17.73 & 444.47 & 2.379 & 92.83 & 46.38 \\*
 & LightGS & 101.20 & 20.56 & 639.04 & 3.516 & 91.48 & 74.23 \\*
 & MesonGS & 66.76 & 19.06 & 454.64 & 2.473 & 92.99 & 47.75 \\*
 & 3DGS & 94.09 & 25.16 & 879.78 & 5.138 & 87.78 & 60.60 \\
\midrule
\multirow{10}{*}{Tanks \& Temples} & Ours & \textbf{68.02} & 27.31 & \textbf{485.90} & \textbf{1.423} & 75.75 & \underline{48.82} \\*
 & PUP & \underline{78.11} & 35.21 & \underline{502.51} & \underline{1.664} & 63.35 & 49.03 \\*
 & RAP & 90.09 & 40.97 & 516.02 & 1.857 & 67.50 & \textbf{48.64} \\*
 & REFINE & 94.55 & 40.76 & 558.38 & 2.013 & 57.56 & 52.03 \\*
 & C3DGS & 94.82 & 36.05 & 641.45 & 2.137 & 67.68 & 55.97 \\*
 & CoreSet & 88.92 & 35.00 & 611.96 & 2.011 & 72.26 & 53.41 \\*
 & EAGLES & 91.21 & 35.12 & 653.72 & 2.144 & 68.77 & 55.10 \\*
 & LightGS & 129.28 & 40.19 & 718.03 & 2.549 & 68.82 & 70.98 \\*
 & MesonGS & 93.30 & 35.64 & 652.65 & 2.158 & 68.94 & 55.60 \\*
 & 3DGS & 123.38 & 41.58 & 1165.29 & 4.221 & 62.58 & 64.50 \\
\midrule
\multirow{10}{*}{NeRF Synthetic} & Ours & \textbf{11.12} & 10.83 & \textbf{106.62} & \textbf{0.297} & 96.80 & \textbf{9.37} \\*
 & PUP & \underline{17.23} & 23.36 & \underline{119.99} & \underline{0.399} & 93.49 & \underline{11.22} \\*
 & RAP & 24.92 & 33.28 & 121.81 & 0.455 & 89.52 & 11.63 \\*
 & REFINE & 42.36 & 43.42 & 145.48 & 0.717 & 87.36 & 12.89 \\*
 & C3DGS & 23.23 & 30.21 & 136.20 & 0.489 & 92.96 & 12.44 \\*
 & CoreSet & 24.15 & 28.06 & 136.62 & 0.489 & 95.66 & 12.76 \\*
 & EAGLES & 19.61 & 24.13 & 140.84 & 0.460 & 94.66 & 11.99 \\*
 & LightGS & 45.79 & 38.25 & 193.48 & 0.820 & 91.75 & 17.20 \\*
 & MesonGS & 21.02 & 26.52 & 141.25 & 0.479 & 94.18 & 12.23 \\*
 & 3DGS & 65.72 & 37.40 & 508.84 & 2.069 & 73.24 & 18.79 \\
\midrule
\multirow{10}{*}{Ref-NeRF} & Ours & \textbf{30.99} & 24.60 & \textbf{169.18} & \textbf{0.614} & 88.20 & \textbf{17.08} \\*
 & PUP & \underline{38.60} & 31.91 & \underline{180.64} & \underline{0.734} & 85.07 & 18.33 \\*
 & RAP & 53.85 & 45.96 & 192.34 & 0.908 & 77.44 & \underline{17.65} \\*
 & REFINE & 77.25 & 55.38 & 206.25 & 1.169 & 71.62 & 19.71 \\*
 & C3DGS & 46.39 & 35.90 & 202.67 & 0.877 & 85.54 & 19.43 \\*
 & CoreSet & 42.65 & 29.41 & 214.14 & 0.791 & 95.80 & 20.83 \\*
 & EAGLES & \underline{38.60} & 30.21 & 202.04 & 0.805 & 87.52 & 18.44 \\*
 & LightGS & 86.32 & 49.07 & 273.14 & 1.384 & 79.57 & 24.23 \\*
 & MesonGS & 43.56 & 33.80 & 206.83 & 0.869 & 85.71 & 19.03 \\*
 & 3DGS & 99.61 & 47.76 & 551.87 & 2.697 & 67.34 & 28.02 \\
\end{longtable}
\endgroup

\subsubsection{Spatial Balance of Reconstruction Quality}
\label{app:spatial-quality-balance}

For Mip-NeRF 360, Deep Blending, and Tanks \& Temples, all pixels
are eligible. For NeRF Synthetic and Ref-NeRF, a four-connected
flood fill on GT excludes boundary-connected white pixels
(all RGB channels $\ge250$).

SSIM is computed between rendered and GT RGB images normalized to $[0,1]$,
using an $11\times11$ Gaussian window ($\sigma=1.5$), stride one, zero
padding, and constants $C_1=0.01^2$, $C_2=0.03^2$. Channel-wise local
SSIM values are averaged over RGB before spatial statistics are taken
over eligible pixels; values are not clipped.

For PSNR, each full image is partitioned into 16 rows and 16 columns,
yielding 256 nonoverlapping cells.
For image dimensions $H\times W$, cell boundaries are
$\lfloor kH/16\rfloor$ and $\lfloor kW/16\rfloor$, $k=0,\ldots,16$.
Cells containing no eligible pixels are excluded. For eligible pixels
$F_j$ in cell $j$, native 8-bit RGB
images $X,Y\in[0,255]$ give
\[
m_j=\frac{1}{3|F_j|}\sum_{p\in F_j}\sum_c
       (X_c(p)-Y_c(p))^2,
\qquad
z_j=\begin{cases}
100, & m_j=0,\\
10\log_{10}(255^2/m_j), & m_j>0.
\end{cases}
\]
Only infinite PSNR values are replaced by 100\,dB; finite values are
not capped. Each valid cell has equal weight regardless of its size.
For either the eligible SSIM values or cell PSNR values
$z_1,\ldots,z_n$, variance and range are
\[
\bar z=\frac1n\sum_i z_i,\qquad
V=\frac1n\sum_i(z_i-\bar z)^2,\qquad
D=\max_i z_i-\min_i z_i.
\]
Variance uses the population convention, and range is computed within
each image. We first averages
test images equally within each scene, then averages scenes equally
within each dataset. SSIM statistics are unitless. PSNR variance and
range use dB$^2$ and dB, respectively.

In the current table, among the pruned methods, Ours achieves the lowest PSNR
variance and range on all five datasets. It also has the lowest SSIM
variance on four datasets, with PUP lower on Deep Blending. SSIM range
shows a less uniform advantage: Ours is lowest on Mip-NeRF 360 and NeRF
Synthetic, while PUP is lowest on Deep Blending and MesonGS on the
remaining two datasets.

These results support more balanced regional reconstruction, particularly
in PSNR. Lower dispersion should be interpreted alongside
average quality, rather than as evidence of higher quality on its own.

\begingroup
\small
\setlength{\tabcolsep}{3pt}
\renewcommand{\arraystretch}{1.05}
\setlength{\LTleft}{0pt}
\setlength{\LTright}{0pt}
\setlength{\LTcapwidth}{\linewidth}
\begin{longtable}{@{}p{0.21\linewidth}p{\dimexpr0.13\linewidth-2\tabcolsep\relax}*{4}{>{\raggedleft\arraybackslash}p{\dimexpr0.165\linewidth-2\tabcolsep\relax}}@{}}
\caption{Dispersion of reconstruction quality across five datasets.
PSNR uses a $16\times16$ grid and its variance and range are reported
in dB$^2$ and dB, respectively. Lower values indicate less
dispersion. Best and second-best displayed values among pruned methods
are \textbf{bold} and \underline{underlined}, respectively, including ties.}
\label{tab:spatial-quality-balance}\\
\toprule
Dataset & Method & SSIM var.$\downarrow$ & \shortstack{PSNR var.$\downarrow$} &
SSIM range$\downarrow$ & \shortstack{PSNR range$\downarrow$} \\
\midrule
\endfirsthead
\multicolumn{6}{c}{Table~\thetable{} (continued)}\\
\toprule
Dataset & Method & SSIM var.$\downarrow$ & \shortstack{PSNR var.$\downarrow$} &
SSIM range$\downarrow$ & \shortstack{PSNR range$\downarrow$} \\
\midrule
\endhead
\midrule
\multicolumn{6}{r}{Continued on next page}\\
\endfoot
\bottomrule
\endlastfoot
\multirow{10}{*}{Mip-NeRF 360} & Ours & \textbf{0.0288} & \textbf{18.76} & \textbf{1.5498} & \textbf{23.63} \\*
 & PUP & \underline{0.0304} & \underline{21.25} & \underline{1.5699} & 24.72 \\*
 & RAP & 0.0345 & 21.60 & 1.5965 & \underline{24.59} \\*
 & REFINE & 0.0407 & 25.89 & 1.6290 & 25.64 \\*
 & C3DGS & 0.0319 & 24.21 & 1.5770 & 25.64 \\*
 & CoreSet & 0.0365 & 26.38 & 1.5993 & 26.08 \\*
 & EAGLES & 0.0321 & 24.06 & 1.5708 & 25.71 \\*
 & LightGS & 0.0347 & 24.18 & 1.5792 & 25.49 \\*
 & MesonGS & 0.0315 & 23.80 & 1.5715 & 25.47 \\*
 & 3DGS & 0.0298 & 22.73 & 1.5730 & 25.39 \\
\midrule
\multirow{10}{*}{Deep Blending} & Ours & \underline{0.0141} & \textbf{23.13} & \underline{1.4574} & \textbf{25.30} \\*
 & PUP & \textbf{0.0139} & 23.55 & \textbf{1.4489} & \underline{25.53} \\*
 & RAP & 0.0170 & \underline{23.25} & 1.5254 & 25.82 \\*
 & REFINE & 0.0221 & 25.84 & 1.5676 & 26.51 \\*
 & C3DGS & 0.0152 & 24.04 & 1.4610 & 25.61 \\*
 & CoreSet & 0.0188 & 25.13 & 1.4964 & 25.93 \\*
 & EAGLES & 0.0155 & 24.22 & 1.4754 & 25.59 \\*
 & LightGS & 0.0171 & 25.07 & 1.4800 & 25.98 \\*
 & MesonGS & 0.0154 & 24.13 & 1.4848 & 25.60 \\*
 & 3DGS & 0.0142 & 23.56 & 1.4881 & 25.65 \\
\midrule
\multirow{10}{*}{Tanks \& Temples} & Ours & \textbf{0.0412} & \textbf{30.41} & 1.6367 & \textbf{29.02} \\*
 & PUP & 0.0496 & 33.55 & 1.6456 & \underline{29.34} \\*
 & RAP & 0.0471 & 31.07 & 1.6363 & 29.71 \\*
 & REFINE & 0.0560 & 33.47 & 1.6662 & 29.71 \\*
 & C3DGS & 0.0432 & 31.74 & \underline{1.6347} & 29.75 \\*
 & CoreSet & 0.0479 & 31.36 & 1.6644 & 29.40 \\*
 & EAGLES & 0.0456 & 33.00 & 1.6367 & 29.83 \\*
 & LightGS & 0.0457 & \underline{31.02} & 1.6542 & 29.98 \\*
 & MesonGS & \underline{0.0420} & 31.21 & \textbf{1.6342} & 29.68 \\*
 & 3DGS & 0.0363 & 30.26 & 1.6496 & 29.02 \\
\midrule
\multirow{10}{*}{NeRF Synthetic} & Ours & \textbf{0.0263} & \textbf{54.39} & \textbf{1.2443} & \textbf{32.13} \\*
 & PUP & 0.0274 & \underline{58.33} & 1.2889 & \underline{33.08} \\*
 & RAP & 0.0378 & 58.82 & 1.4022 & 33.22 \\*
 & REFINE & 0.0422 & 62.86 & 1.4468 & 34.76 \\*
 & C3DGS & 0.0272 & 59.90 & 1.2862 & 33.70 \\*
 & CoreSet & 0.0396 & 62.30 & 1.4173 & 34.95 \\*
 & EAGLES & 0.0270 & 59.77 & 1.2805 & 33.71 \\*
 & LightGS & 0.0415 & 62.70 & 1.4105 & 34.47 \\*
 & MesonGS & \underline{0.0269} & 59.62 & \underline{1.2699} & 33.71 \\*
 & 3DGS & 0.0221 & 58.60 & 1.2315 & 33.71 \\
\midrule
\multirow{10}{*}{Ref-NeRF} & Ours & \textbf{0.0405} & \textbf{43.65} & \underline{1.5152} & \textbf{29.93} \\*
 & PUP & \underline{0.0406} & \underline{45.41} & 1.5167 & \underline{30.14} \\*
 & RAP & 0.0446 & 46.98 & 1.5233 & 30.72 \\*
 & REFINE & 0.0502 & 54.61 & 1.5476 & 32.15 \\*
 & C3DGS & 0.0413 & 47.20 & 1.5161 & 30.87 \\*
 & CoreSet & 0.0527 & 52.29 & 1.5356 & 32.64 \\*
 & EAGLES & 0.0414 & 47.57 & 1.5158 & 31.45 \\*
 & LightGS & 0.0452 & 50.63 & 1.5165 & 31.46 \\*
 & MesonGS & 0.0410 & 46.60 & \textbf{1.5113} & 30.86 \\*
 & 3DGS & 0.0413 & 49.37 & 1.5378 & 30.70 \\
\end{longtable}
\endgroup

\end{document}